\documentclass[9pt,twocolumn,letterpaper]{article}
\usepackage{graphicx,epsfig,fullpage,svg}
\usepackage[hyphens]{url}
\usepackage{caption}
\usepackage{subfig}
\usepackage{float}
\usepackage{comment}
\usepackage{caption}
\usepackage{mathtools} 
\usepackage{adjustbox}
\usepackage{tabularx}
\usepackage{multirow}
\usepackage{booktabs} 
\usepackage{hyperref}
\usepackage{dirtytalk}

\title{TRECVID 2020: A comprehensive campaign for evaluating video retrieval tasks across multiple application domains }
\author{
George Awad
  \{gawad@nist.gov\}\\
  Georgetown University;\\ Information Access Division,
National Institute of Standards and Technology, USA\\\\
Asad A. Butt
  \{asad.butt@nist.gov\}\\
  Johns Hopkins University;\\ Information Access Division,
National Institute of Standards and Technology, USA\\\\
Keith Curtis
  \{keith.curtis@nist.gov\}\\
  Information Access Division,
National Institute of Standards and Technology, USA\\\\
Jonathan Fiscus
  \{jfiscus@nist.gov\}
Afzal Godil
   \{godil@nist.gov\}\\
Yooyoung Lee
  \{yooyoung@nist.gov\}
Andrew Delgado
  \{andrew.delgado@nist.gov\}\\
Jesse Zhang
  \{jesse.zhang@nist.gov\}\
Eliot Godard
  \{eliot.godard@nist.gov\}\\
Baptiste Chocot 
  \{baptiste.chocot@nist.gov\}\\
Information Access Division,
National Institute of Standards and Technology, USA\\\\
Lukas Diduch
  \{lukas.diduch@nist.gov\}\\
  Dakota-consulting,  USA\\\\
Jeffrey Liu
  \{jeffrey.liu@ll.mit.edu\}\\
  MIT Lincoln Laboratory, USA\\\\
Alan F. Smeaton
  \{alan.smeaton@dcu.ie\}\\
 Insight Centre, School of Computing, Dublin City University, Ireland\\\\
Yvette Graham
  \{graham.yvette@gmail.com\}\\
 ADAPT Centre, School of Computing, Dublin City University, Ireland\\\\
Gareth J. F. Jones
  \{gareth.jones@dcu.ie\}\\
  ADAPT Centre, School of Computer Science and Statistics, Trinity College Dublin, Ireland\\\\
Wessel Kraaij
  \{w.kraaij@liacs.leidenuniv.nl\}\\
  Leiden University; TNO, Netherlands\\\\
Georges Qu\'{e}not
  \{Georges.Quenot@imag.fr\}\\
  Laboratoire d'Informatique de Grenoble, France\\\\
}

\newlength{\spacelen}
\begin{document}
\graphicspath{{tv20.figures/}}
\maketitle

\section{Introduction}
The TREC Video Retrieval Evaluation (TRECVID) is a TREC-style
video analysis and retrieval evaluation with the goal of
promoting progress in research and development of content-based exploitation and retrieval of 
information from digital video via open, metrics-based evaluation. 

Over the last twenty years this effort has yielded a better understanding of how systems can effectively
accomplish such processing and how one can reliably benchmark their performance. 
TRECVID has been funded by NIST (National Institute of Standards and Technology) and other US government agencies. In addition, many organizations and individuals worldwide contribute significant time and effort.

TRECVID 2020 represented a continuation of four tasks and the addition of two new tasks. In total, 52 teams from various research organizations worldwide signed up to join the evaluation campaign this year, where 29 teams
(Table \ref{participants}) completed one or more of the following six tasks, and 23 teams registered but did not submit any runs (Table \ref{nonparticipants}):

\begin{enumerate} \itemsep0pt \parskip0pt
\item Ad-hoc Video Search (AVS)
\item Instance Search (INS)
\item Disaster Scene Description and Indexing (DSDI)
\item Video to Text Description (VTT)
\item Activities in Extended Video (ActEV)
\item Video Summarization (VSUM)
\end{enumerate}

This year TRECVID continued the usage of the Vimeo Creative Commons collection dataset (V3C1) \cite{rossetto2019v3c} of about 1000 hours in total and segmented into 1 million short video shots to support the Ad-hoc video search task. The dataset is drawn from the Vimeo video sharing website under the Creative Commons licenses and reflects a wide variety of content, style, and source device determined only by the self-selected donors. 

The Instance Search task continued working with the 464 hours of the BBC (British Broadcasting Corporation) EastEnders video as used before since 2013, while the Video to Text description task started using a subset of 1700 short videos from the Vimeo V3C2 dataset.

For the Activities in Extended Video task, about 10 hours of the VIRAT (Video and Image Retrieval and Analysis Tool) dataset was used which was designed to be realistic, natural and challenging for video surveillance domains in terms of its resolution, background clutter, diversity in scenes, and human activity/event categories.

The new Video Summarization task also made use of the BBC Eastenders dataset, while the DSDI task worked on public natural disaster 5\thinspace h videos collected from a Nepal earthquake event in 2015. 

The Ad-hoc search, Instance Search, and Video Summarization results were judged by NIST human assessors, while the Video to Text was annotated by NIST human assessors and scored automatically later on using Machine Translation (MT) metrics and Direct Assessment (DA) by Amazon Mechanical Turk workers on sampled runs. The Disaster Scene Description and Indexing task was also annotated by human assessors and scored automatically using Mean Average Precision (MAP).

The systems submitted for the ActEV (Activities in Extended Video) evaluations were scored by NIST using reference annotations created by Kitware, Inc.

\begin{table*} 
\caption{Participants and tasks}
\label{participants}
  \vspace{1.0cm}
  \centering{
   \scriptsize{
    \begin{tabular}{|c|c|c|c|c|c|l|l|l|}
      \hline 
\multicolumn{6}{|c|}{Task}&Location&TeamID&Participants \\
      \hline 
$IN$ & $VT$ & $AV$ & $AH$ &  $DS$& $VS$ &  &  &    \\
\hline

$--$ & $--$ & $--$ & $--$ & $DS$ & $--$ & $Eur$ & $VCL$&Information Technologies Institute \\&&&&&&&& (ITI) Centre of Research and \\&&&&&&&& Technology Hellas (CERTH)\\
$--$ & $VT$ & $--$ & $--$ & $--$ & $--$ & $Eur$ & $PicSOM$&Aalto University\\
$IN$ & $--$ & $AV$ & $--$ & $**$ & $--$ & $Asia$ & $BUPT\_MCPRL$&Beijing University of Posts \\&&&&&&&& and Telecommunications\\
$--$ & $--$ & $**$ & $AH$ & $--$ & $--$ & $Asia$ & $VIdeoREtrievalGrOup$&City University of Hong Kong\\
$--$ & $VT$ & $--$ & $**$ & $--$ & $--$ & $SAm$ & $IMFD\_IMPRESEE$&University of Chile; Millennium \\&&&&&&&& Institute of Data Foundation \\&&&&&&&& (IMFD), Chile; Impresee Inc, Chile\\
$--$ & $--$ & $--$ & $--$ & $--$ & $VS$ & $Eur$ & $MeMAD$&Eurecom and Aalto for MeMAD\\
$--$ & $--$ & $--$ & $AH$ & $DS$ & $--$ & $NAm$ & $FIU\_UM$&Florida International University;\\&&&&&&&& University of Miami\\
$--$ & $--$ & $AV$ & $AH$ & $--$ & $--$ & $Asia$ & $kindai_ogu$& Kindai University; \\&&&&&&&& Osaka Gakuin University\\
$--$ & $--$ & $**$ & $--$ & $DS$ & $--$ & $Asia$ & $VAS$&Hitachi, Ltd. R\&D\\
$--$ & $--$ & $--$ & $AH$ & $--$ & $--$ & $Asia$ & $DVA\_Researchers$&Indian Institute of Space \\&&&&&&&& Science \& Technology (IIST), \\&&&&&&&& Thiruvananthapuram Development \\&&&&&&&& and Educational Communication \\&&&&&&&& Unit  (DECU), Indian Space \\&&&&&&&& Research Organisation (ISRO)\\
$--$ & $--$ & $AV$ & $AH$ & $DS$ & $--$ & $Eur$ & $ITI\_CERTH$&Information Technologies Institute \\&&&&&&&&, Centre for Research and \\&&&&&&&& Technology Hellas\\
$--$ & $VT$ & $--$ & $--$ & $--$ & $--$ & $Asia$ & $KU\_ISPL$&korea university\\
$--$ & $--$ & $--$ & $--$ & $DS$ & $--$ & $Eur$ & $SHIELD$&LINKS Foundation\\
$--$ & $VT$ & $--$ & $--$ & $--$ & $--$ & $Asia$ & $KsLab\_NUT$&Nagaoka University of Technology\\
$--$ & $--$ & $--$ & $--$ & $DS$ & $--$ & $Asia$ & $NIIICT$&National Institute of Information \\&&&&&&&& and Communications Technology \\&&&&&&&& (Japan), and National Institute of \\&&&&&&&& Informatics (Japan)\\
$IN$ & $**$ & $**$ & $**$ & $DS$ & $VS$ & $Asia$ & $NII\_UIT$&National Institute of Informatics\\&&&&&&&&, Japan; University of Information \\&&&&&&&& Technology, VNU-HCMC, Vietnam\\
$IN$ & $**$ & $--$ & $--$ & $**$ & $--$ & $Asia$ & $PKU\_WICT$&Peking University\\
$--$ & $VT$ & $--$ & $AH$ & $--$ & $--$ & $Asia$ & $RUC\_AIM3$&Renmin University of China\\
$--$ & $--$ & $--$ & $AH$ & $--$ & $--$ & $Asia$ & $RUCMM$&Renmin University of China\\
$IN$ & $--$ & $AV$ & $**$ & $--$ & $--$ & $Asia$ & $UEC$&The University of \\&&&&&&&& Electro-Communications, Tokyo\\
$--$ & $--$ & $AV$ & $--$ & $--$ & $--$ & $Asia$ & $TokyoTech\_AIST$&Tokyo Institute of Technology\\&&&&&&&&, National Institute of Advanced \\&&&&&&&& Industrial Science and \\&&&&&&&& Technology (AIST)\\
$--$ & $VT$ & $--$ & $--$ & $--$ & $--$ & $Eur$ & $MMCUniAugsburg$&University of Augsburg\\
$--$ & $**$ & $--$ & $--$ & $DS$ & $**$ & $Aus$ & $UTSVideo$&University of Technology Sydney\\
$**$ & $--$ & $--$ & $--$ & $DS$ & $--$ & $NAm$ & $COVIS$&UNT College of Engineering;\\&&&&&&&& UNT Dept. of Computer Science \\&&&&&&&& and Engineering; UNT Dept. of \\&&&&&&&& Electrical Engineering\\
$--$ & $**$ & $**$ & $AH$ & $--$ & $**$ & $Asia$ & $WasedaMeiseiSoftbank$&Waseda University; \\&&&&&&&& Meisei University;\\&&&&&&&& SoftBank Corporation\\
$IN$ & $--$ & $--$ & $--$ & $--$ & $--$ & $Asia$ & $WHU\_NERCMS$&Wuhan University\\
$--$ & $**$ & $--$ & $AH$ & $--$ & $--$ & $Asia$ & $ZY\_BJLAB$&XinHuaZhiYun Technology\\
$--$ & $--$ & $AV$ & $--$ & $--$ & $--$ & $NAm$ & $INF$&Carnegie Mellon University\\
$--$ & $--$ & $AV$ & $--$ & $--$ & $--$ & $NAm$ & $CRCV\_UCF$&University of Central Florida\\

      \hline 
   \end{tabular} \\
\vspace{.2cm}
Task legend. IN:Instance Search; VT:Video to Text; AV:Activities in Extended videos; AH:Ad-hoc search; DS: Disaster Scene Description and Indexing; VS: Video Summarization; $--$:no run planned; $***$:planned but not submitted\\
   } 
  } 
\end{table*}

\begin{table*} 
\caption{Participants who did not submit any runs}
\label{nonparticipants}
  \vspace{1.0cm}
  \centering{
   \scriptsize{
    \begin{tabular}{|c|c|c|c|c|c|l|l|l|}
      \hline 
\multicolumn{6}{|c|}{Task}&Location&TeamID&Participants \\
      \hline 
        $IN$ & $VT$ & $AV$ & $AH$ & $DS$ & $VS$ &         &               &    \\
        \hline

$--$ & $**$ & $--$ & $**$ & $--$ & $**$ & $Asia$ & $ATL$&Alibaba group, \\&&&&&&&& ZheJiang University\\
$--$ & $**$ & $--$ & $--$ & $--$ & $--$ & $NAm$ & $Arete$&Arete\_Associates\\
$--$ & $--$ & $**$ & $--$ & $--$ & $--$ & $Eur+Asia$ & $SYMBEN$&Athlone Institute of Technology, \\&&&&&&&& Ireland Aligarh Muslim \\&&&&&&&& University, India Lahore College \\&&&&&&&& for Women Univesity, Lahore, \\&&&&&&&& Pakistan Islamia University \\&&&&&&&& Bahawalpur, Pakistan\\
$--$ & $**$ & $--$ & $--$ & $--$ & $--$ & $Asia$ & $BDVIDEO$&BAIDU\\
$--$ & $**$ & $--$ & $--$ & $--$ & $**$ & $Asia$ & $NDKS$ &Charotar University Of \\&&&&&&&& Science \& Technology\\
$--$ & $--$ & $**$ & $--$ & $--$ & $--$ & $Asia$ & $Byte\_Karma$&CHARUSAT\\
$--$ & $--$ & $--$ & $**$ & $--$ & $--$ & $Asia$ & $UPC\_VIT2020$&China university of petroleum \\&&&&&&&& (east China)\\
$**$ & $--$ & $--$ & $**$ & $**$ & $--$ & $NAm$ & $VCUB$&CSE Dept UB\\
$--$ & $--$ & $--$ & $**$ & $--$ & $**$ & $NAm$ & $drylwlsn\_visual$&drylwlsn\_visual\\
$--$ & $--$ & $**$ & $--$ & $**$ & $--$ & $Eur$ & $IOSBVID\_TV20$&Fraunhofer IOSB Research \\&&&&&&&& Institute Karlsruhe Institute \\&&&&&&&& of Technology\\
$--$ & $--$ & $--$ & $**$ & $--$ & $--$ & $NAm$ & $ark\_20$&Huawei Noah's Ark lab\\
$**$ & $**$ & $**$ & $**$ & $**$ & $**$ & $Asia$ & $aalekhn$&Independent Researcher\\
$--$ & $--$ & $**$ & $--$ & $--$ & $--$ & $NAm$ & $usf\_bulls$&Institute for Artificial \\&&&&&&&& Intelligence (AI+X), University of \\&&&&&&&& South Florida\\
$--$ & $**$ & $--$ & $**$ & $--$ & $**$ & $Asia$ & $KNU.visual\_lab$&Kangwon national university\\
$--$ & $**$ & $--$ & $**$ & $--$ & $--$ & $Eur$ & $LIG$&Multimedia Information Modeling \\&&&&&&&& and Retrieval group of \\&&&&&&&& LIG Explainable and Responsible \\&&&&&&&& Artificial Intelligence Chair of \\&&&&&&&& the MIAI Institute.\\
$**$ & $**$ & $**$ & $**$ & $**$ & $**$ & $Asia$ & $DAMILAB$&NIT Warangal\\
$--$ & $--$ & $--$ & $--$ & $**$ & $**$ & $Asia$ & $PKUMI$&Peking University\\
$--$ & $**$ & $--$ & $--$ & $--$ & $**$ & $Afr$ & $REGIM\_Lab\_VSUM$&Research Groups in \\&&&&&&&& Intelligent Machines\\
$**$ & $--$ & $--$ & $**$ & $--$ & $--$ & $Eur$ & $AIT\_SRI\_2020$&Software Research Institute \\&&&&&&&& Athlone IT\\
$**$ & $**$ & $--$ & $--$ & $--$ & $--$ & $Eur+Asia$ & $Sheffield\_UETLahore$&University of Sheffield \\&&&&&&&& Department of Computer Science \\&&&&&&&& University of Engineering and \\&&&&&&&& Technology, Lahore \\&&&&&&&& Department of Computer Science\\
$--$ & $--$ & $--$ & $**$ & $--$ & $--$ & $Asia$ & $ustcmcc$&University of Science and \\&&&&&&&& Technology of China officially \\&&&&&&&& Huawei Technologies Co. Ltd.\\
$**$ & $--$ & $**$ & $**$ & $**$ & $--$ & $Eur$ & $Aptitude$&Universite de Mons\\
$**$ & $--$ & $--$ & $**$ & $--$ & $--$ & $NAm$ & $VSR$&Visionary Systems and \\&&&&&&&& Research (VSR)\\

       \hline 
    \end{tabular} \\
\vspace{.2cm}
Task legend. IN:Instance Search; VT:Video to Text; AV:Activities in extended videos; AH:Ad-hoc search; DS: Disaster Scene Description and Indexing; VS: Video Summarization; $--$:no run planned; $**$:planned but not submitted\\
   } 
  } 
\end{table*}

This paper is an introduction to the evaluation framework, tasks, data, and measures used in the 2020 evaluation campaign. For detailed information about the approaches and results, the reader should see the various site reports and the results pages available at the workshop proceeding online page \cite{tv20pubs}.
Finally we would like to acknowledge that all work presented here has been cleared by RPO (Research Protection Office) under RPO number: \#ITL-17-0025

\emph{Disclaimer: Certain commercial entities, equipment, or
materials may be identified in this document in order to describe an
experimental procedure or concept adequately. Such identification is
not intended to imply recommendation or endorsement by the National
Institute of Standards and Technology, nor is it intended to imply 
that the entities, materials, or equipment are necessarily the best 
available for the purpose. The views and conclusions contained herein 
are those of the authors and~should not be interpreted as necessarily
representing the official policies or endorsements, either expressed or 
implied, of IARPA (Intelligence Advanced Research Projects Activity), NIST, or the U.S. Government.}

\section{Datasets}
Many datasets have been adopted and used across the years since TRECVID started in 2001 and all available resources and datasets from previous years can be accessed from our website\footnote{https://trecvid.nist.gov/past.data.table.html}. In the following sections we will give an overview of the main datasets used this year across the different tasks.

\subsection{BBC EastEnders Instance Search Dataset}
The BBC in collaboration with the European Union's AXES project made 464 h of the popular and long-running soap opera EastEnders available to TRECVID for research since 2013. 
The data comprise 244 weekly ``omnibus'' broadcast files (divided into 471\,527 shots), transcripts, and a small amount of additional metadata. This dataset was adopted to test systems on retrieving target persons (characters) doing specific everyday actions in the Instance Search task and also adopted for the Video Summarization task to summarize the major events in 3 characters during a time period of about 6 to 8 weeks of episodes.

\subsection{Vimeo Creative Commons Collection (V3C) Dataset}
The V3C1 dataset (drawn from a larger V3C video dataset \cite{rossetto2019v3c}) is composed of 7475 Vimeo videos (1.3 TB, 1000 h) with Creative Commons licenses and mean duration of 8 min. All videos have some metadata available such as title, keywords, and description in json files. The dataset has been segmented into 1\,082\,657 short video segments according to the provided master shot boundary files. In addition, keyframes and thumbnails per video segment have been extracted and made available. While the V3C1 dataset was adopted for testing the Ad-hoc video search systems, the previous Internet Archive datasets (IACC.1-3) of about 1800 h were available for development and training. In addition to the above, a small subset of 1700 short videos from V3C2 dataset (also drawn from the V3C video dataset) was used to test the Video to Text systems.

\subsection{Activity Detection VIRAT Dataset} 
\begin{figure}[hbt]
\begin{centering}
\includegraphics[width=3.16667in,height=1.86000in]{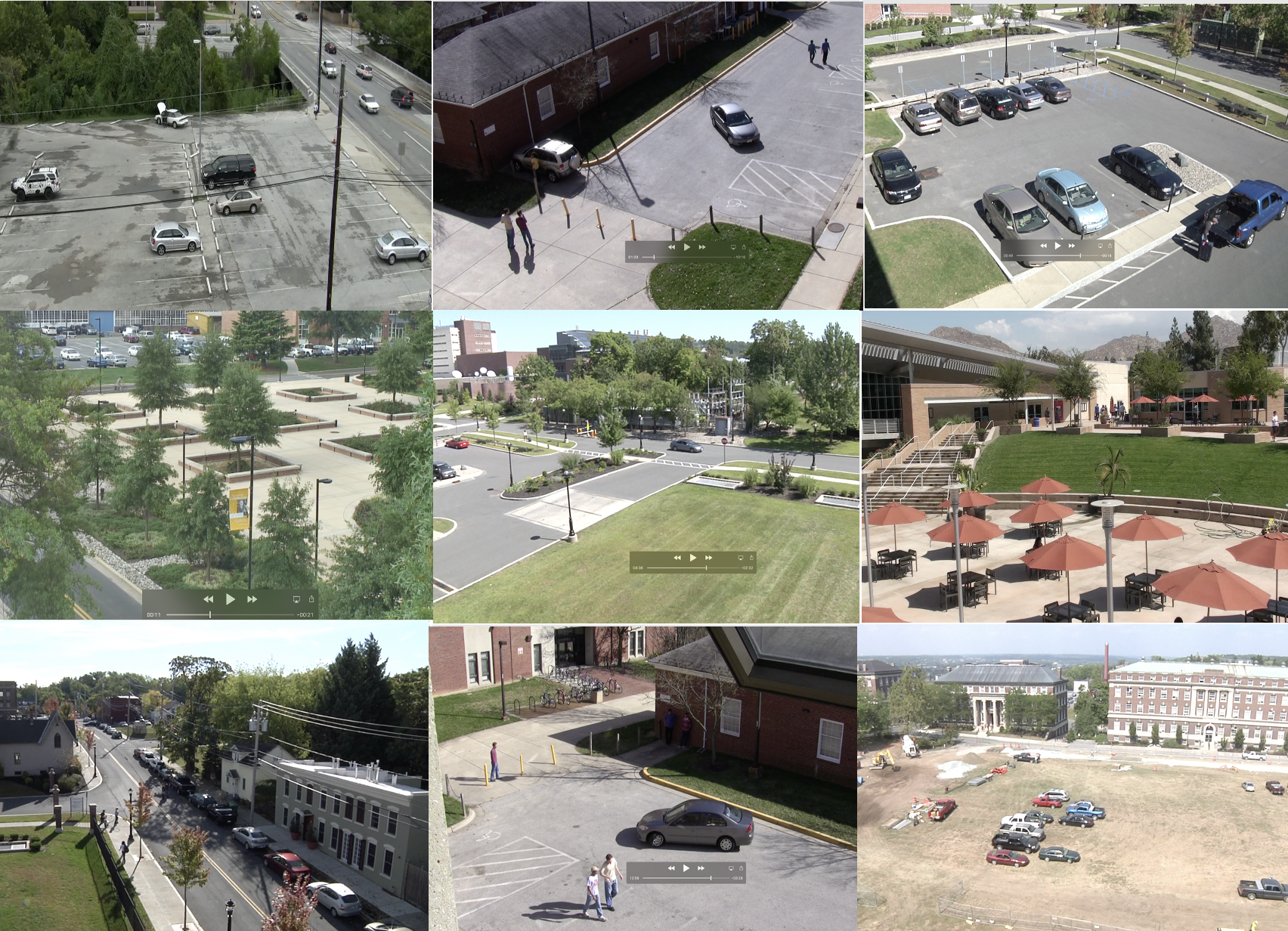}
\caption{Shows the different VIRAT videos montage of few selected video clips.}
\label{fig_actev_dataset}
\end{centering}
\end{figure}

The VIRAT Video Dataset \cite{oh2011large} is a large-scale surveillance video dataset designed to assess the performance of activity detection algorithms in realistic scenes. The dataset was collected outdoor to facilitate both detection of activities and spatio-temporal localization of objects associated with activities from a large continuous video.  The data was collected at different buildings and parking lots at multiple sites distributed throughout the United States. A variety of camera viewpoints and resolutions were included, with different levels of cluttered backgrounds, and activities are performed by many ordinary people. The spatial resolution of the  cameras is either
1920x1080 or 1920x1072. The VIRAT dataset is closely aligned with real-world video surveillance analytics. The 35 activities used for this evaluation could
be broadly categorized as: person/multi-person activity, person object interaction,
vehicle activity, and person vehicle/facility interaction. Figure \ref{fig_actev_dataset} shows the different VIRAT image montages of randomly selected videos. In addition, we have built a larger Multiview Extended Video
with Activities (MEVA) dataset \cite{MEVAdata} which is used for different ActEV Sequestered Data Leaderboard (SDL) competitions \cite{SDL}. The main purpose of the VIRAT data is to stimulate the computer vision community to develop advanced algorithms with improved performance and robustness of human activity detection of multi-camera systems that cover a large area.

\subsection{TRECVID-VTT}
This dataset contains short videos (with durations ranging from 3 seconds to 10 seconds) previously used for the TRECVID VTT task since 2016. In total, there are 9185 videos with captions. Each video has between 2 and 5 captions, which have been written by dedicated annotators. The collection includes 6475 URLs from Twitter Vine and 2710 video files in webm format and Creative Commons License. Those 2710 videos have been extracted from Flickr and the V3C2 dataset (1700 V3C2 videos were used as a testing set this year).

\subsection{Low Altitude Disaster Imagery (LADI)}
The LADI dataset consists of over 20\,000 annotated images, each at least 4 MB in size, and was available as development dataset for the DSDI systems. The images are collected by the Civil Air Patrol from various natural disaster events. The raw images were previously released into the public domain. Two key distinctions are the low altitude (less than 304.8 m (1000 ft)), oblique perspective of the imagery and disaster-related features, which are rarely featured in computer vision benchmarks and datasets. The dataset currently employs a hierarchical labeling scheme of five coarse categories and then more specific annotations for each category. The initial dataset focuses on the Atlantic Hurricane and spring flooding seasons since 2015.

\section{Evaluated Tasks}
\subsection{Ad-hoc Video Search}
The Ad-hoc Video Search (AVS) task was resumed at TRECVID again in 2016 utilizing the Internet Archive Creative Commons (IACC.3) dataset and in 2019 a new Vimeo dataset (V3C1) was adopted instead. The task is trying to model the end user video search use-case, who is looking for segments of video containing people, objects, activities, locations, etc. and combinations of the former. It was coordinated by NIST and by the Laboratoire d'Informatique de Grenoble.

The task for participants was defined as the following: given a standard set of master shot boundaries (about 1 million shots) from the V3C1 test collection and a list of 30 ad-hoc textual queries (see Appendix \ref{appendixA} and \ref{appendixB}), participants were asked to return for each query, at most the top 1000 video clips from the master shot boundary reference set, ranked according to the highest probability of containing the target query. The presence of each query was assumed to be binary, i.e., it was either present or absent in the given standard video shot. 

Judges at NIST followed several rules in evaluating system output. For example, if the query was true for some frame (sequence) within the shot, then it was true for the shot. This is a simplification adopted for the benefits it offered in pooling of results and approximating the basis for calculating recall. In addition, query definitions such as ``contains x" or words to that effect are short for ``contains x to a degree sufficient for x to be recognizable as x by a human". This means among other things that unless explicitly stated, partial visibility or audibility may suffice. Lastly, the fact that a segment contains video of a physical object representing the query target, such as photos, paintings, models, or toy versions of the target (e.g picture of Barack Obama vs Barack Obama himself), was NOT grounds for judging the query to be true for the segment. Containing video of the target within video (such as a television showing the target query) may be grounds for doing so. Three main submission types were accepted:

\begin{itemize}
\item{Fully automatic runs (no human input in the loop): System takes a query as input and produces results without any human intervention.}
\item{Manually-assisted runs: where a human can formulate the initial query based on topic and query interface, not on knowledge of collection or search results. Then system takes the formulated query as input and produces results without further human intervention.}
\item{Relevance-Feedback: System takes the official query as input and produces initial results, then a human judge can assess the top-30 results and input this information as a feedback to the system to produce a final set of results. This feedback loop is strictly permitted only up to 3 iterations.}
\end{itemize}

In general, runs submitted were allowed to choose any of the below four training types:

\begin{itemize}
\item{A - used only IACC training data}
\item{D - used any other training data}
\item{E - used only training data collected automatically using only the official query textual description}
\item{F - used only training data collected automatically using a query built manually from the given official query textual description}
\end{itemize}

The training categories "E" and "F" are motivated by the idea of promoting the development of methods that permit the indexing of concepts in video clips using only data from the web or archives without the need of additional annotations. The training data could for instance consist of images or videos retrieved by a general-purpose search engine (e.g. Google) using only the query definition with only automatic processing of the returned images or videos. 

A new progress subtask was introduced in 2019 with the objective of measuring system progress on a set of 20 fixed topics (Appendix \ref{appendixB}). As a result, 2019 systems were allowed to submit results for 20 common topics (not evaluated in 2019) that will be fixed for three years (2019-2021). This year NIST evaluated progress runs submitted in 2019 and 2020 so that teams can measure their progress against two years (2019-2020) while in 2021 they can measure their progress against three years. In general, the 20 fixed progress topics are divided equally into two sets of 10 topics to be evaluated in 2020 and 2021.

A "Novelty" run type was also allowed to be submitted within the main task. The goal of this run is to encourage systems to submit novel and unique relevant shots not easily discovered by other runs. Finally, teams were allowed to submit an optional explainability parameter with each shot. This was formulated as a keyframe and bounding box to localize the region that supports the query evidence.

\subsubsection{Dataset}
The V3C1 dataset (drawn from a larger V3C video dataset \cite{rossetto2019v3c}) was adopted as a testing dataset. It is composed of 7475 Vimeo videos (1.3 TB, 1000 h) with Creative Commons licenses and mean duration of 8 min. All videos have some metadata available e.g., title, keywords, and description in json files. The dataset has been segmented into 1\,082\,657 short video segments according to the provided master shot boundary files. In addition, keyframes and thumbnails per video segment have been extracted and made available.
For training and development, all previous Internet Archive datasets (IACC.1-3) with about 1\,800 h were made available with their ground truth and xml meta-data files. Throughout this report we do not differentiate between a clip and a shot and thus they may be used interchangeably.

\subsubsection{Evaluation}
Each group was allowed to submit up to 4 prioritized runs per submission type, and per task type (main or progress) and two additional if they were of training type "E" or "F" runs. In addition, one novelty run type was allowed to be submitted within the main task.

In fact, 9 groups submitted a total of 75 runs with 39 main runs and 36 progress runs. Two groups submitted novelty runs. The 39 main runs consisted of 26 fully automatic, and 13 manually-assisted runs, while the progress runs consisted of 24 fully automatic and 12 manually-assisted runs. 

To prepare the results from teams for human judgments, a workflow was adopted to pool results from runs submitted. For each query topic, a top pool was created using 100 \% of clips at ranks 1 to 250 across all submissions after removing duplicates. A second pool was created using a sampling rate at 11.1 \% of clips at ranks 251 to 1000, not already in the top pool, across all submissions and after removing duplicates. Using these two master pools, we divided the clips in them into small pool files with about 1000 clips in each file. Ten human judges (assessors) were presented with the pools - one assessor per topic - and they judged each shot by watching the associated video and listening to the audio then voting if the clip contained the query topic or no. Once the assessor completed judging for a topic, he or she was asked to rejudge all clips submitted by at least 10 runs at ranks 1 to 200 and was voted as false positive by the assessor. This final step was done as a secondary check on the assessors judging work and to give them an opportunity to fix any judgment mistakes. In all, 147\,950 clips were judged while 226\,097 clips fell into the unjudged part of the overall samples. Total hits across the 30 topics reached 22\,859 with 12\,210 hits at submission ranks from 1 to 100, 7969 hits at submission ranks 101 to 250 and 2725 hits at submission ranks between 251 to 1000. Table \ref{avsstats} presents information about the pooling and judging per topic.

\subsubsection{Measures}
Work at Northeastern University \cite{yilmaz06} has resulted in methods for estimating standard system performance measures using relatively small samples of the usual judgment sets so that larger numbers of features can be evaluated using the same amount of judging effort. Tests on past data showed the measure inferred average precision (infAP) to be a good estimator of average precision \cite{tv6overview}. This year mean extended inferred average precision (mean xinfAP) was used which permits sampling density to vary \cite{yilmaz08}. This allowed the evaluation to be more sensitive to clips returned below the lowest rank ($\approx$250) previously pooled and judged. It also allowed adjustment of the sampling density to be greater among the highest ranked items that contribute more average precision than those ranked lower.
The {\em sample\_eval} software \footnote{http://www-nlpir.nist.gov/projects/trecvid/\\trecvid.tools/sample\_eval/}, a tool implementing xinfAP, was used to calculate inferred recall, inferred precision, inferred average precision, etc., for each result, given the sampling plan and a submitted run. Since all runs provided results for all evaluated topics, runs can be compared in terms of the mean inferred average precision across all evaluated query topics. 

\begin{table*}[t]
\caption{Ad-hoc search pooling and judging statistics}
\label{avsstats} 
  \vspace{0.5cm}
  \centering{
   \small{
    \begin{tabular}{|c|c|c|c|c|c|c|c|}
      \hline 

\parbox{1.3cm}{Topic \\ number} & 
\parbox{1.5cm}{Total \\ submitted} & 
\parbox{1.5cm}{Unique \\ submitted} & 
\parbox{1.0cm}{total \\ that \\ were \\ unique\\ \%} & 
\parbox{1.2cm}{Number \\ judged} & 
\parbox{1.0cm}{unique \\ that \\ were \\ judged\\ \%} & 
\parbox{1.2cm}{Number \\ relevant} & 
\parbox{1.2cm}{judged \\ that \\ were \\ relevant\\ \%}

 \\ \hline
 
1591 & 72692 & 64555 & 88.81 & 6115 & 9.47 & 705 & 11.53\\ \hline 
1593 & 73856 & 70481 & 95.43 & 7705 & 10.93 & 345 & 4.48\\ \hline 
1594 & 72936 & 65249 & 89.46 & 6043 & 9.26 & 547 & 9.05\\ \hline 
1596 & 73996 & 67095 & 90.67 & 5321 & 7.93 & 57 & 1.07\\ \hline 
1597 & 73996 & 66281 & 89.57 & 5355 & 8.08 & 213 & 3.98\\ \hline 
1598 & 73936 & 62872 & 85.04 & 5675 & 9.03 & 230 & 4.05\\ \hline 
1602 & 73996 & 68596 & 92.70 & 6238 & 9.09 & 1585 & 25.41\\ \hline 
1604 & 73996 & 64148 & 86.69 & 6495 & 10.13 & 905 & 13.93\\ \hline 
1606 & 73996 & 61256 & 82.78 & 9626 & 15.71 & 277 & 2.88\\ \hline 
1610 & 72942 & 64411 & 88.30 & 7072 & 10.98 & 953 & 13.48\\ \hline 
1641 & 39000 & 32867 & 84.27 & 3416 & 10.39 & 723 & 21.17\\ \hline 
1642 & 39000 & 31640 & 81.13 & 2602 & 8.22 & 1042 & 40.05\\ \hline 
1643 & 39000 & 34885 & 89.45 & 5287 & 15.16 & 302 & 5.71\\ \hline 
1644 & 39000 & 33874 & 86.86 & 4041 & 11.93 & 1152 & 28.51\\ \hline 
1645 & 37502 & 30863 & 82.30 & 4344 & 14.08 & 1339 & 30.82\\ \hline 
1646 & 38734 & 33868 & 87.44 & 4319 & 12.75 & 461 & 10.67\\ \hline 
1647 & 39000 & 36846 & 94.48 & 5094 & 13.83 & 1678 & 32.94\\ \hline 
1648 & 39000 & 32881 & 84.31 & 4331 & 13.17 & 826 & 19.07\\ \hline 
1649 & 39000 & 30802 & 78.98 & 3026 & 9.82 & 1804 & 59.62\\ \hline 
1650 & 39000 & 33807 & 86.68 & 3879 & 11.47 & 322 & 8.30\\ \hline 
1651 & 39000 & 34875 & 89.42 & 3772 & 10.82 & 518 & 13.73\\ \hline 
1652 & 38592 & 31836 & 82.49 & 3363 & 10.56 & 597 & 17.75\\ \hline 
1653 & 39000 & 33888 & 86.89 & 4178 & 12.33 & 972 & 23.26\\ \hline 
1654 & 37502 & 36810 & 98.15 & 4778 & 12.98 & 529 & 11.07\\ \hline 
1655 & 38756 & 36879 & 95.16 & 5139 & 13.93 & 569 & 11.07\\ \hline 
1656 & 39000 & 31773 & 81.47 & 5158 & 16.23 & 1234 & 23.92\\ \hline 
1657 & 39000 & 31930 & 81.87 & 5535 & 17.33 & 837 & 15.12\\ \hline 
1658 & 39000 & 35877 & 91.99 & 5011 & 13.97 & 832 & 16.60\\ \hline 
1659 & 39000 & 31813 & 81.57 & 2155 & 6.77 & 441 & 20.46\\ \hline 
1660 & 39000 & 32758 & 83.99 & 2877 & 8.78 & 900 & 31.28\\ \hline

    \end{tabular}
   } 
  } 
\end{table*}

\subsubsection{Ad-hoc Results}

\begin{figure}[hbtp]
\begin{center}
\includegraphics[height=2.5in,width=3.0in,angle=0]{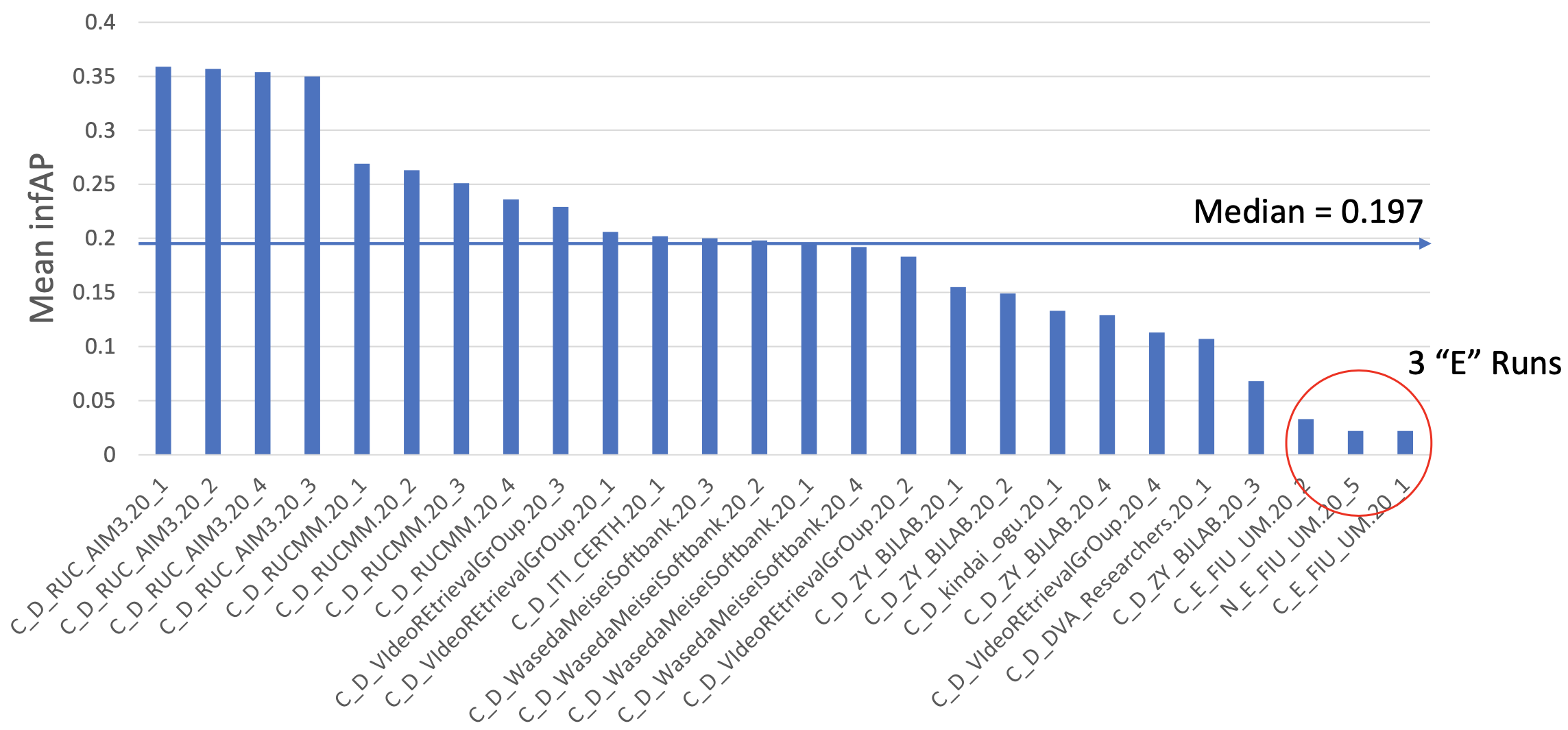}
\caption{AVS: 26 Automatic Runs across 20 Main Queries}
\label{avs.f.all}
\end{center}
\end{figure}

\begin{figure}[hbtp]
\begin{center}
\includegraphics[height=2.5in,width=3.0in,angle=0]{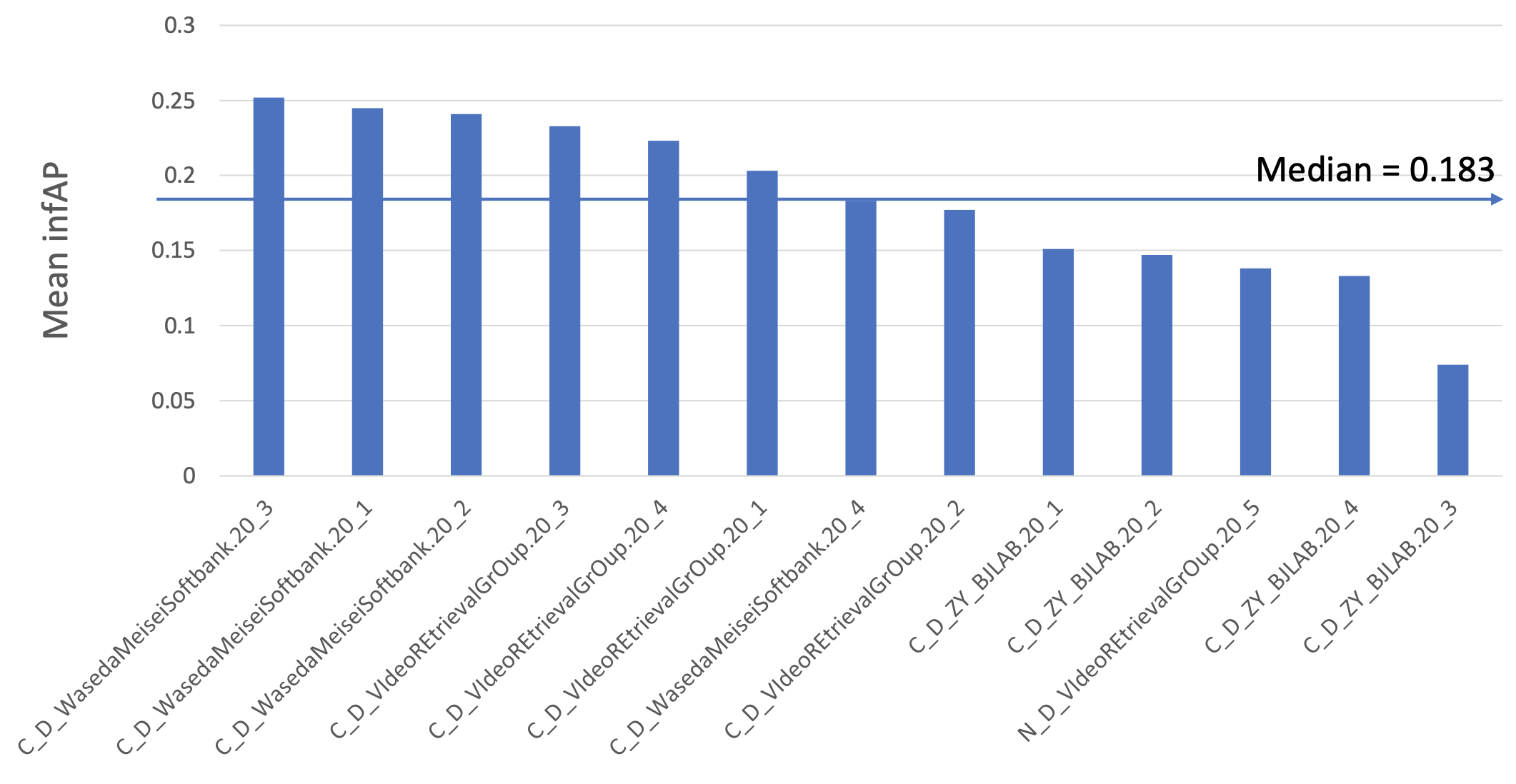}
\caption{AVS: 13 Manually-Assisted Runs across 20 Main queries}
\label{avs.m.all}
\end{center}
\end{figure}

Figures \ref{avs.f.all} and \ref{avs.m.all} show the results of all the 26 fully automatic runs and 13 manually-assisted submissions respectively. 

This is the second year for the ad-hoc task to work with the V3C1 dataset. As tested queries in the main task are different each year, we can not directly compare the performance the same way we do in the progress subtask. However, we can see that the best automatic team runs outperformed the top manually-assisted runs. Also, we should note that the only 3 submitted E runs performed the lowest among all automatic runs. This shows that collecting automatic training data is still very hard and challenging to systems.

We should also note here that while the majority of runs were of type "D", no runs using category "F" were submitted. Also, while the evaluation supported a relevance feedback run types, this year no submissions were received under this category.

To test if there were significant differences between the runs submitted, we applied a randomization test \cite{manly97} on the top 10 runs for manually-assisted and automatic run submissions using a significance threshold of p$<$0.05. For automatic runs, the analysis showed that there were no significant differences between runs 1,2 and 4 of team RUC\_AIM3, while run 2 was significantly better than run 3. For team RUCMM, there were no significant differences between their 4 runs. Finally, for the team VIdeoREtrievalGrOup, their run 3 is better than run 1. For manually-assisted runs, the analysis showed that there was no significant differences between runs 1,2 and 3 for team WasedaMeiseiSoftbank, and they are all better than run 4. For team VIdeoREtrievalGrOup, run 3 is better than runs 1 and 2. And finally, run 1 of team ZY\_BJLAB is better than run 2.

\begin{figure}[hbtp]
\begin{center}
\includegraphics[height=2.5in,width=3.0in,angle=0]{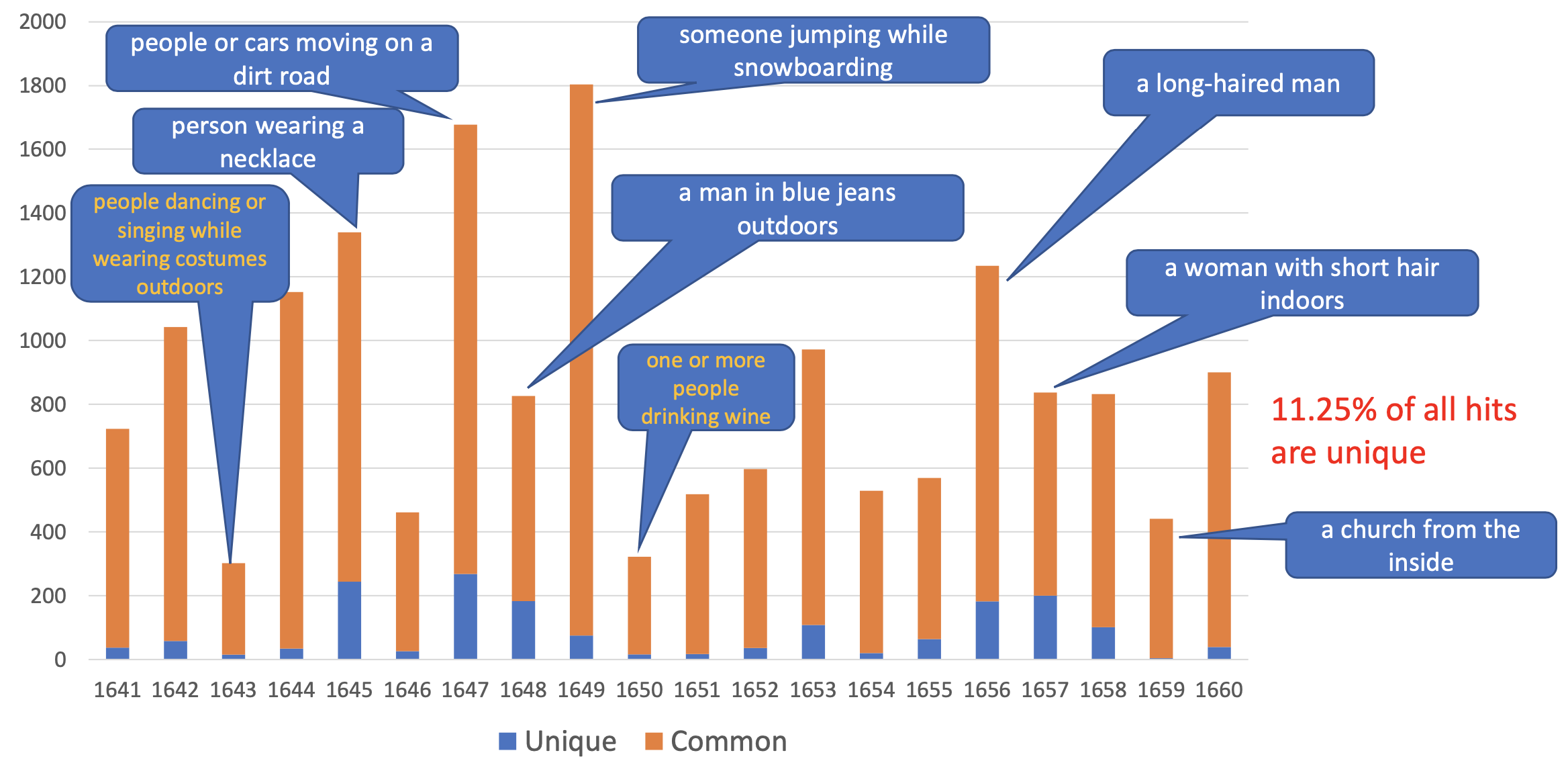}
\caption{AVS: Unique vs overlapping results in main task}
\label{avs.unique.overlapped}
\end{center}
\end{figure}

Figure \ref{avs.unique.overlapped} shows for each topic the number of relevant and unique shots submitted by all teams combined (blue color). On the other hand, the orange bars show the total non-unique true shots submitted by at least 2 or more teams. The four topics: 1647, 1645, 1657, and 1656 achieved the most unique hits overall while also reporting a high number of hits overall, while the 3 topics: 1643, 1650, and 1659 reported the lowest unique hits and lowest hits overall. In general, topics that reported a high number of hits consisted of high number of unique as well as non-unique hits, while topics that reported low number of hits mainly only consisted of non-unique hits representing the difficulty of the query. 

\begin{figure}[htbp]
\begin{center}
\includegraphics[height=2.0in,width=3in,angle=0]{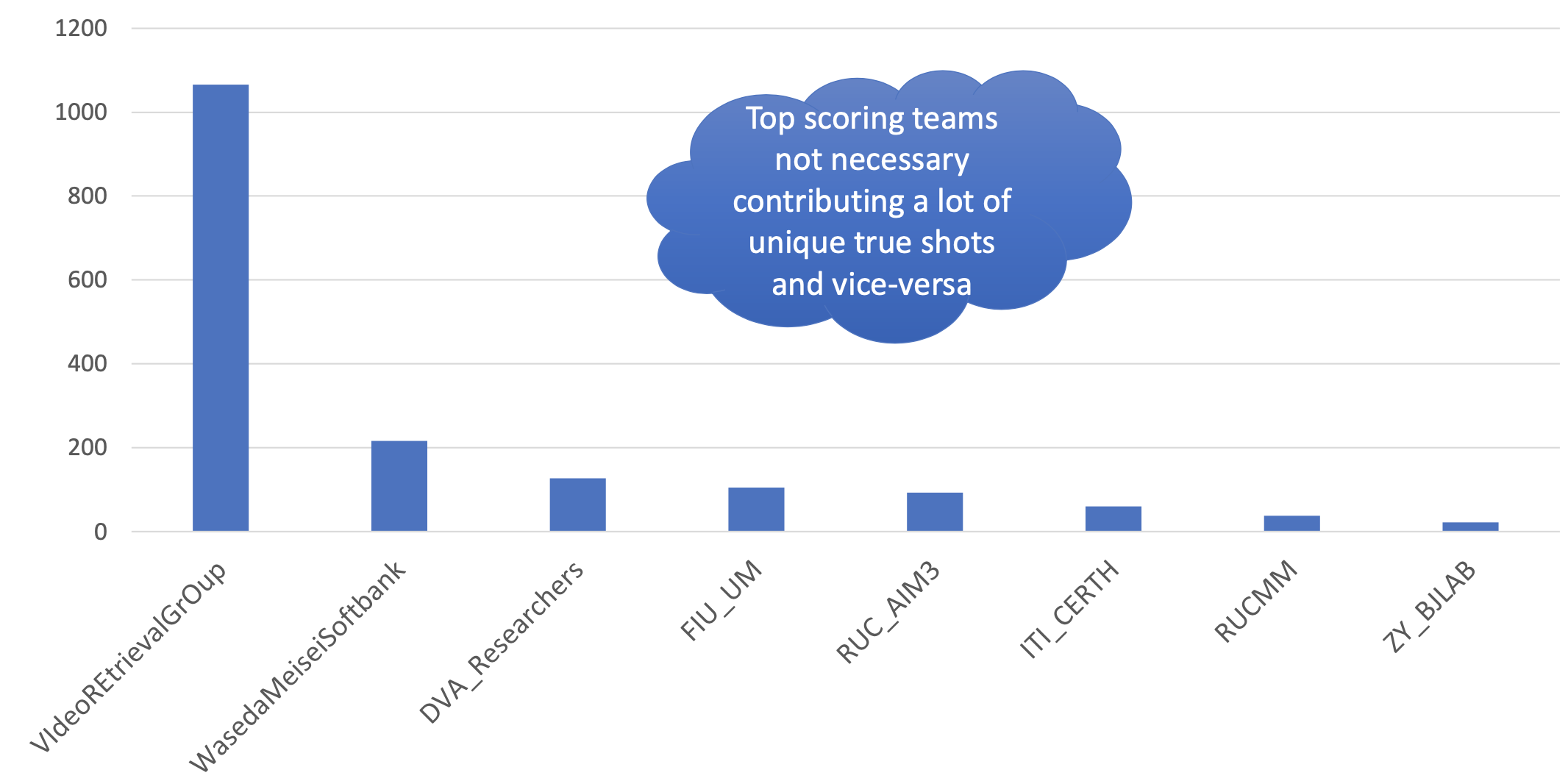}
\caption{AVS: 1727 Unique shots contributed by teams in main task}
\label{avs.unique.byteam}
\end{center}
\end{figure}

Figure \ref{avs.unique.byteam} shows the number of unique clips found by the different participating teams. From this figure and the overall scores in figures \ref{avs.f.all} and \ref{avs.m.all} it can be shown that there is no clear relation between teams who found the most unique shots and their total performance. Many of the top-performing teams did not contribute a lot of unique relevant shots. While the top two teams producing the largest number of unique relevant shots are not among the top performing teams with automatic runs. This observation is consistent with that of the past few years. 

\begin{figure}[htbp]
\begin{center}
\includegraphics[height=2.0in,width=3.0in,angle=0]{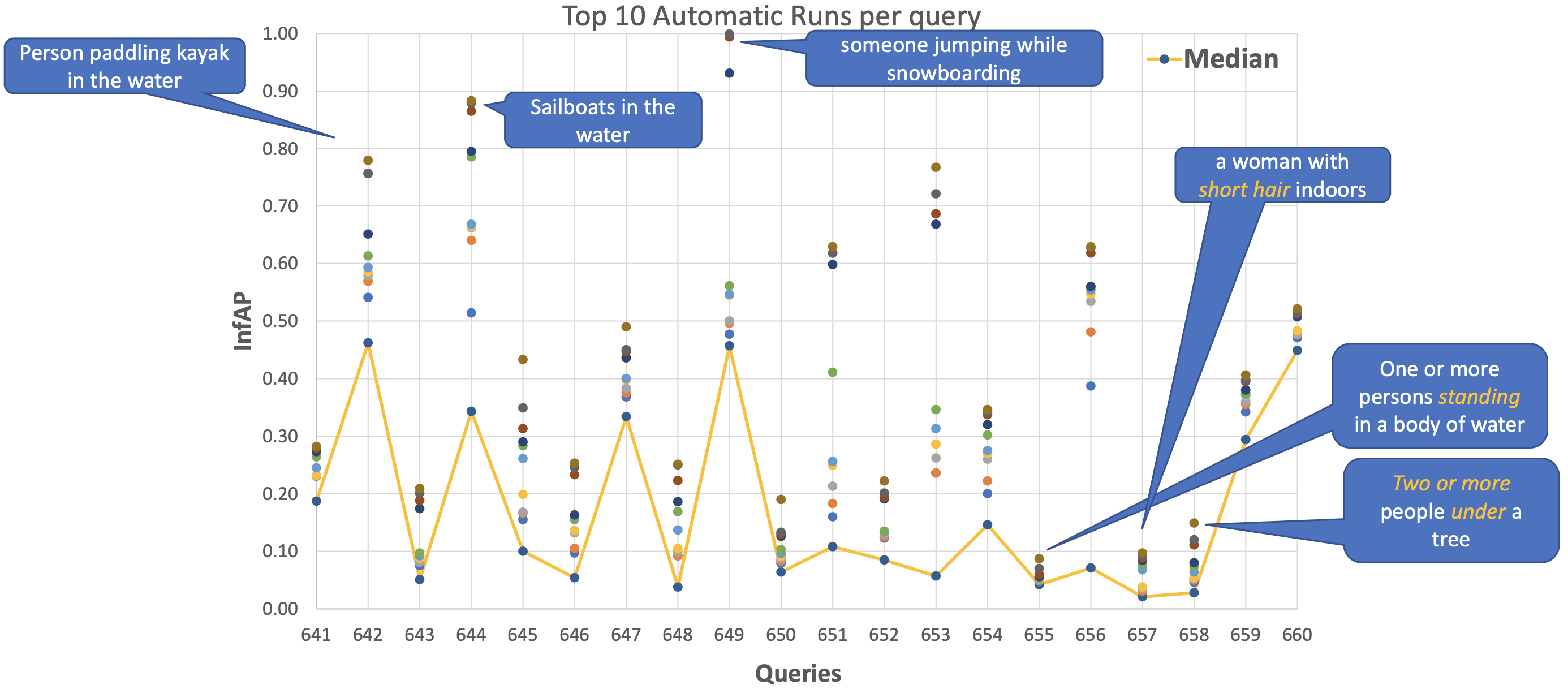}
\caption{AVS: Top 10 runs (xinfAP) per query (fully automatic)}
\label{avs.top10.f}
\end{center}
\end{figure}

\begin{figure}[htbp]
\begin{center}
\includegraphics[height=2.0in,width=3.0in,angle=0]{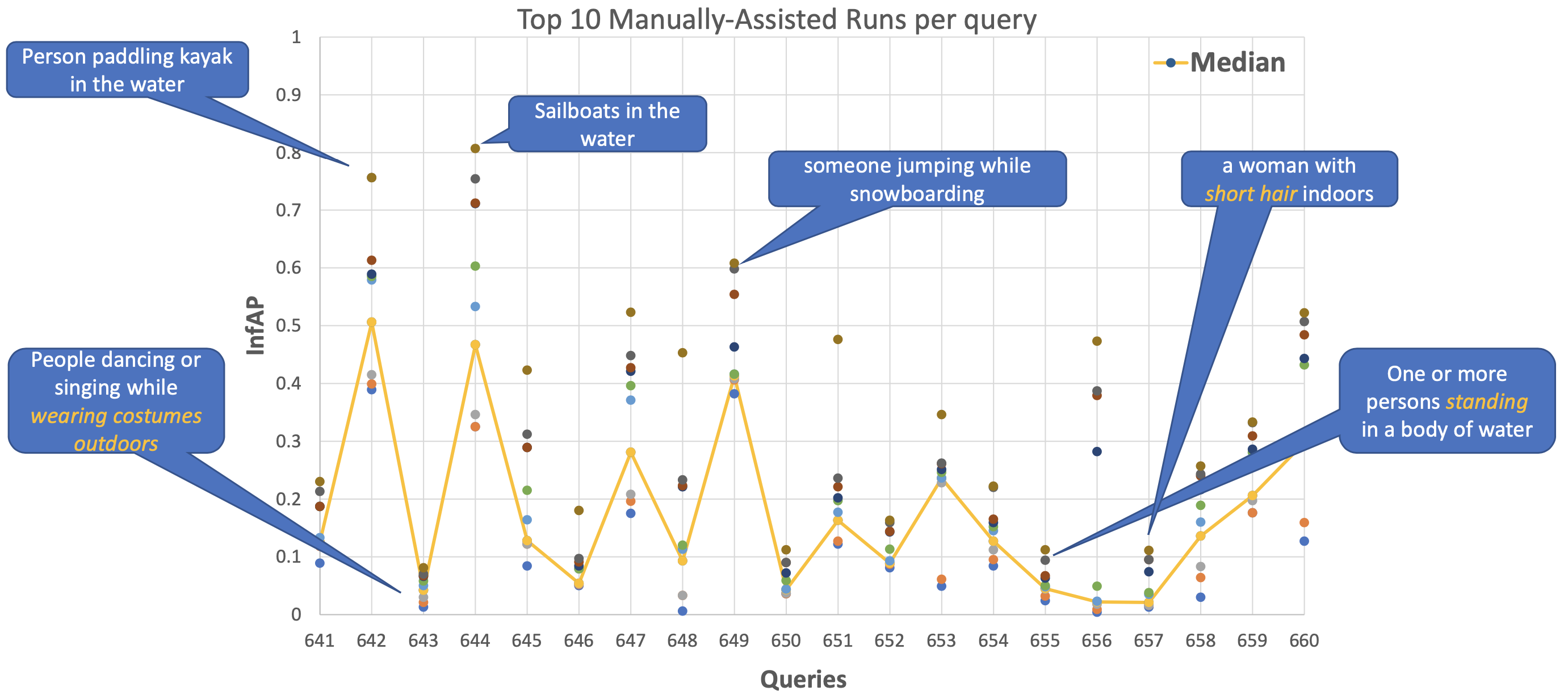}
\caption{AVS: Top 10 runs (xinfAP) per query (manually assisted)}
\label{avs.top10.m}
\end{center}
\end{figure}

Figures \ref{avs.top10.f} and \ref{avs.top10.m} show the performance of the top 10 teams across the 20 main queries. Note that each series in this plot represents a rank (from 1 to 10) of the scores, but all scores at a given rank do not necessarily belong to a specific team.
A team's scores can rank differently across the 20 queries. Some samples of top and bottom performing queries are highlighted with the query text.

A main theme among the top-performing queries is their composition of more common visual concepts (e.g kayak, sailboat, snowboard, skydiving, etc) compared to the bottom ones which require more temporal analysis for some activities and combination of one or more facets of who,what and where/when (e.g person standing in body of water, two or more people under a tree, people dancing or singing while wearing costumes outdoors). 

In general, there is a noticeable spread in score ranges among the top 10 runs, especially with high-performing topics which may indicate the variation in the performance of the used techniques and that there is still room for further improvement. However for topics not performing well, usually all top 10 runs are condensed together with low spread between their scores. In addition, from the two figures, we can see that automatic and manually-assisted systems had comparable performance across all queries.

A new novelty run type was introduced last year to encourage submitting unique (hard to find) relevant shots. Systems were asked to label their runs as either of novelty type or common type runs. A new novelty metric was designed to score runs based on how good they are in detecting unique relevant shots. A weight was given to each topic and shot pairs such as follows:

\[TopicX\_ShotY_{weight} (x) = 1 - \frac{N}{M}\]

\noindent where N is the number of times Shot Y was retrieved for topic X by any run submission, and M is the number of total runs submitted by all teams. For instance, a unique relevant shot weight will be close to 1.0 while a shot submitted by all runs will be assigned a weight of 0.

\begin{figure}[htbp]
\begin{center}
\includegraphics[height=2.5in,width=3.0in,angle=0]{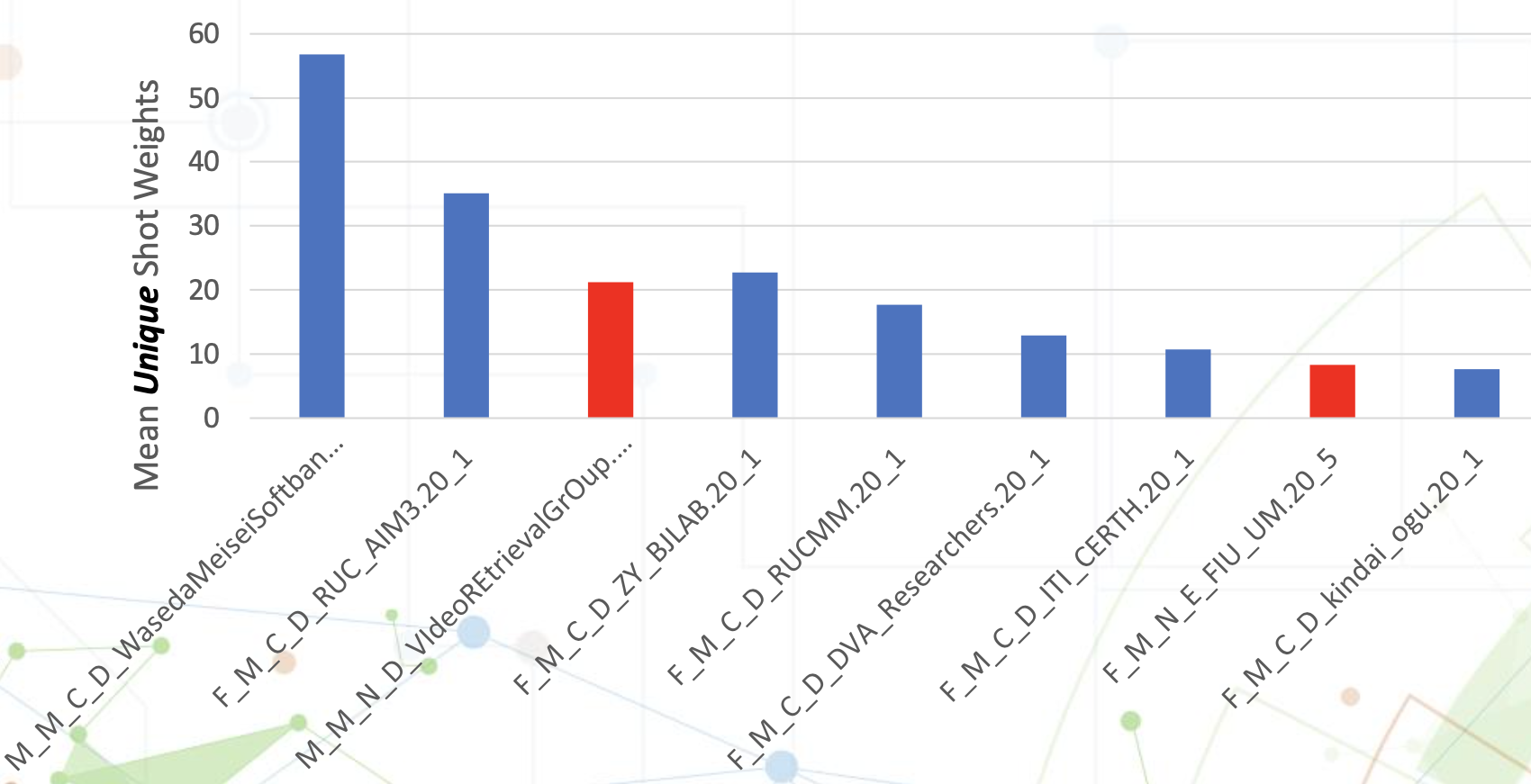}
\caption{AVS: Novelty runs vs best common run from each team}
\label{avs.novelty.scores}
\end{center}
\end{figure}

For Run R and for all topics, we calculate the summation S of all unique shot weights only and the final novelty metric score is the mean score across all evaluated 20 topics. Figure \ref{avs.novelty.scores} shows the novelty metric scores. The red bars indicate the submitted novelty runs. 

We should note here that in running this experiment, for a team that submitted a novelty run, we removed all its other common runs submitted. The reason for doing this was the fact that usually for a given team there would be many overlapping shots within all its submitted runs. For other teams who did not submit novelty runs, we chose the best (top scoring) run for each team for comparison purposes. As shown in the figure, novelty runs did not achieve the best score compared to other top single common runs. As more teams submit novelty runs, it will become more reliable to measure and compare novelty approaches.

\begin{figure}[htbp]
\begin{center}
\includegraphics[height=2.0in,width=3.0in,angle=0]{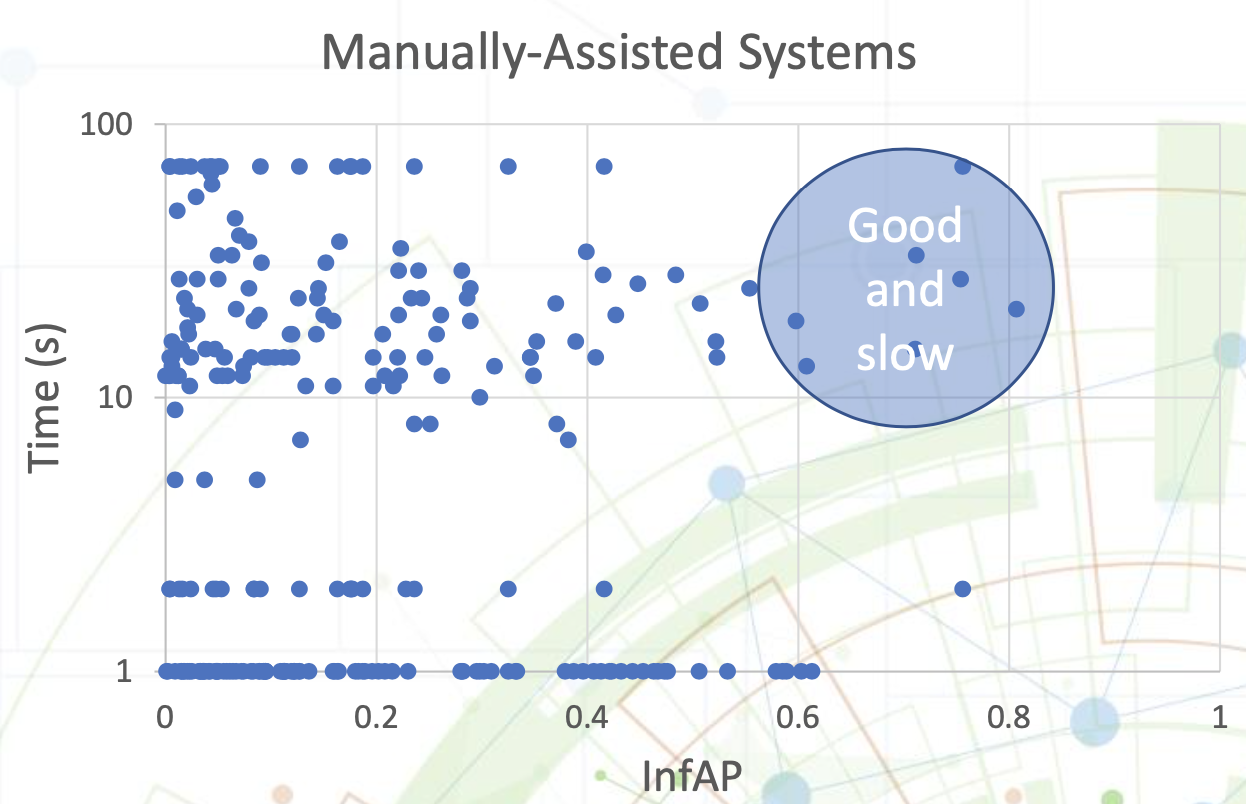}
\caption{AVS: Processing time vs Scores (fully automatic)}
\label{avs.f.time.score}
\end{center}
\end{figure}

\begin{figure}[htbp]
\begin{center}
\includegraphics[height=2.0in,width=3.0in,angle=0]{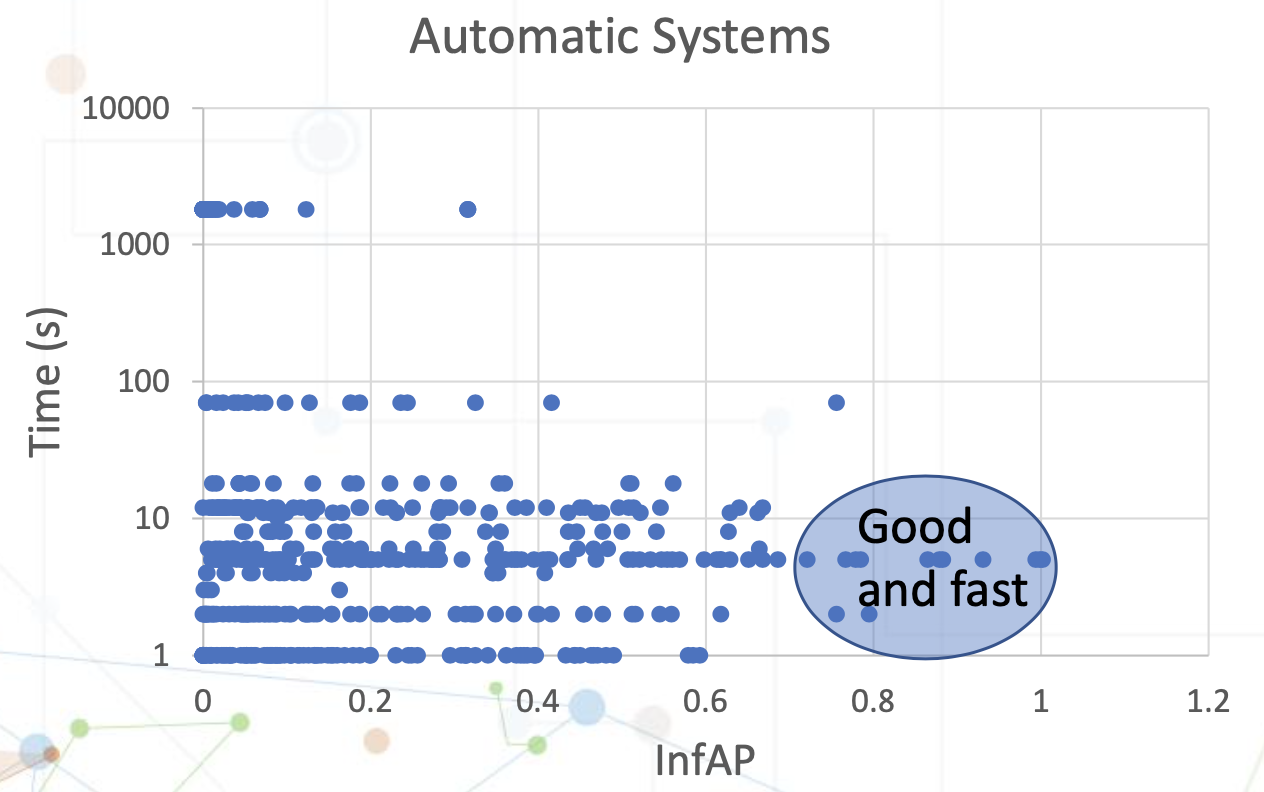}
\caption{AVS: Processing time vs Scores (Manually assisted)}
\label{avs.m.time.score}
\end{center}
\end{figure}

Among the submission requirements, we asked teams to submit the processing time that was consumed to return the result sets for each query. Figures \ref{avs.f.time.score} and \ref{avs.m.time.score} plot the reported processing times vs the InfAP scores among all run queries for automatic and manually-assisted runs respectively. It can be seen that spending more time did not necessarily help in many cases and few queries achieved high scores in less time. There is more work to be done to make systems efficient and effective at the same time. In general, most automatic systems reported processing time below 10 s. While most manually-assisted systems reported processing times above 10 s. 

\begin{figure}[htbp]
\begin{center}
\includegraphics[height=2.0in,width=3.0in,angle=0]{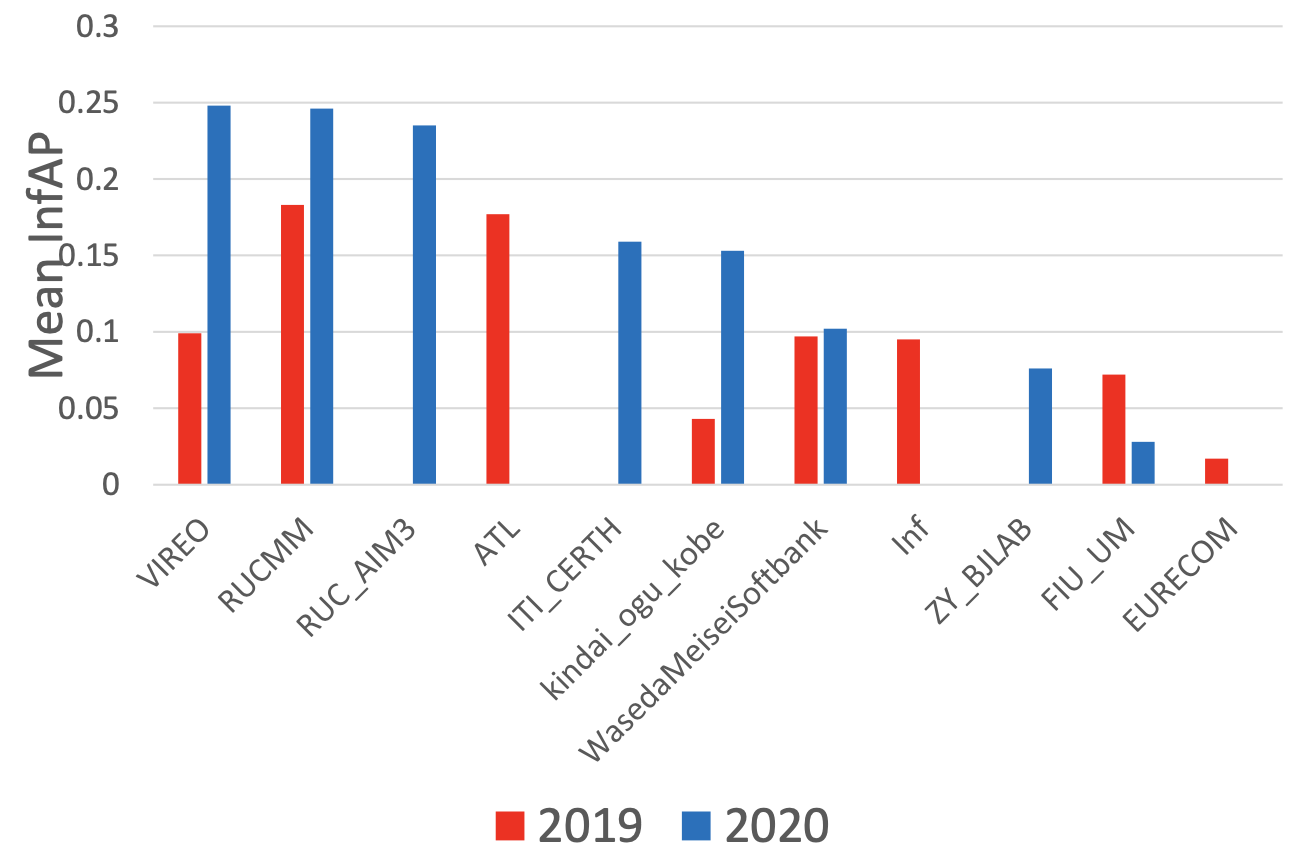}
\caption{AVS: Max performance per team on 10 progress queries (automatic systems)}
\label{avs.progress.f}
\end{center}
\end{figure}

\begin{figure}[htbp]
\begin{center}
\includegraphics[height=2.0in,width=3.0in,angle=0]{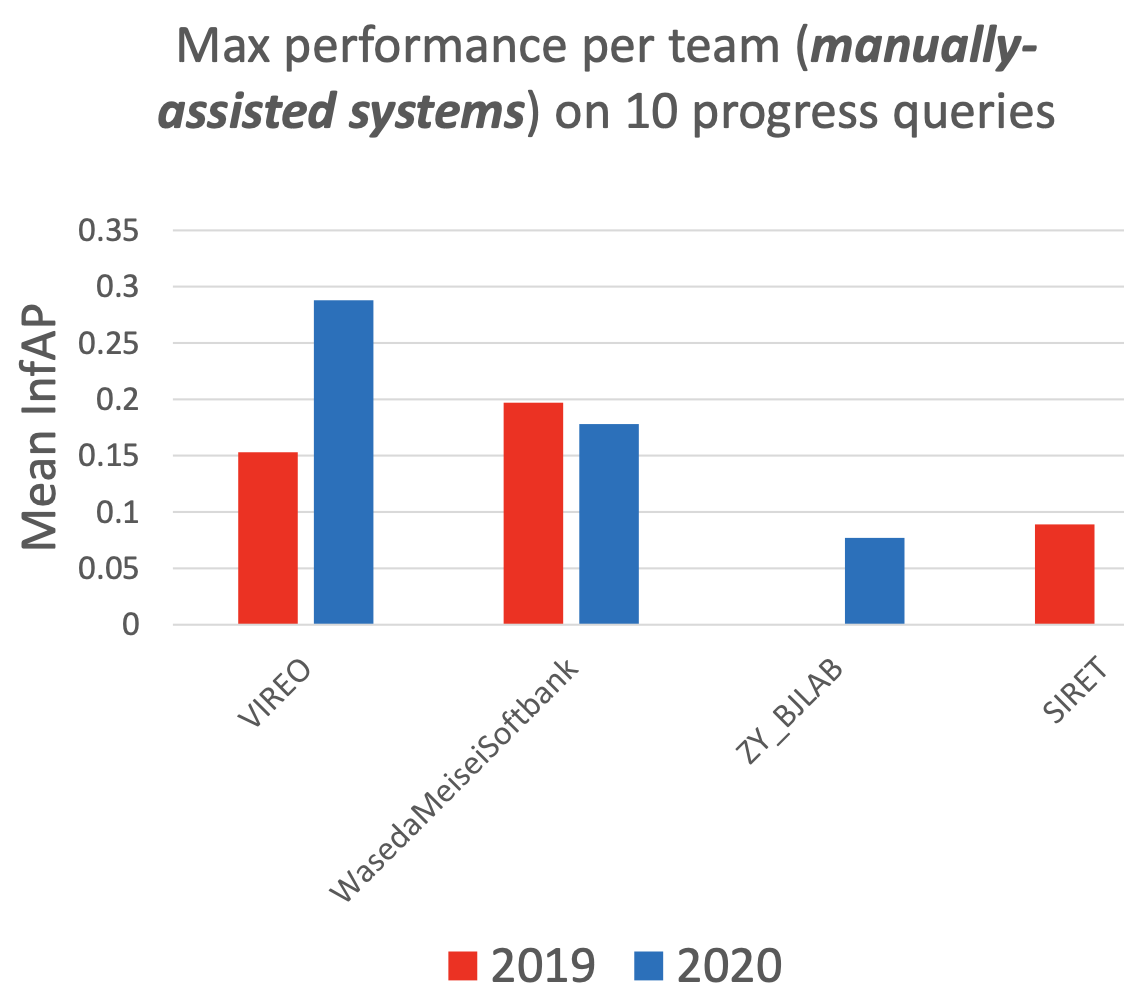}
\caption{AVS: Max performance per team on 10 progress queries (manually-assisted systems)}
\label{avs.progress.m}
\end{center}
\end{figure}

The progress task results are shown in figures \ref{avs.progress.f} and \ref{avs.progress.m} for automatic and manually-assisted systems respectively. Comparing the best run in 2019 vs 2020 for each team, we can see that most systems achieved better performance in 2020. There are few teams who participated in just one of the two years. One team in each of both categories had a better 2019 system (team FIU\_UM in automatic results, and team WasedaMeiseiSoftbank in manually-assisted results). Some of the other teams doubled or more their performance which is very promising and significant progress.

To analyze in general which topics were the easiest and most difficult we sorted topics by the number of runs that scored xInfAP $>$= 0.5 for any given topic and assumed that those were the easiest topics, while topics with xInfAP $<$ 0.5 were assumed hard topics. 
From this analysis, it can be concluded that the top 5 hard topics were: \say{Find shots of people dancing or singing while wearing costumes outdoors}, \say{Find shots of a man in blue jeans outdoors}, \say{Find shots of a church from the inside}, \say{Find shots of a person wearing a necklace}, and \say{Find shots of a little boy smiling}. On the other hand, the top 5 easiest topics were: \say{Find shots of a person paddling kayak in the water}, \say{Find shots of sailboats in the water}, \say{Find shots of someone jumping while snowboarding}, \say{Find shots of a long-haired man}, and \say{Find shots of train tracks during the daytime}.

It can be concluded that hard topics are associated more with conditions that must be satisfied in the retrieved shots (e.g. \say{people in costumes singing/dancing outdoors}, or the type of person in certain condition as in \say{man in blue jeans outdoors}, or \say{little boy smiling}) compared to easily identifiable visual concepts within the easy topics.

\begin{figure}[htbp]
\begin{center}
\includegraphics[height=3.5in,width=2.0in,angle=0]{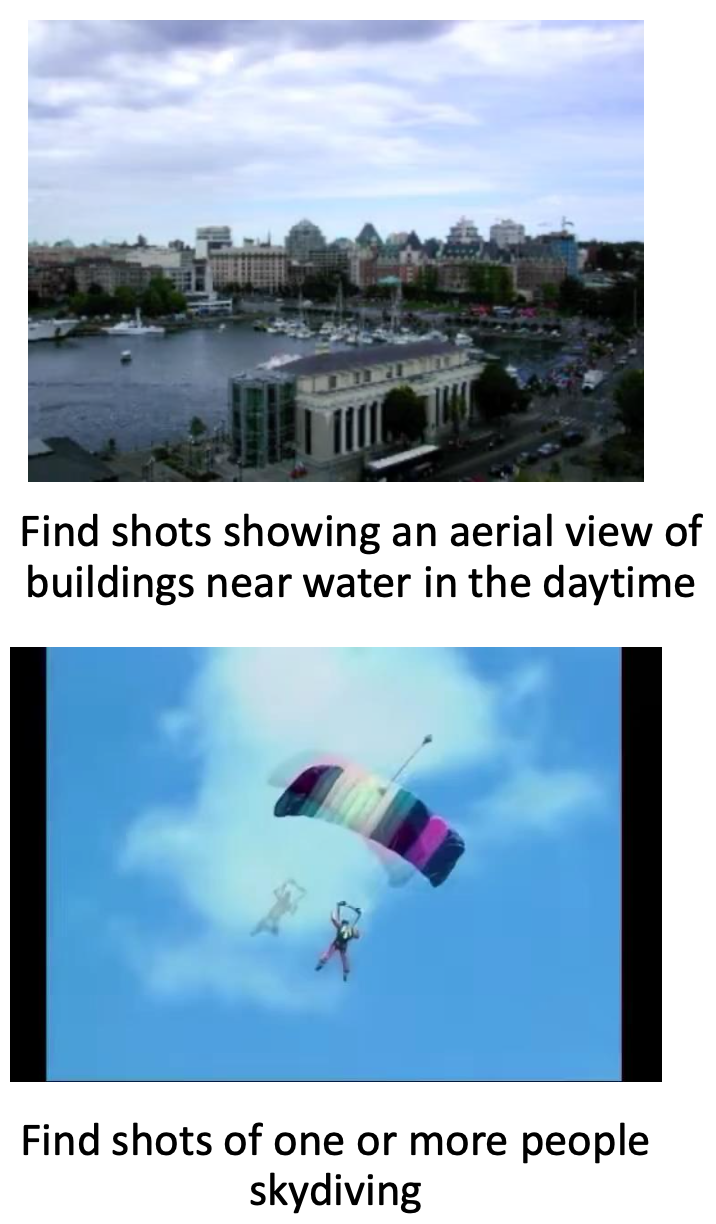}
\caption{AVS: Samples of frequent false positive results (1)}
\label{avs.fp1}
\end{center}
\end{figure}

\begin{figure}[htbp]
\begin{center}
\includegraphics[height=3.5in,width=2.0in,angle=0]{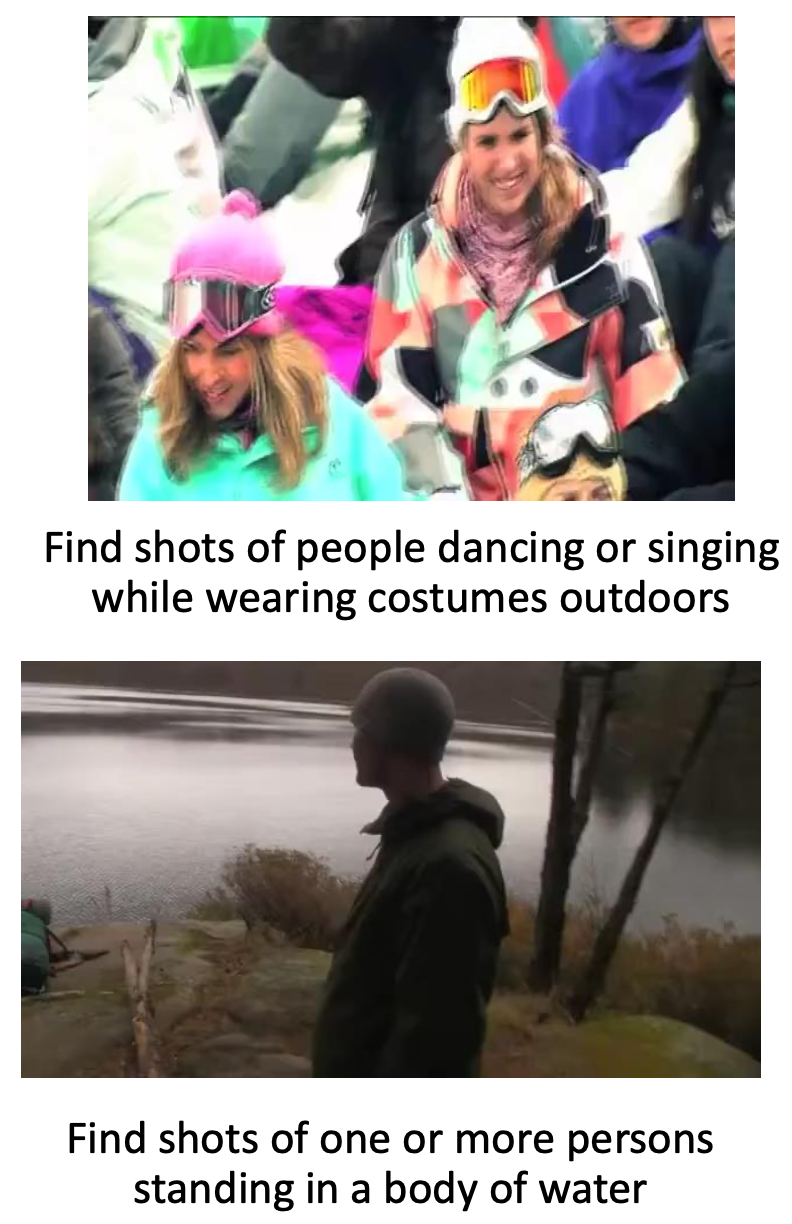}
\caption{AVS: Samples of frequent false positive results (2)}
\label{avs.fp2}
\end{center}
\end{figure}

\begin{figure}[htbp]
\begin{center}
\includegraphics[height=3.5in,width=2.0in,angle=0]{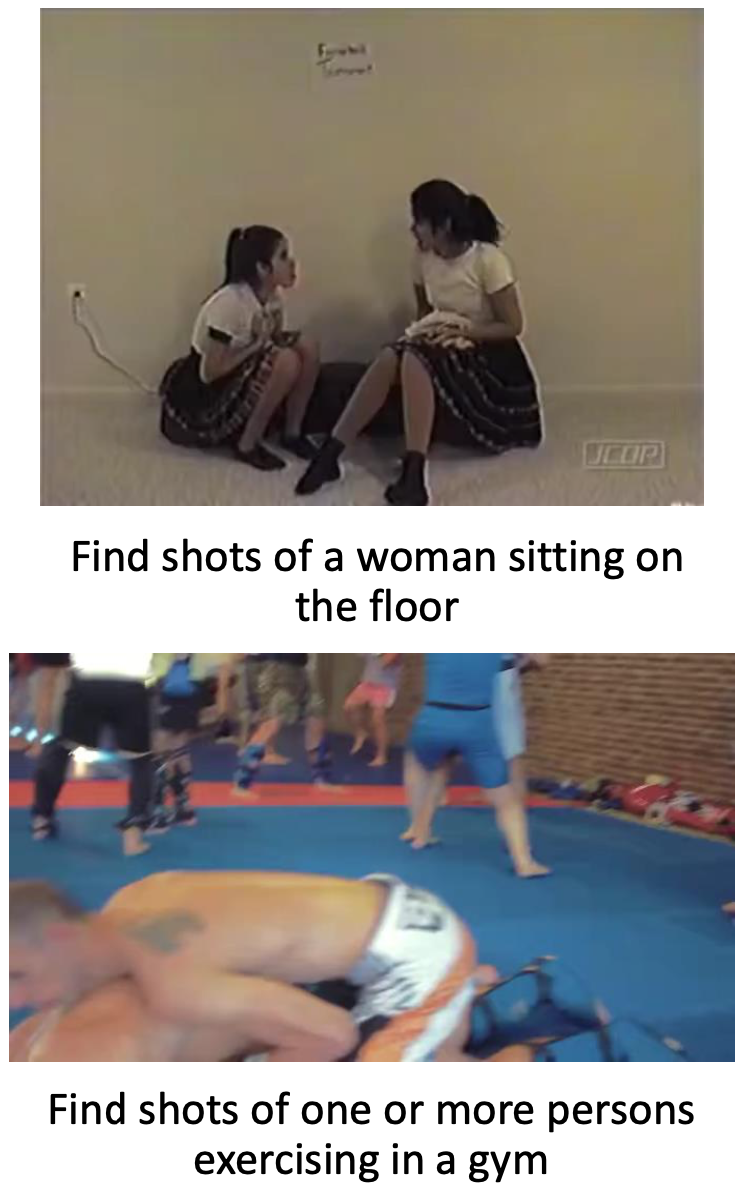}
\caption{AVS: Samples of frequent false positive results (3)}
\label{avs.fp3}
\end{center}
\end{figure}

Sample results of frequently submitted false positive shots are demonstrated\footnote{All figures are in the public domain and permissible under RPO \#ITL-17-0025} in Figures \ref{avs.fp1} to \ref{avs.fp3} .

\subsubsection{Ad-hoc Observations and Conclusions}

Compared to the semantic indexing task that was conducted to detect single concepts (e.g., airplane, animal, bridge) from 2010 to 2015 it can be seen from this year's results that the ad-hoc task is still very hard and systems still have a lot of room to research methods that can deal with unpredictable queries composed of one or more concepts including their interactions.

In 2018 we concluded 1-cycle of three years of Ad-hoc task using the Internet Archive (IACC.3) dataset \cite{2016trecvidover}. Last year, a new dataset, Vimeo Creative Commons Collection (V3C1), was introduced and adopted for testing at least for a 3 year cycle (2019-2021). NIST Developed a set of 90 queries to be used for 3 years including a progress subtask.

To summarize major observations in 2020 we can see that overall, team participation and task completion rates are stable. Most submitted runs were of training type “D”, only 3 runs of type "E" were submitted, and no relevance feedback submissions were received. In addition, no run was submitted with the optional explainability results. The majority of 2020 systems performed higher than their 2019 versions in the progress subtask. Fully automatic and manually-assisted performance are almost similar. Among high scoring topics, there is more room for improvement among systems. Among low scoring topics, most systems scores fall in the same narrow range.
Top scoring teams did not necessarily report unique relevant shots (thus they are good in ranking relevant shots). Queries with unusual combinations of facets constitute the most difficult queries. Most systems are computationally inefficient. Few systems are efficient and effective at retrieving fast and accurate results. Finally, the task still remains challenging.

As a general high-level system overview, we can see that there are two main competing approaches among participating teams: “concept-based banks” and “visual-textual embedding spaces”. There is a clear trend towards the embedding approaches due to their performance. Concept-based banks are often used as a complement to embedding approaches. Training data for semantic spaces included MSR-VTT (MSR Video to Text) and TRECVID VTT, TGIF (Tumblr GIF), IACC.3, Flickr8k, Flickr30k, MS COCO (MS Common Objects in Context), and Conceptual Captions, and VATEX (Video And TEXt). In general teams such as RUC\_AIM3 used a two-branch framework with global Visual-Semantic Embedding (VSE++) and fine-grain matching with Hierarchical Graph Reasoning (HGR) \cite{rucaim3}, while team RUCMM applied a dual encoding network with Word to Visual Vector (W2VV++) and BERT as text encoder plus Sentence Encoder Assembly (SEA) \cite{rucmm} using multi-space multi-loss learning. The VIREO team used a dual-task model that learns feature embedding and concept decoding simultaneously for automatic runs, while for manually-assisted runs, they assigned a user to filter the query from unrelated concepts. Team Waseda used a VSE++ based approach for their automatic system and a concept-based method similar to previous years' concept-bank approach and fused it with VSE \cite{waseda}. Team ITI-CERTH used an attention-based cross-modal deep network inspired by the dual encoding approach \cite{iticerth}, while team ZY\_BJLAB used a multi-modal video representation from collaborative experts to retrieve video shots.

For detailed information about the approaches and results for individual teams, we refer the reader to the reports \cite{tv20pubs} in the online workshop notebook proceedings.

\subsection{Instance search}

\begin{table*}[t]
\caption{Instance search pooling and judging statistics}
\label{searchstats} 
  \vspace{0.5cm}
  \centering{
   \small{
    \begin{tabular}{|c|c|c|c|c|c|c|c|c|}
      \hline 

\parbox{1.3cm}{Topic \\ number} & 
\parbox{1.5cm}{Total \\ submitted} & 
\parbox{1.5cm}{Unique \\ submitted} & 
\parbox{1.0cm}{total \\ that \\ were \\ unique\\ \%} & 
\parbox{1.0cm}{Max.\\ result\\ depth \\ pooled\\} & 
\parbox{1.2cm}{Number \\ judged} & 
\parbox{1.0cm}{unique \\ that \\ were \\ judged\\ \%} & 
\parbox{1.2cm}{Number \\ relevant} & 
\parbox{1.2cm}{judged \\ that \\ were \\ relevant\\ \%}

 \\ \hline
 
9279 & 26880 & 25315 & 94.18 & 400 & 4260 & 16.83 & 533 & 12.51\\ \hline 
9282 & 27102 & 23137 & 85.37 & 320 & 2897 & 12.52 & 105 & 3.62\\ \hline 
9284 & 27999 & 24769 & 88.46 & 200 & 2068 & 8.35 & 75 & 3.63\\ \hline 
9285 & 27999 & 21807 & 77.88 & 280 & 2783 & 12.76 & 110 & 3.95\\ \hline 
9287 & 27378 & 21046 & 76.87 & 260 & 2593 & 12.32 & 264 & 10.18\\ \hline 
9290 & 26950 & 18193 & 67.51 & 240 & 1922 & 10.56 & 73 & 3.80\\ \hline 
9292 & 27999 & 21836 & 77.99 & 340 & 3282 & 15.03 & 107 & 3.26\\ \hline 
9294 & 27110 & 21762 & 80.27 & 220 & 2336 & 10.73 & 37 & 1.58\\ \hline 
9295 & 28000 & 20386 & 72.81 & 340 & 3327 & 16.32 & 389 & 11.69\\ \hline 
9298 & 27090 & 18549 & 68.47 & 200 & 1882 & 10.15 & 22 & 1.17\\ \hline 
9299 & 16563 & 14323 & 86.48 & 440 & 3484 & 24.32 & 389 & 11.17\\ \hline 
9300 & 15170 & 13708 & 90.36 & 300 & 2038 & 14.87 & 261 & 12.81\\ \hline 
9301 & 16999 & 12451 & 73.25 & 280 & 2036 & 16.35 & 237 & 11.64\\ \hline 
9302 & 16999 & 12379 & 72.82 & 200 & 1385 & 11.19 & 127 & 9.17\\ \hline 
9303 & 16969 & 12680 & 74.72 & 500 & 3977 & 31.36 & 270 & 6.79\\ \hline 
9304 & 17000 & 13818 & 81.28 & 500 & 2590 & 18.74 & 187 & 7.22\\ \hline 
9305 & 16998 & 13781 & 81.07 & 280 & 1775 & 12.88 & 83 & 4.68\\ \hline 
9306 & 16937 & 10061 & 59.40 & 200 & 1474 & 14.65 & 28 & 1.90\\ \hline 
9307 & 16997 & 10823 & 63.68 & 220 & 1490 & 13.77 & 122 & 8.19\\ \hline 
9308 & 16979 & 15452 & 91.01 & 320 & 2675 & 17.31 & 92 & 3.44\\ \hline 
9309 & 16999 & 15411 & 90.66 & 520 & 3373 & 21.89 & 191 & 5.66\\ \hline 
9310 & 16960 & 11998 & 70.74 & 240 & 1870 & 15.59 & 67 & 3.58\\ \hline 
9311 & 16978 & 12093 & 71.23 & 480 & 3407 & 28.17 & 397 & 11.65\\ \hline 
9312 & 16978 & 12025 & 70.83 & 200 & 1653 & 13.75 & 28 & 1.69\\ \hline 
9313 & 16982 & 14349 & 84.50 & 300 & 1887 & 13.15 & 123 & 6.52\\ \hline 
9314 & 16965 & 14127 & 83.27 & 200 & 1319 & 9.34 & 103 & 7.81\\ \hline 
9315 & 17000 & 12307 & 72.39 & 340 & 2135 & 17.35 & 48 & 2.25\\ \hline 
9316 & 16999 & 11794 & 69.38 & 200 & 1378 & 11.68 & 12 & 0.87\\ \hline 
9317 & 17000 & 14124 & 83.08 & 420 & 2295 & 16.25 & 285 & 12.42\\ \hline 
9318 & 16998 & 13231 & 77.84 & 280 & 1660 & 12.55 & 155 & 9.34\\ \hline

    \end{tabular}
   } 
  } 
\end{table*}

An important need in many situations involving video collections
(archive video search/reuse, personal video organization/search,
surveillance, law enforcement, protection of brand/logo use) is to find more video segments of a certain specific person, object, or place, given one or more visual examples of the specific item. Building on the work from previous years in the concept detection task \cite{awad2016trecvid} the instance search task seeks to address some of these needs. For six years (2010-2015) the instance search task tested systems on retrieving specific instances of individual objects, persons and locations. A more challenging task and important goal in some applications is to combine two or more entities. Therefore, starting in 2016 a new query type, to retrieve specific persons in specific locations was introduced. The task spanned 3 years till 2018 and
since 2019 a similar query type has been adopted to retrieve instances of named persons doing named actions.

\subsubsection{Dataset}
Finding realistic test data, which contains sufficient recurrences of various specific objects/persons/locations under varying conditions has been difficult. Initially, the task was run for three years starting in 2010 to explore task definition and evaluation issues using data of three sorts: Sound and Vision (2010), British Broadcasting Corporation (BBC) rushes (2011), and Flickr (2012). 

In 2013 the task embarked on a multi-year effort using 464 h of the BBC soap opera EastEnders. 244 weekly ``omnibus'' files were divided by the BBC into 471\,523 video clips to be used as the unit of retrieval. The videos present a ``small world'' with a slowly changing set of recurring people (several dozen), locales (homes, workplaces, pubs, cafes, restaurants, open-air market, clubs, etc.), objects (clothes, cars, household goods, personal possessions, pets, etc.), and views (various camera positions, times of year, times of day). One dedicated video (Id 0) was provided for development where participants could use it in any way they wish, while the rest of the dataset episodes were used for testing. The usage of the BBC Eastenders proved to be very useful and adequate for the task and TRECVID has been using this same dataset since 2013.

\subsubsection{System task}

The instance search task for the systems was as follows. Given a collection of test videos, a master shot reference, a set of known predefined actions with example videos, and a collection of topics (queries) that delimit a specific person in some example images and videos, locate for each topic up to the 1000 clips most likely to contain a recognizable instance of the person performing one of the predefined named actions. Each query consisted of a set of:
\begin{itemize}
\item{The name of the target person}
\item{The name of the target action}
\item{4 example frame images drawn at intervals from videos
  containing the person of interest. For each frame image:}
\begin{itemize}
  \item{a binary mask covering one instance of the target person}
  \item{the ID of the shot from which the image was taken}
\end{itemize}
\item{4 - 6 short sample video clips of the target action}
\item{A text description of the target action}
\end{itemize}

Information about the use of the examples was reported by participants
with each submission. The possible categories for use of examples
were as follows:
\begin{enumerate} \itemsep0pt \parskip0pt
\item [A] - one or more provided images - no video used 
\item [E] - video examples (+ optional image examples) 
\end{enumerate}

Each run was also required to state the source of the training data used. This year participants were allowed to use training data from an external source, instead of, or in addition to the NIST provided training data. The following training options were available for the 2020 evaluation:
\begin{enumerate} \itemsep0pt \parskip0pt
\item[A] Only sample video 0
\item[B] Other external data
\item[C] Only provided images/videos in the query
\item[D] Sample video 0 AND provided images/videos in the query (A+C)
\item[E] External data AND NIST provided data (sample video 0 OR query images/videos)
\end{enumerate}

The task supported 2 types of runs that teams could submit for evaluation:
\begin{enumerate}
    \item Fully automatic (F) runs: System takes official query as input and produces results without any human intervention.
    \item Interactive humans in the loop (I) runs: System takes official query as input and produces results where humans can filter or re-rank search results for up to a period of 5 elapsed minutes per search and 1 user per system run.
\end{enumerate}
In the above both run types, all provided official query image/video examples should be frozen with no human modifications to them.

\subsubsection{Query Topics}

NIST reviewed a sample of test videos and developed a list of recurring
actions and the persons performing these actions. In order to test the effect of persons or actions on the performance of a given query, the topics tested different target persons performing the same actions. 
Besides the main task with unique queries each year, starting in 2019, a progress subtask was introduced to measure system progress on a set of fixed queries. In total, 20 common queries were released in 2019 and participating systems were allowed to submit results against those queries such that in 2020, NIST could evaluate 10 of those 20 queries to measure progress across two years (2019 - 2020) and evaluate the other 10 queries in 2021 measuring progress across 3 years (2019 - 2021). The 20 common queries comprised of 9 individual persons and 10 specific actions (Appendix \ref{appendixD}). 

A set of 20 unique queries (Appendix \ref{appendixC}) were released in the main task comprising of 8 individuals and 9 specific actions. In total, we evaluated those 20 queries in addition to 10 queries from the progress subtask set.

The guidelines for the task allowed the use of metadata assembled by the EastEnders fan community as long as its use was documented by participants and shared with other teams.

\subsubsection{Evaluation}

Each group was allowed to submit up to 4 runs (8 if submitting pairs
that differ only in the sorts of examples used). In total, 5 groups
submitted 33 runs including 31 automatic and 2 interactive runs.
From the 33 runs, 16 runs belonged to the progress subtask, while 17 belonged to the main 2020 task.
In addition to the 16 progress runs in 2020, a set of 12 progress runs were submitted by 3 separate teams in 2019. All 28 runs were evaluated and scored on 10 queries this year.

All run submissions were pooled and then divided into strata based on the
rank of the result items. Each stratum comprised of 20 rank levels (1-20, 21-40, 41-60, etc) 
up to rank 520. Finally, all duplicates in each stratum were removed.

For a given topic\footnote{Please refer to Appendix \ref{appendixC} and \ref{appendixD} for query descriptions.}, the submissions for that
topic were judged by a NIST human assessor who played each submitted shot
and determined if the topic target was present (the target person was seen doing the specific action).  The assessor started with the highest ranked stratum and worked his/her way down until too
few relevant clips were being found or time ran out. 

In general, submissions were pooled and judged down to at least rank 200, resulting
in 71\,251 judged shots including 4\,920 total relevant shots (6.9\%).
Table \ref{searchstats} presents information about the pooling and judging.

\subsubsection{Measures}
This task was treated as a form of search, and evaluated accordingly with average precision for each query in each run and per-run mean average precision (MAP) over all queries. While run-time and location accuracy
were also of interest here, of these two, only run-time was reported.

For detailed information about the approaches and results for individual teams' performance and runs, the reader should see the various site reports \cite{tv20pubs} in the online workshop notebook proceedings.

\subsubsection{Results}
\begin{figure}[htbp]
\begin{center}
\includegraphics[height=2.5in,width=3in,angle=0]{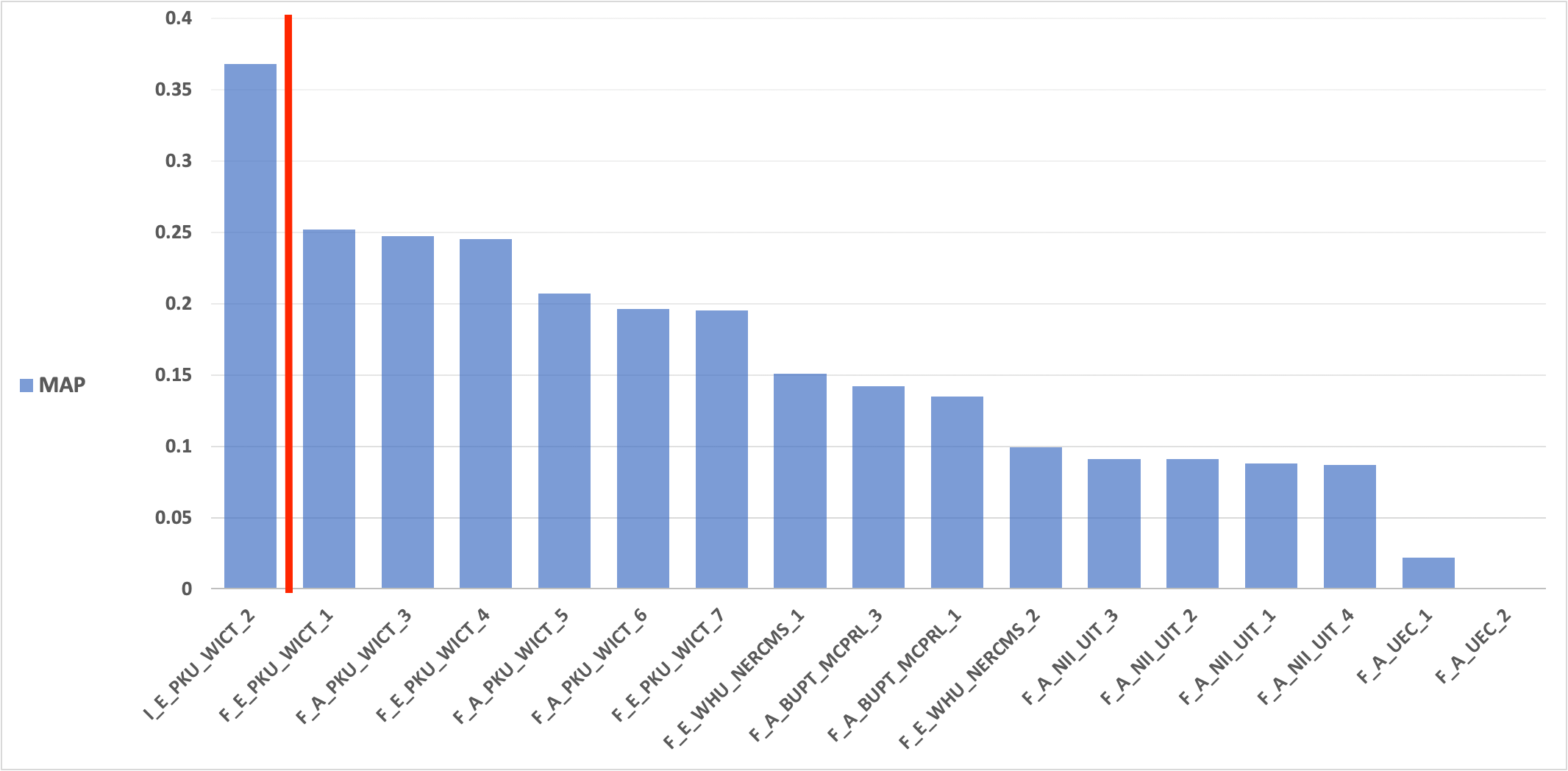}
\caption{INS: Mean average precision scores for automatic and interactive systems}
\label{ins.auto.scores}
\end{center}
\end{figure}

Figure \ref{ins.auto.scores} shows the sorted scores
of runs for both automatic and interactive systems. With only one interactive run submitted this year, this run has been included in the automatic runs chart. 

\begin{figure}[htbp]
\begin{center}
\includegraphics[height=2.5in,width=3in,angle=0]{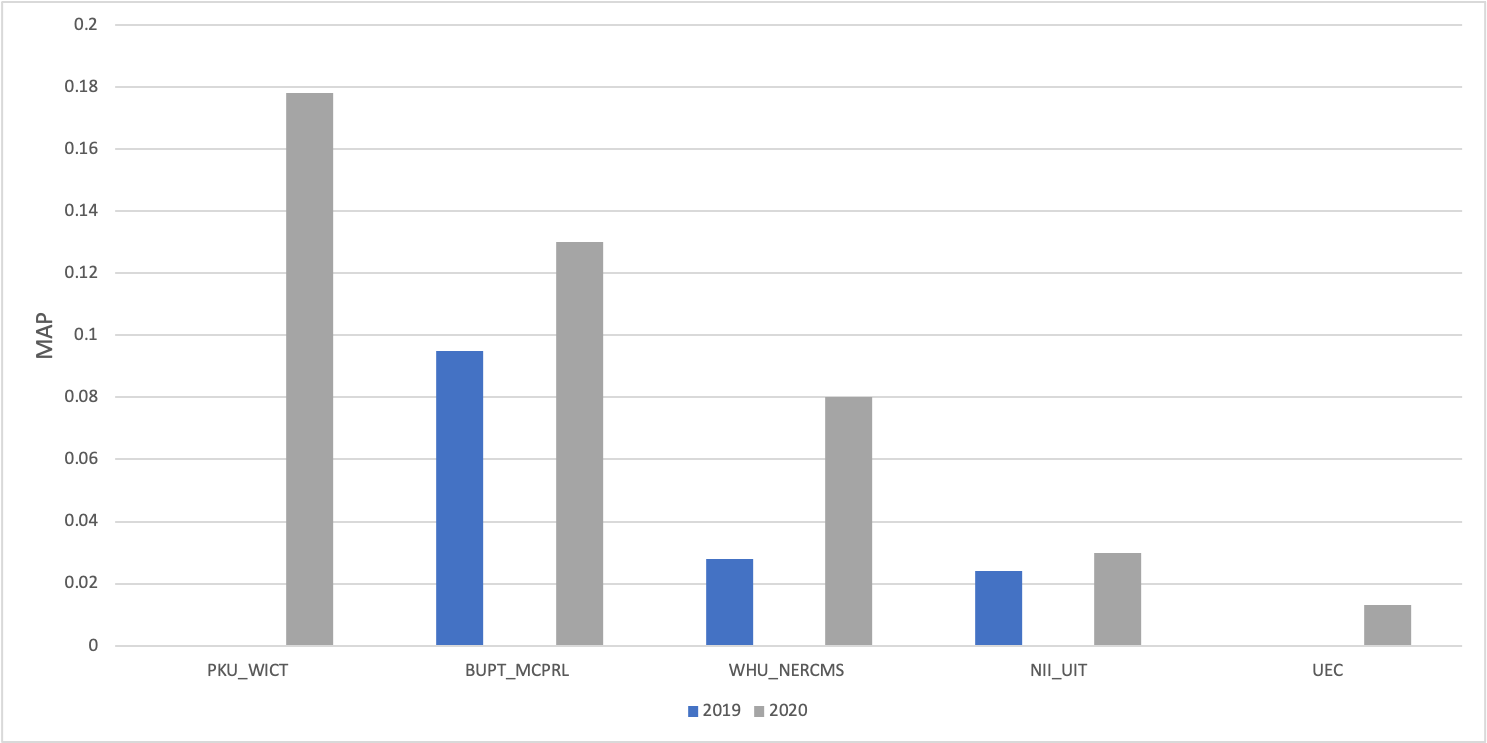}
\caption{INS: Mean average precision scores comparing results on 2019 and 2020 progress topics}
\label{ins.progress.scores}
\end{center}
\end{figure}

Figure \ref{ins.progress.scores} shows the progress topics scores for 2019 and 2020. From this chart we can see that teams who submitted progress runs in both years saw an increase in performance in 2020 compared to 2019, with two teams achieving a significant improvement in performance and one team achieving a more moderate improvement.

\begin{figure}
\begin{center}
\includegraphics[height=2.5in,width=3in,angle=0]{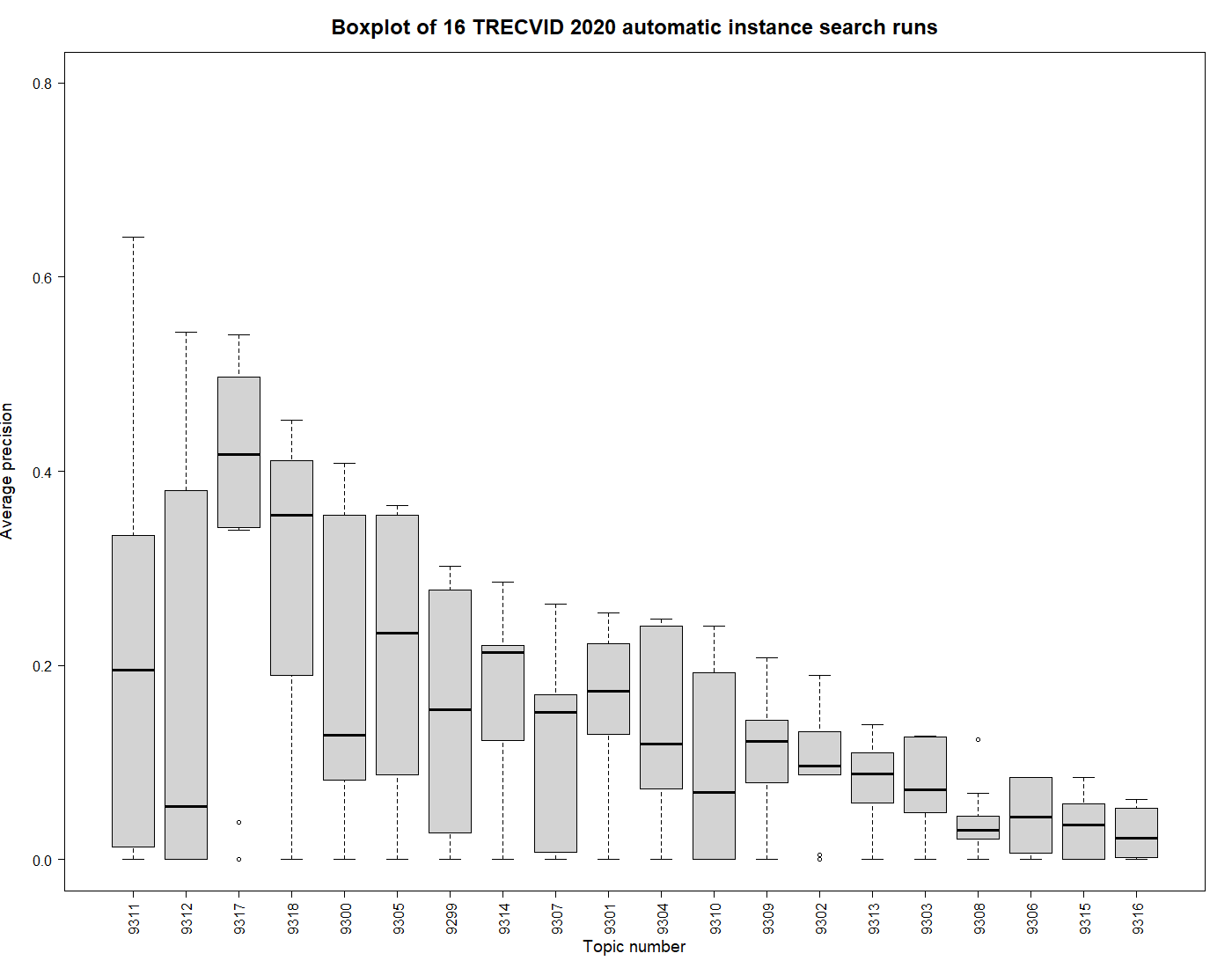}
\caption{INS: Boxplot of average precision by topic for automatic runs.}
\label{ins.auto.bp.topic.v.ap}
\end{center}
\end{figure}

Figure \ref{ins.auto.bp.topic.v.ap} shows the distribution of
automatic run scores (average precision) by topic as a box plot. The
topics are sorted by the maximum score with the best performing topic
on the left. Median scores vary from 0.4175 down to 0.022. 
The main factor affecting topic difficulty this year is the target action.

This figure shows interesting results for topics 9317, and 9318: Max, and Stacey holding phone. These topics do not score among the highest for maximum scores, but do have the highest median scores.

\begin{figure}
\begin{center}
\includegraphics[height=2.5in,width=3in,angle=0]{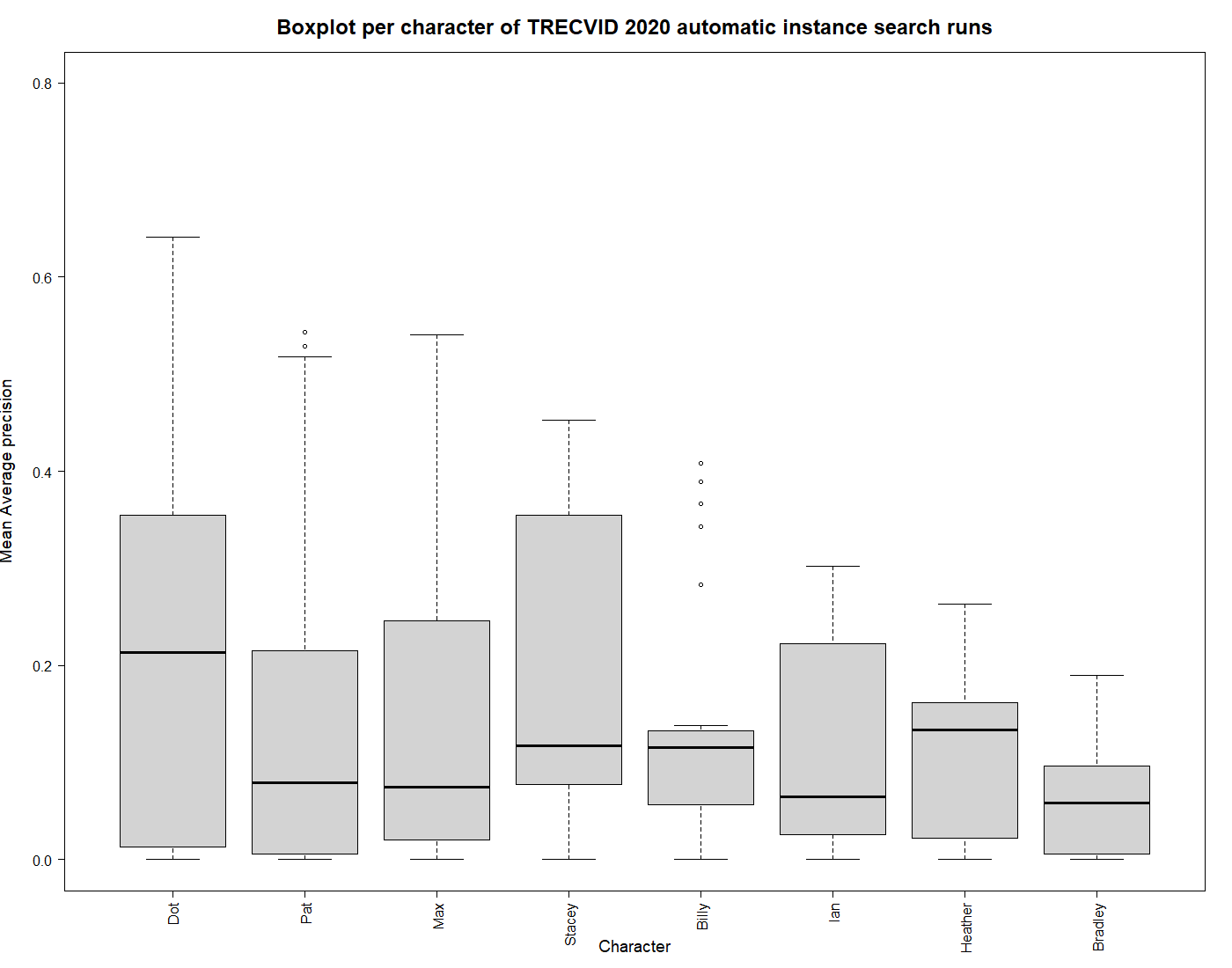}
\caption{INS: Boxplot of average precision by character for automatic runs.}
\label{ins.auto.bp.char.v.ap}
\end{center}
\end{figure}

\begin{figure}
\begin{center}
\includegraphics[height=2.5in,width=3in,angle=0]{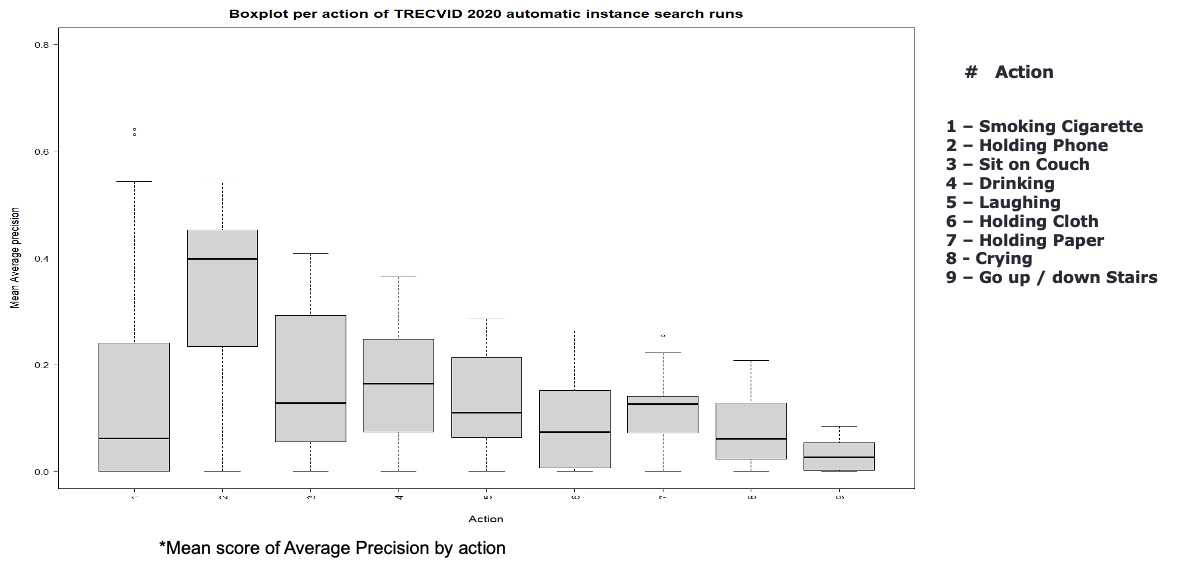}
\caption{INS: Boxplot of average precision by action for automatic runs.}
\label{ins.auto.bp.action.v.ap}
\end{center}
\end{figure}

Figures \ref{ins.auto.bp.char.v.ap} and \ref{ins.auto.bp.action.v.ap} show the distribution of automatic run scores by character and action. These are sorted by maximum score with the best performing character and action on the left.

\begin{figure}
\begin{center}
\includegraphics[height=2.5in,width=3in,angle=0]{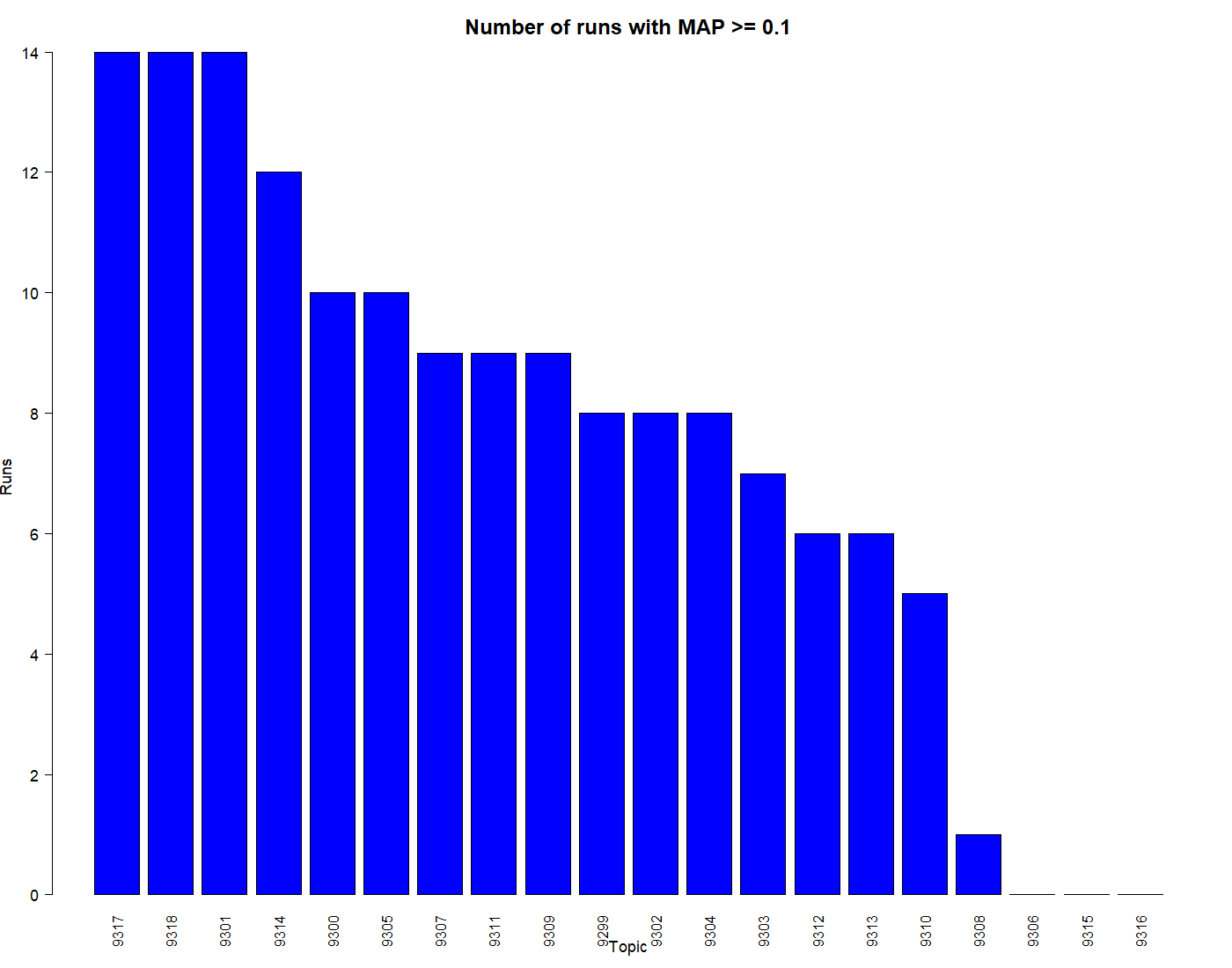}
\caption{INS: Easiest topics for automatic systems}
\label{ins.easy.topics}
\end{center}
\end{figure}

\begin{figure}
\begin{center}
\includegraphics[height=2.5in,width=3in,angle=0]{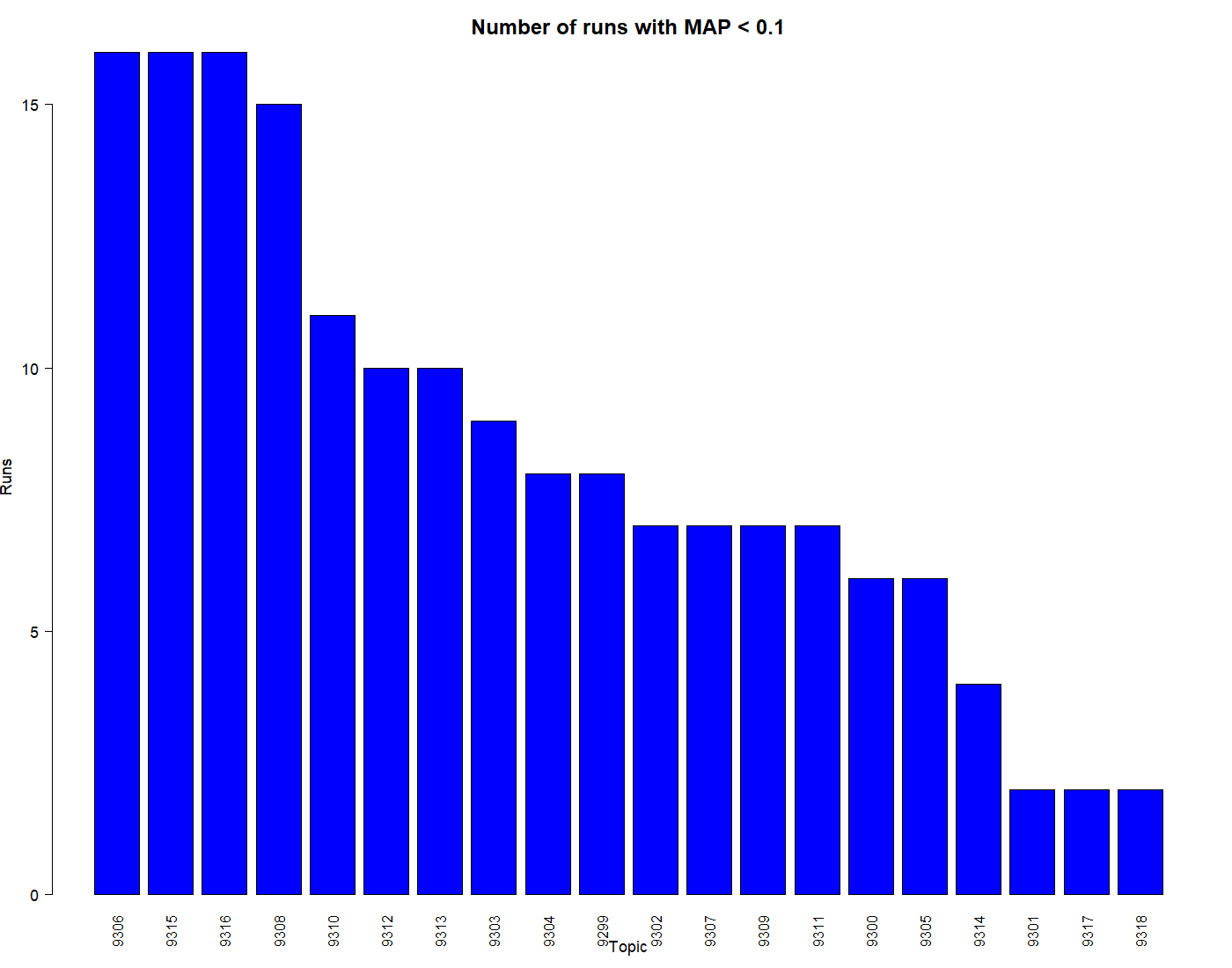}
\caption{INS: Hardest topics for automatic systems}
\label{ins.hard.topics}
\end{center}
\end{figure}

Figures \ref{ins.easy.topics} and \ref{ins.hard.topics} show the easiest and hardest topics, calculated by the number of runs with average precision scores above 0.1 and below 0.1 respectively. These figures show that holding a phone was the easiest action to find, while Going up / down stairs was the hardest action to find.

\begin{figure}[htbp]
\begin{center}
\includegraphics[height=2.5in,width=3in,angle=0]{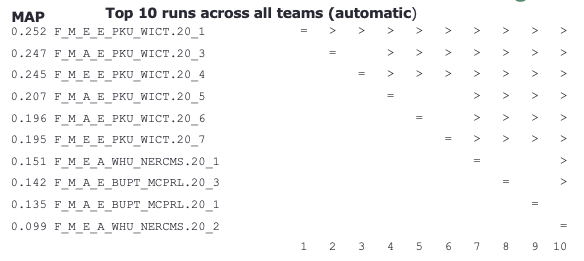}
\caption{INS: Randomization test results for top automatic runs. "E":runs used video examples. "A":runs used image examples only.}
\label{ins.auto.random.test}
\end{center}
\end{figure}

Figure \ref{ins.auto.random.test} documents the raw scores of the top 10
automatic runs and the results of a partial randomization test \cite{manly97} and sheds some light on which differences in ranking are
likely to be statistically significant. One angled bracket indicates p
$<$ 0.05. There are little significant differences between the top runs this year.

\begin{figure}[htbp]
\begin{center}
\includegraphics[height=2.5in,width=3in,angle=0]{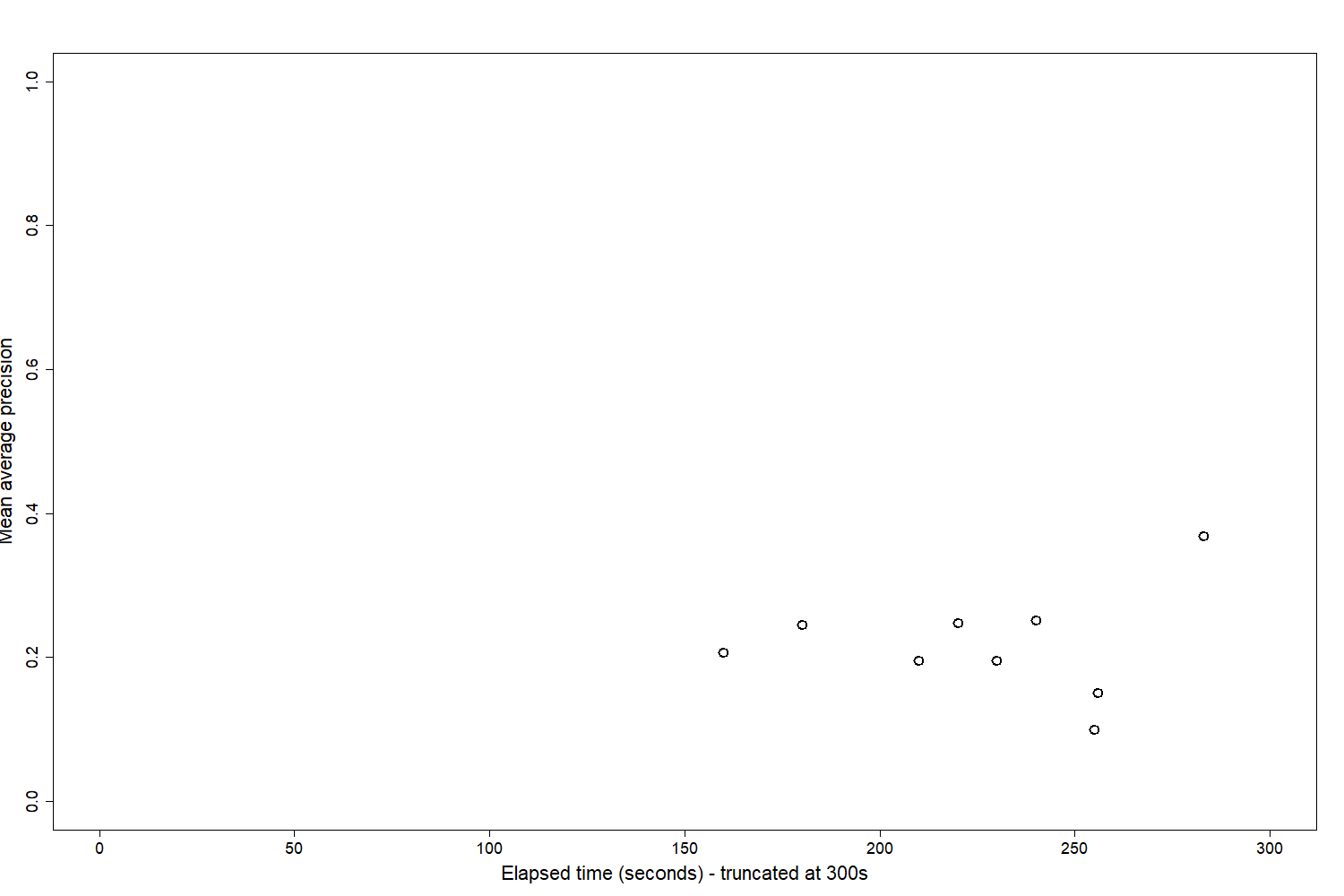}
\caption{INS: Mean average precision versus time for fastest runs}
\label{ins.map.vs.fastest}
\end{center}
\end{figure}

The relationship between effectiveness (mean average precision)
and elapsed processing time is depicted in Figure \ref{ins.map.vs.fastest} for the automatic runs with elapsed times less than or equal to 300s. Of those reported times below 300s, we can see that except for a couple of outliers, the most accurate systems take longer processing times.

\begin{figure}[htbp]
\begin{center}
\includegraphics[height=2.5in,width=3in,angle=0]{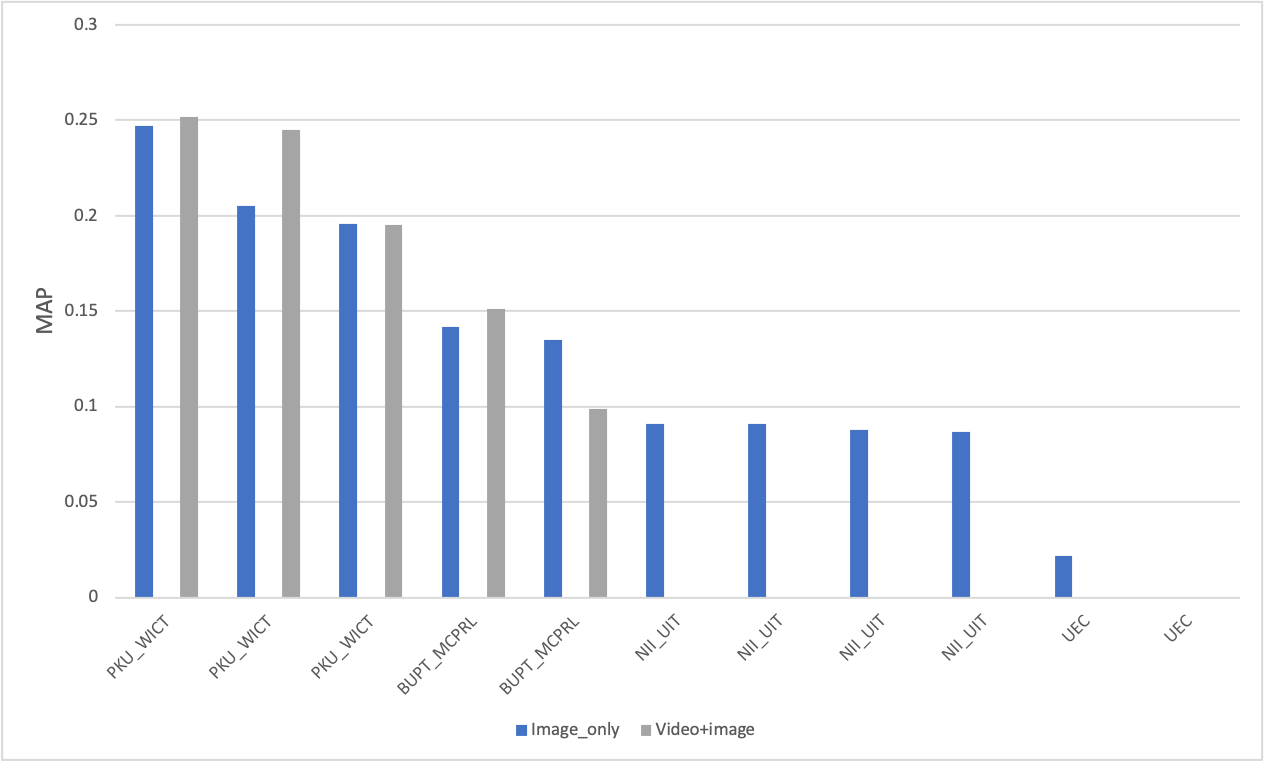}
\caption{INS: Effect of image vs video data type}
\label{ins.effect.datatype}
\end{center}
\end{figure}

Figure \ref {ins.effect.datatype} shows the relationship between the
two categories of runs (images only for training OR video and images) and 
the effectiveness of the runs. These show that far more runs make use of video and image examples than just image examples. Comparing results however for systems making use of both show that there was actually very little difference between results for systems that differed only in the category of runs (images only for training OR video and images).

\begin{figure}[htbp]
\begin{center}
\includegraphics[height=2.5in,width=3in,angle=0]{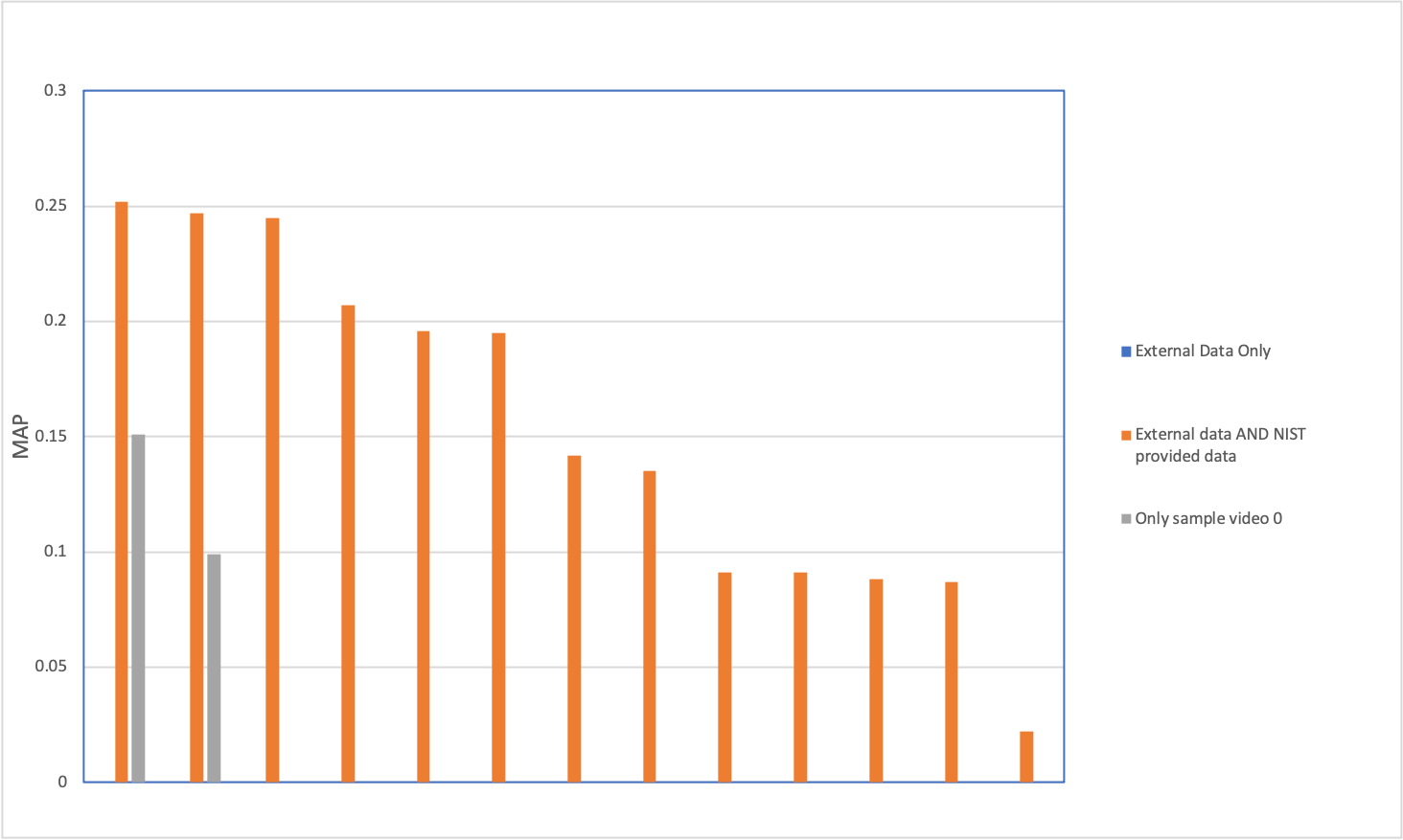}
\caption{INS: Effect of data source used}
\label{ins.effect.datasource}  
\end{center}
\end{figure}

Figure \ref{ins.effect.datasource} shows the effect of the data source used for training, with participants being able to use an external data source instead of or in addition to the NIST provided training data. The use of external data in addition to the NIST provided data provides by far the best results. The use of external data in addition to the NIST provided data is adopted by the vast majority of participating teams. Results for other external data only and sample video '0' only are similar, however, these are way below results for teams that use external data in addition to the NIST provided data, and very few teams use these data sources.

\subsubsection{Observations}
This was the second year the task used the new query type of person+action, and it was the fifth year using the Eastenders dataset. Although there was a slight increase in the number of participants who signed up for the task, this year there was one fewer team that completed the task, bringing the finishers this year to 5 out of 13 registered teams, compared to 6 out of 12 registered teams last year.

We should also note that this year a time consuming process was spent trying to get the data agreement set with the donor (BBC) which happened but may have affected the number of teams who did not get enough time to work on and finish the task.

Once again for this year of the task, participating teams could use external data instead of or in addition to NIST provided data. Results have shown that the use of external data in addition to the NIST provided data consistently provide better results. However, results also show that the use of external data instead of the NIST provided data, or NIST provided data only, provides weaker results. Teams could also again make use of video examples or image only examples. A majority of teams used video examples in this new task, however results from runs which differed only in the examples used showed very little difference between video examples and image examples only. 
 
BUPT\_MCPRL adopted a multi task CNN model, extracted face features based on dlib, \url{https://github.com/davisking/dlib}, to get 128-dim face embeddings and computed cosine distance between queries and detected persons. To address the problem of false negatives when a target person's face is turned away from the camera they adopted person tracking. The first frame in each shot where the target person appears was used as an initialization frame and the object tracker is set to the bounding box of the target person. They used ResNet50-SiamRPN as the object tracking module. For instance retrieval they divided instances into three categories: emotion related, human object interactions and general actions. For the emotion related category: They used crying, laughing and shouting as sad, happy or angry - emotion recognition models based on Region Attention Network taking FERPlus and CK+ as main training set. For Human-object interactions, they explored dependencies between semantic objects and human keypoints using object detection and pose estimation models. Human bounding boxes were fed into HRNet to estimate human pose. They calculated distance between object location and target persons interactive keypoint. This was used for holding glass, holding a phone, sitting on couch etc. For general action retrieval: kissing, hugging, go upstairs / downstairs, they used two different models to extract feature for action recognition. The first was image-based method ECOfull, which extracts action features directly from raw video. They also introduced STGCN to extract features only based on the human pose. Both models are trained on the Kinetics-600 dataset.

UEC performed person search by comparing the facial features of the person in each query with the people in each video to obtain a person similarity score. Faces in each frame were first detected using RetinaFae and then cropped. Facial features were then extracted from cropped images using ArcFace. Action retrieval consists of three parts: Emotion related action retrieval, Human-Object Interaction retrieval, and General action Retrieval. Emotions are recognized through facial expressions for which they used a model rained on the FER2013 dataset. For Human-Object Interaction retrieval they used EfficientDet which was pre-trained on MS-COCO. Other general classes were recognized using SlowFast. They fine-tuned the SlowFast pre-trained by Kinetics-600 with INS data.

PKU\_WICT proposed a two-stage approach consisting of similarity computing and result re-ranking. They used four aspects for action specific recognition: frame-level action recognition, video-level action recognition (trained using Kinetics-400), object detection (pre-trained on MS-COCO) and facial expression recognition. They finally computed the fusion value of all prediction scores of a shot as the final prediction score \textit{ActScore}. For person specific recognition they first detected faces in query examples and filtered out bad faces of low detection confidence and complimented with good faces of high detection confidence. Next, face features from queries and shots were extracted based on deep convolutional neural networks and calculated the similarity. Top N query expansion strategy was used conducted for further improving the retrieval results. They then used two fusion strategies to fuse scores from action specific and person specific recognition.

For person retrieval, WHU\_NERCMS used a face detection and recognition model pre-trained on a wilder face dataset to compute person retrieval scores. For action retrieval, they utilized a common action recognition model (TSM) pre-trained on kinetics dataset, and a human-object interaction model (PPDM) pre-trained on HICO-DET dataset to compute action retrieval scores. Person and action scores were fused to obtain the final result.

NII\_UIT used VGGFace2 to extract face embeddings / representations. Face representations were then matched and reranked using cosine similarity. For finding actions, actions were categorized into Human Object Interaction (HOI) actions and Facial Expression (FE) actions. They also proposed a heuristic method based on person search result and the distance between the target person and the desired object. This heuristic method specifies faces locations of the target in the top ranked shots. EfficientDet detector was used to specify object locations. For the facial expression actions, ESR was applied on target faces which predicted 10 facial expressions. For fusion of person and action search, they selected top shots from person rank list and re-rank these shots based on action score.

\subsubsection{Conclusions}
This was the second year of the updated Instance Search task in which queries comprised of a specific person doing a specific action. The action recognition part of the task made this task a more challenging problem than before the updated task, with maximum and average results still far below those of previous years for the specific person in a specific location queries. Results did however show an improvement over last year, which was the first year of the updated task.

There were a total of 5 finishers out of 13 participating teams in this year's task. All 5 finishers submitted notebook papers. All 5 teams submitted runs for the progress queries, 3 of which can be directly compared against their progress runs from last year which showed an improvement in results. We hope that all 5 teams will also submit progress runs next year which will allow more solid conclusions to be drawn.

\subsection{Disaster Scene Description and Indexing}
\begin{table*} \begin{center}

\begin{tabular}[htbp]{|c|c|c|c|c|}
\hline
Damage & Environment & Infrastructure & Vehicles & Water\\ \hline\hline
Misc. Damage & Dirt & Bridge & Aircraft & Flooding\\ \hline
Flooding/Water Damage& Grass & Building & Boat & Lake/Pond\\ \hline
Landslide& Lava & Dam/Levee & Car & Ocean\\ \hline
Road Washout& Rocks & Pipes & Truck & Puddle\\ \hline
Rubble/Debris& Sand & Utility or Power Lines/Electric Towers & & River/Stream\\ \hline
Smoke/Fire& Shrubs & Railway & &\\ \hline
& Snow/Ice & Wireless/Radio Communication Towers & &\\ \hline
& Trees & Water Tower & &\\ \hline
& & Road & &\\ \hline
\end{tabular}
\caption{DSDI: The testing dataset has 5 coarse categories, each divided into 4-9 more specific labels.}
\label{tab:dsdi.categories}
\end{center}
\end{table*}

Computer vision capabilities have rapidly been advancing and are expected to become an important component for incident and disaster response. Having prior knowledge about affected areas can be very helpful for the first responders. Communication systems often go down in major disasters, which makes it very difficult to get any information regarding the damage. Automated systems, such as robots or low flying drones, can therefore be used to gather information before rescue workers enter the area.

With the popularity of deep learning, computer vision research groups have access to very large image and video datasets for various tasks and the performances of systems have dramatically improved. However, the majority of computer vision capabilities are not meeting public safety’s needs, such as support for search and rescue, due to the lack of appropriate training data and requirements. Most current datasets do not have public safety hazard labels due to which state-of-the-art systems trained on these datasets fail to provide helpful labels in disaster scenes.

In response, the New Jersey Office of Homeland Security and MIT Lincoln Laboratory developed a dataset of images collected by the Civil Air Patrol of various natural disasters. The Low Altitude Disaster Imagery (LADI) dataset was developed as part of a larger NIST Public Safety Innovator Accelerator Program (PSIAP) grant. Two key properties of the dataset are as follows:

\begin{enumerate}
    \item Low altitude
    \item Oblique perspective of the imagery and disaster-related features.
\end{enumerate}

These are rarely featured in computer vision benchmarks and datasets. The LADI dataset acted as a starting point to help label a new video dataset with disaster-related features to be used as testing data in the DSDI task. The image dataset could be used for the training and development of systems for the DSDI task.

\begin{figure*}[htbp]
    \centering
    \includegraphics[width=1.0 \linewidth]{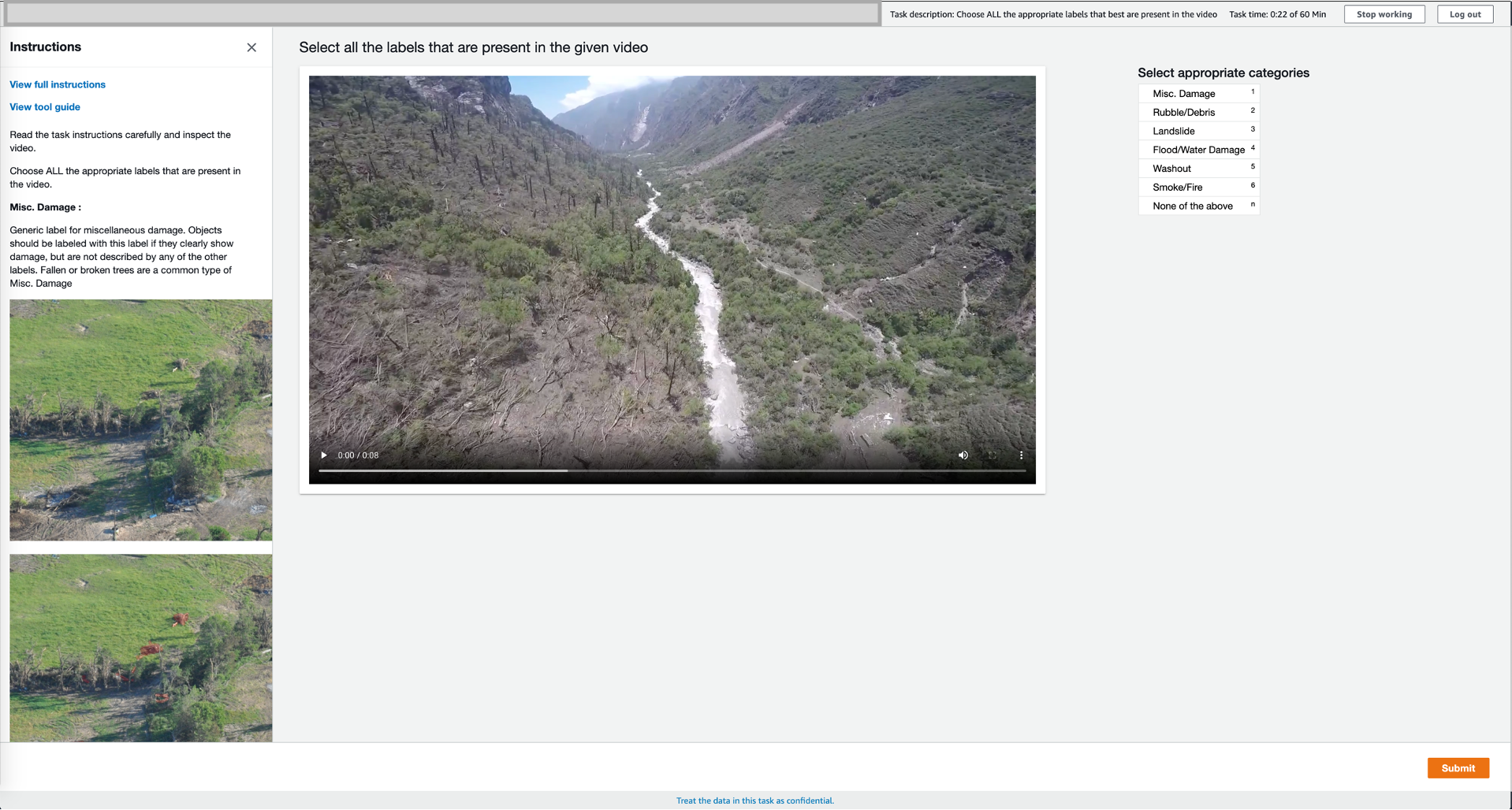}
    \caption{DSDI: Screenshot of a video being annotated for the Damage category. The annotator watches the video and marks all the labels that are visible in the video.}
    \label{fig:dsdi.annotation.page}
\end{figure*}

\subsubsection{Datasets}
\paragraph {\textbf{Training Dataset}}
The training dataset is based on the LADI dataset hosted as part of the AWS Public Dataset program. It consists of 20\,000+ annotated images. The images are from the Atlantic hurricane season. The lower altitude criterion distinguishes the LADI dataset from satellite datasets to support the development of computer vision capabilities with small drones operating at low altitudes. A minimum image size (4MB) was selected to maximize the efficiency of the crowd source workers, since lower resolution images are harder to annotate.

\paragraph {\textbf{Test Dataset}}
A pilot test dataset of about 5 hours of video was distributed for this task. The test dataset was segmented into small video clips (or shots) of a maximum duration of 20 seconds. The videos were from earthquake, hurricane, and flood affected areas. There were a total of 1825 shots with a median length of 16 seconds.

\begin{figure}[htbp]
    \centering
    \includegraphics[width=1.0 \linewidth]{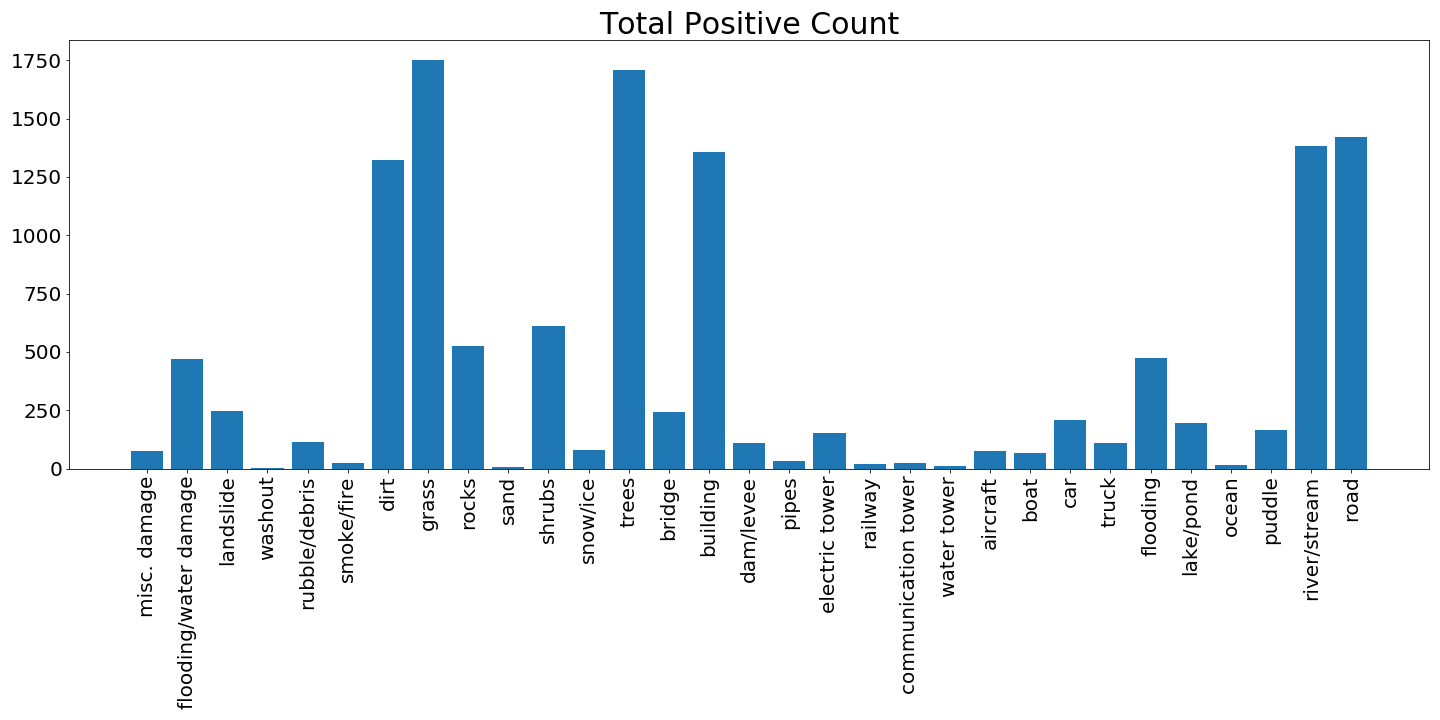}
    \caption{DSDI: Number of shots containing each feature (excluding Lava, which does not appear in any shots).}
    \label{fig:dsdi.features.positive}
\end{figure}

\paragraph{\textbf{Categories}}
The categories used for the testing dataset are the same as those used for the LADI training dataset. Five coarse categories were selected based on their importance for the task, and each of these categories is divided into 4-9 more specific labels. The hierarchical labeling scheme is shown in Table~\ref{tab:dsdi.categories}.

As can be expected from a real-world dataset, features appear with varied frequency within the videos. Some features such as grass, trees, buildings, roads, etc. appear much more frequently than others. The lava feature does not appear in any of the shots in the testing dataset. Figure~\ref{fig:dsdi.features.positive} shows the number of shots that contain each feature. 

\paragraph{\textbf{Annotation}}
The video annotation was done using full time annotators instead of crowdsourcing. It is essential that the annotators become familiar with the task and the labels before they start a category. For this reason, we created a practice page for each category with multiple examples for each label within that category. The annotators were given 2 videos as a test to mark the labels visible to them, and the answers were compared to ours. We also had regular discussions with the annotators to understand their process and clarify any confusion during the labeling of the dataset. 

Two full time annotators labeled the testing dataset. The Amazon Augmented AI (Amazon A2I) tool was used during the process. The annotators worked independently on each category. Figure~\ref{fig:dsdi.annotation.page} shows a screenshot of the annotation page as visible to annotators. To create the final ground truth, for each shot, the union of the labels was used.

\subsubsection{System Task}

Systems were required to return a ranked list of up to 1000 shots for each of the 32 features. Each submitted run specified its training type:
\begin{itemize}
    \item LADI-based (L): The run only used the supplied LADI dataset for development of its system.
    \item Non-LADI (N): The run did not use the LADI dataset, but only trained using other dataset(s).
    \item LADI + Others (O): The run used the LADI dataset in addition to any other dataset(s) for training purposes.
\end{itemize}

\subsubsection{Evaluation and Metrics}
The evaluation metric used for the task was mean average precision (MAP). The average precision is calculated for each feature, and the mean average precision is reported for each submission. Furthermore, the true positive, true negative, false positive, and false negative rates are also reported. Teams self reported the clock time per inference to compare the speeds of the various systems. 

\subsubsection{Results}
In this first year for the task, 17 teams signed up to join the task and finally 9 teams submitted runs. In total, we received 30 runs including 9 LADI+Others (O) runs and 21 LADI-based (L) runs. For detailed information about the approaches and results for individual teams' performances and runs, we refer the reader to the site reports \cite{tv20pubs} in the online workshop notebook proceedings. We present the overall results in this section.

None of the videos in the testing dataset had any occurrences of the lava feature, and so that feature was removed from all result calculations.

\begin{figure}[htbp]
    \centering
    \includegraphics[width=1.0 \linewidth]{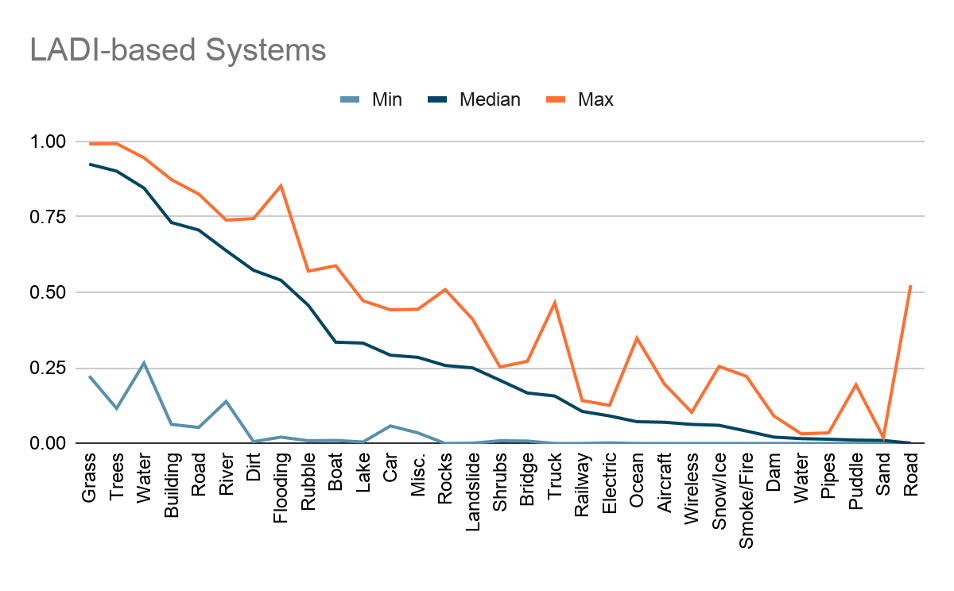}
    \caption{DSDI: Average precision values for each feature for systems with training type L.}
    \label{fig:dsdi.avg.prec.ladi}
\end{figure}

\begin{figure}[htbp]
    \centering
    \includegraphics[width=1.0 \linewidth]{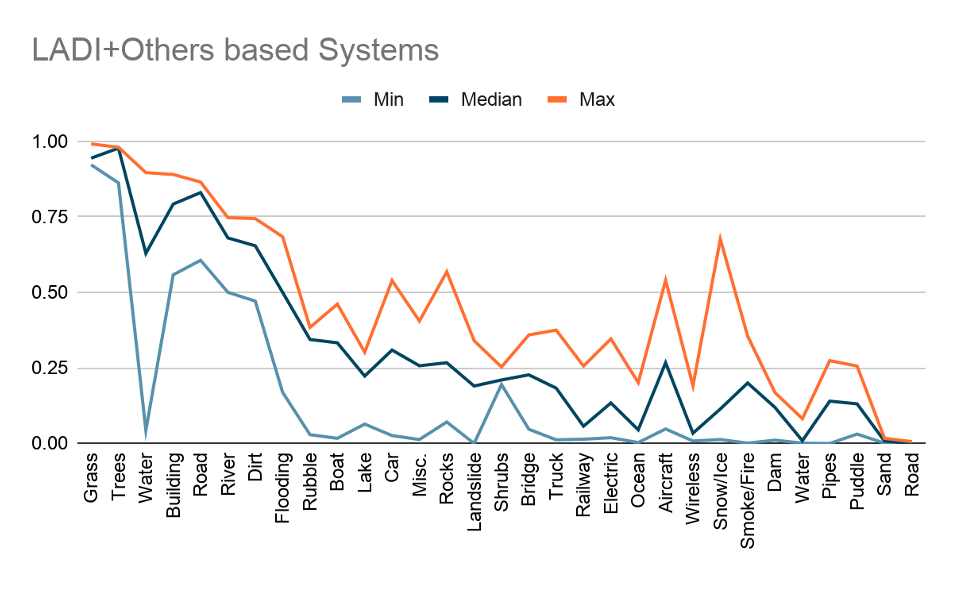}
    \caption{DSDI: Average precision values for each feature for systems with training type O.}
    \label{fig:dsdi.avg.prec.ladi.others}
\end{figure}

Figures~\ref{fig:dsdi.avg.prec.ladi} and~\ref{fig:dsdi.avg.prec.ladi.others} show the average precision scores for each feature for systems with run types L and O respectively. Systems tend to perform well on features that are commonly seen in training data, such as grass, trees, buildings, etc. 

\begin{figure*}[htbp]
    \centering
    \includegraphics[width=1.0 \linewidth]{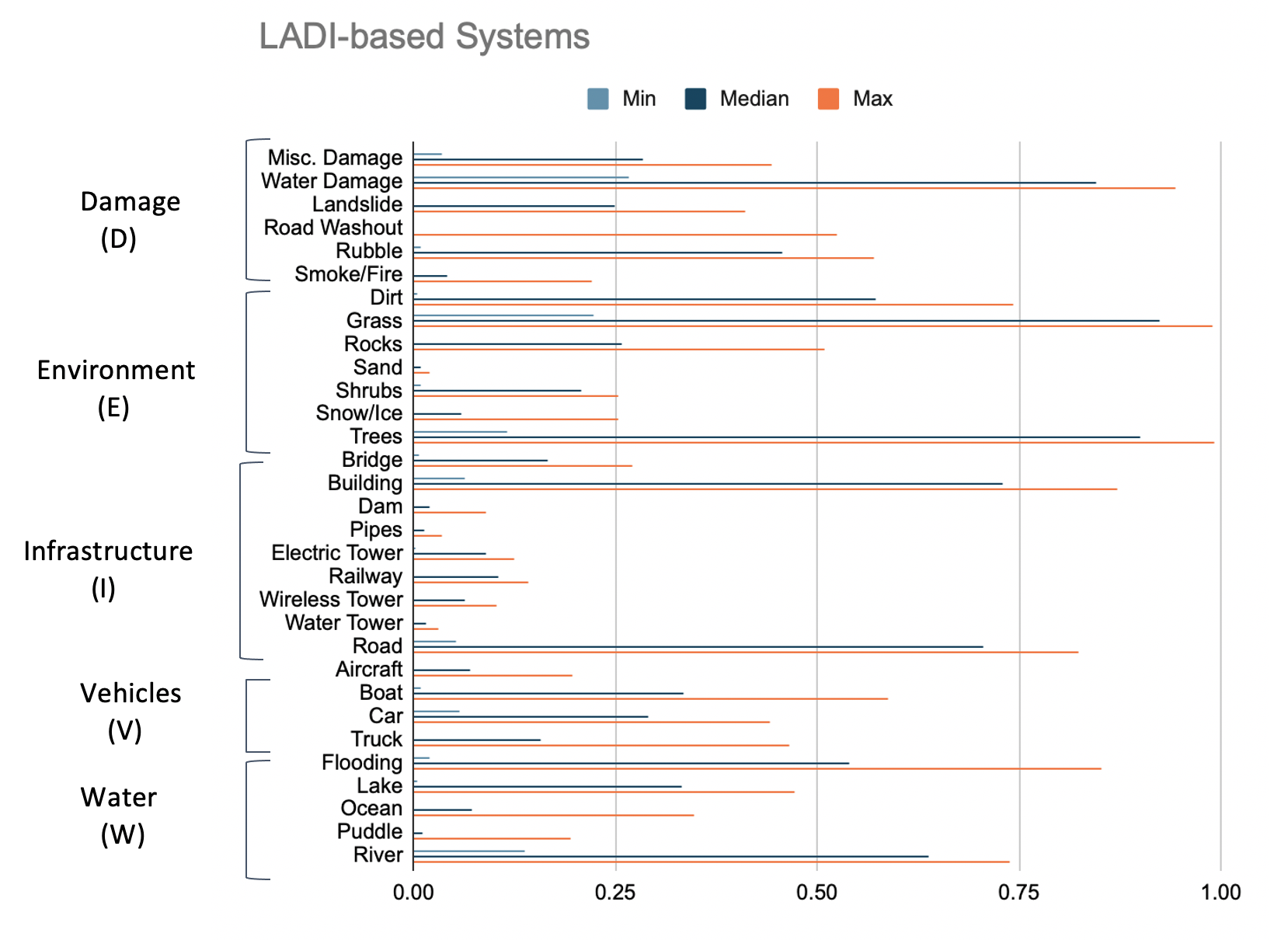}
    \caption{DSDI: Average precision values organized by categories for systems with training type L.}
    \label{fig:dsdi.results.categories.ladi}
\end{figure*}

\begin{figure*}[htbp]
    \centering
    \includegraphics[width=1.0 \linewidth]{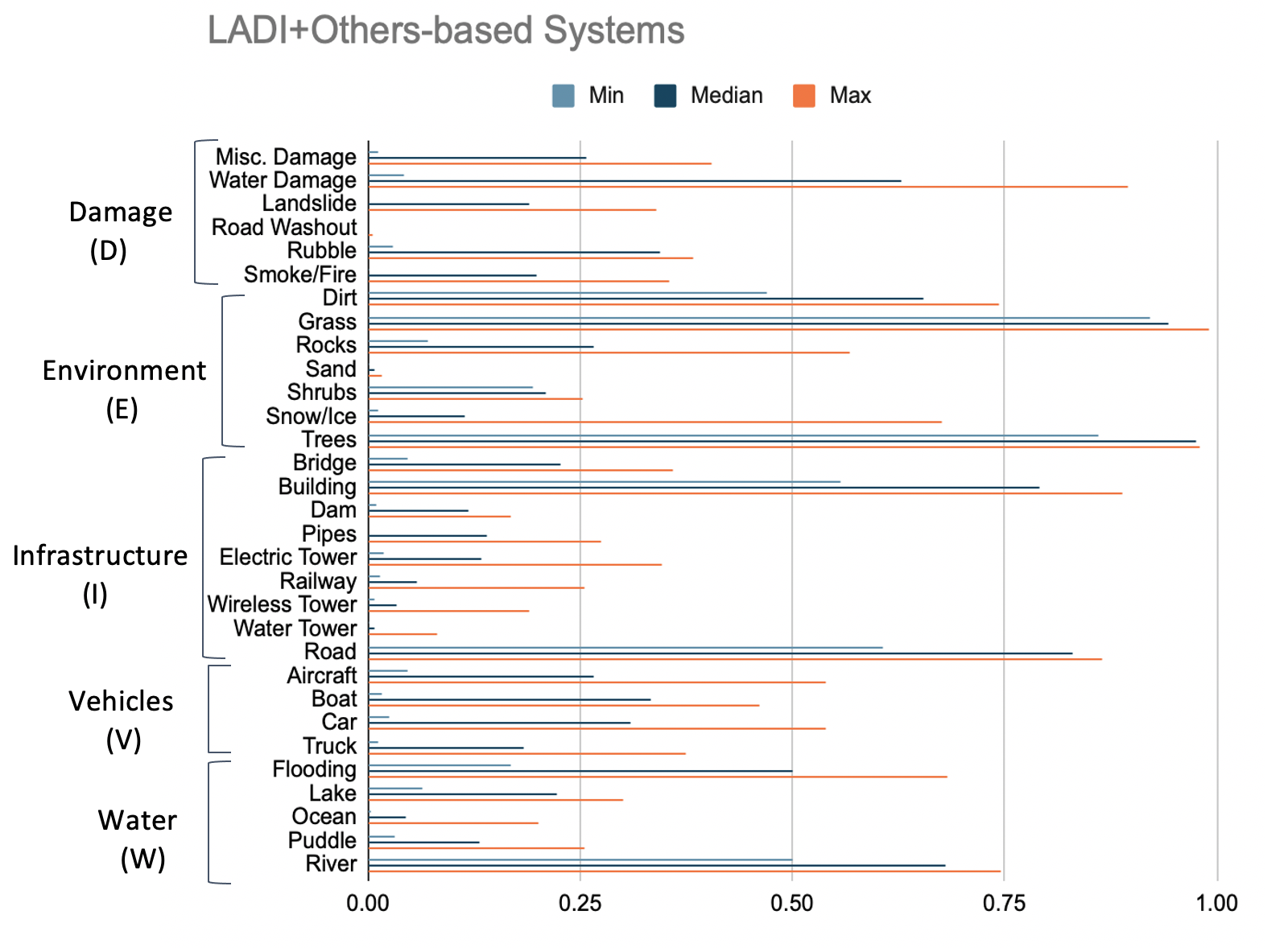}
    \caption{DSDI: Average precision values organized by categories for systems with training type O.}
    \label{fig:dsdi.results.categories.ladi.others}
\end{figure*}

Figures~\ref{fig:dsdi.results.categories.ladi} and~\ref{fig:dsdi.results.categories.ladi.others} show the average precision values organized by categories for run types L and O respectively. These charts show how the systems perform on features within each category. For example, the water damage feature in the damage category has a much higher score than any of the other features within that category. The main reason for this is that the LADI training dataset contains a large number of images with this label. 

\begin{figure}[htbp]
    \centering
    \includegraphics[width=1.0 \linewidth]{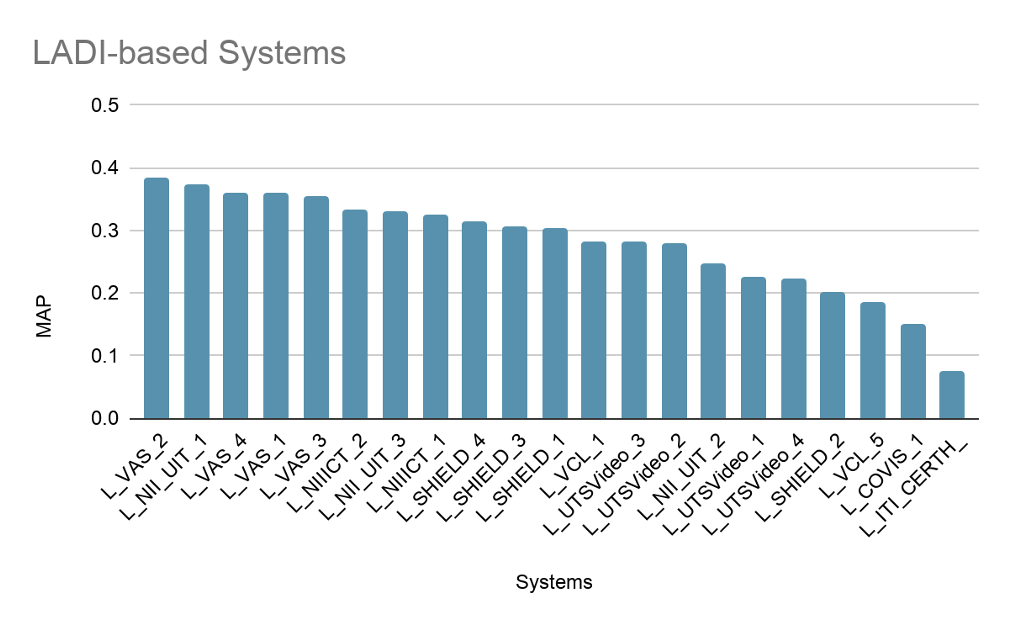}
    \caption{DSDI: Mean average precision score for each run with training type L.}
    \label{fig:dsdi.results.teams.ladi}
\end{figure}

\begin{figure}[htbp]
    \centering
    \includegraphics[width=1.0 \linewidth]{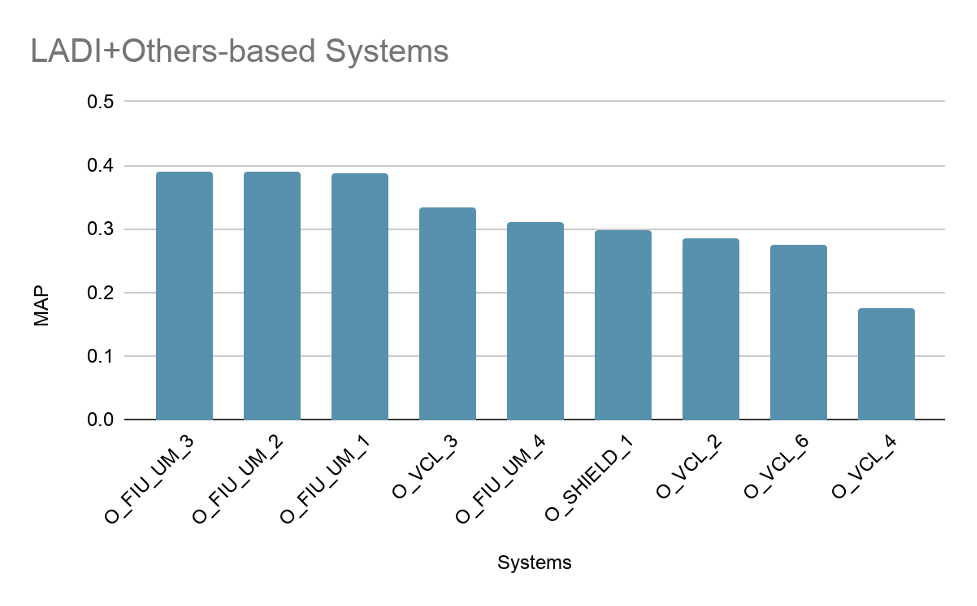}
    \caption{DSDI: Mean average precision score for each run with training type O.}
    \label{fig:dsdi.results.teams.ladi.others}
\end{figure}

Finally, Figures~\ref{fig:dsdi.results.teams.ladi} and~\ref{fig:dsdi.results.teams.ladi.others} show the mean average precision score for each run with training type L and O respectively. 

\begin{figure*}[htbp]
    \centering
    \includegraphics[width=1.0 \linewidth]{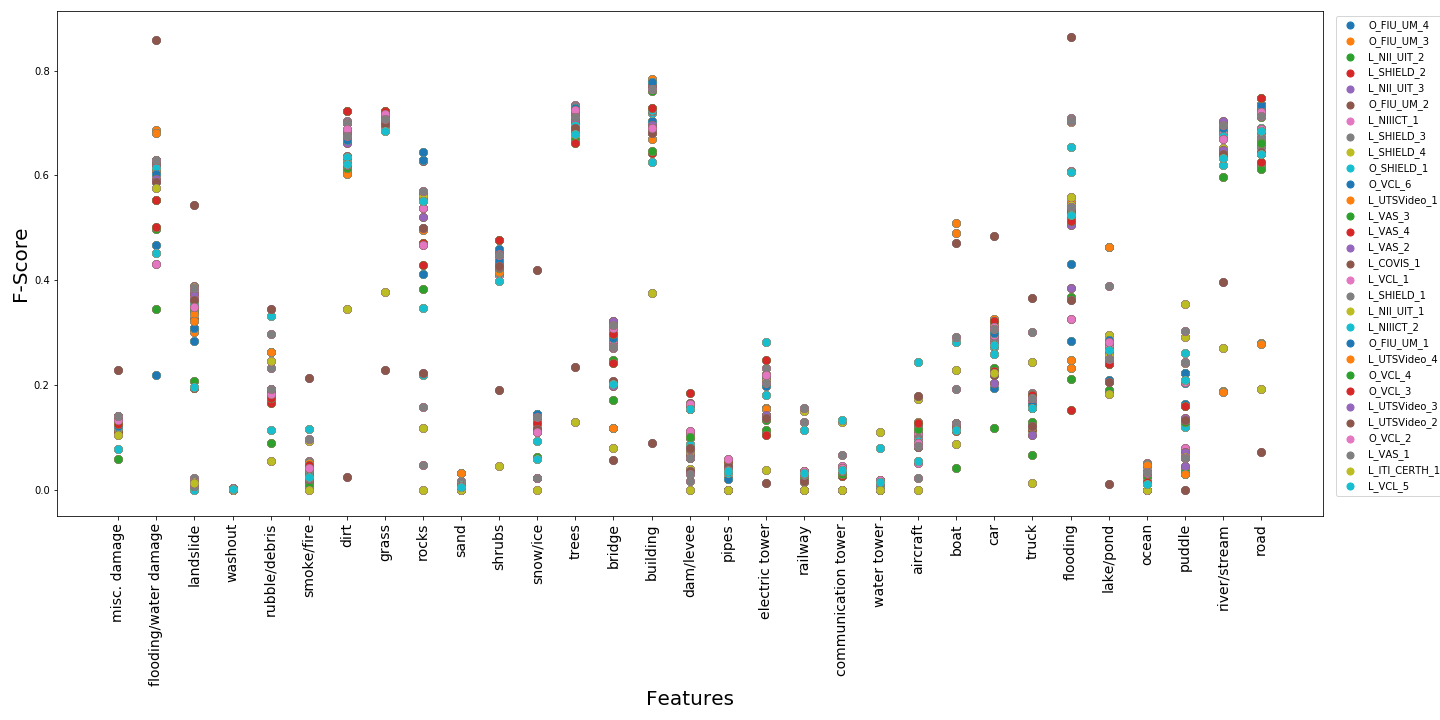}
    \caption{DSDI: F-measure for all the runs.}
    \label{fig:dsdi.F.measure}
\end{figure*}

We also reported the true positives, true negatives, false positives, and false negatives for each run. The F-measure using these values is shown in Figure~\ref{fig:dsdi.F.measure}.

\subsubsection{Conclusion and Future Work}

The DSDI pilot task was successful and it shows the need for datasets and benchmarks in the public safety domain. The teams performed reasonably well on labels that were well represented in the training dataset. Some known issues with the LADI dataset are:
\begin{enumerate}
    \item Labels can be noisy since crowd-sourced annotation is used.
    \item There is a class imbalance as certain classes are more prevalent in nature. There is also a limited budget that requires data collection efforts to focus on certain types of events.
    \item Classes that are larger or more widespread (e.g. floods, grass, trees) are easier to recognize than smaller ones (e.g. vehicles, pipes, water towers).
\end{enumerate}

The DSDI testing dataset was labeled by dedicated annotators, which resulted in cleaner annotation. Multiple improvements are expected in the future training dataset including incorporating community label improvements, as well spatio-temporal data integration to include additional GIS features based on image location.

The task will continue next year with a similar amount of testing video data. 

\subsection{Video to Text Description}
\begin{table*} \begin{center}

\begin{tabular}[htbp]{|c|c|c|}
\hline
&Matching \& Ranking (4 Runs) & Description Generation (19 Runs)\\ \hline\hline
IMFD\_IMPRESEE&  & X\\ \hline
KSLAB&  & X\\ \hline
KU\_ISPL&  & X\\ \hline
MMCUniAugsburg&  & X\\ \hline
PICSOM&  & X\\ \hline
RUC\_AIM3& X & X\\ \hline
\end{tabular}
\caption{VTT: List of teams participating in each of the subtasks. Description Generation is a core task, whereas Matching and Ranking is optional.}
\label{tab:vtt.participants}
\end{center}
\end{table*}

Automatic annotation of videos using natural language text descriptions has been a long-standing goal of computer vision. The task involves understanding many concepts such as objects, actions, scenes, person-object relations, the temporal order of events throughout the video, to mention a few. In recent years there have been major advances in computer vision techniques which enabled researchers to start practical work on solving the challenges posed in automatic video captioning. 

There are many use-case application scenarios which can greatly benefit from the technology, such as video summarization in the form of natural language,  facilitating the searching and browsing of video archives using such descriptions, describing videos as an assistive technology, etc. In addition, learning video interpretation and temporal relations among events in a video will likely contribute to other computer vision tasks, such as prediction of future events from the video. 

The ``Video to Text Description'' (VTT) task was introduced in TRECVID 2016. Since then, there have been substantial improvements in the dataset and evaluation.

\subsubsection{System Task}

The VTT task is divided into two subtasks:
\begin{itemize}
    \item Description Generation Subtask
    \item Matching and Ranking Subtask
\end{itemize}

The description generation subtask has been designated as core/mandatory, which means that teams participating in the VTT task must submit at least one run to this subtask. The matching and ranking subtask is optional for the participants. This subtask was initially introduced to ease teams into the difficult video description task. However, with improvements over subsequent years, the subtask was made optional. 

Details of the two subtasks are as follows:

\begin{itemize}
\item \textbf{Description Generation} (Core): For each video, automatically generate a text description of 1 sentence independently and without taking into consideration the existence of any annotated descriptions for the videos.
\item \textbf{Matching and Ranking} (Optional): In this subtask, 5 sets of text descriptions are provided along with the videos. Each set contains a description for each video in the dataset, but the order of descriptions is randomized. The goal of the subtask is to return for each video a ranked list of the most likely text description that corresponds (was annotated) to that video from each of the 5 sets. An interesting addition this year was to include fake sentences, i.e. sentences not corresponding to any videos, in the sets of descriptions. The goal was to check how such sentences would be ranked.
\end{itemize}
Up to 4 runs were allowed per team for each of the subtasks.

For this year, 6 teams participated in the VTT task. Only 1 team participated in the optional matching and ranking subtask with a total of 4 runs. There were 19 runs submitted for the description generation subtask. A summary of participating teams is shown in Table~\ref{tab:vtt.participants}.

\subsubsection{Data}

The VTT data for 2020 was taken from the V3C2 data collection. In previous years, the VTT testing dataset consisted of Twitter Vine videos, which generally had a duration of 6 seconds. In 2019, we supplemented the dataset with videos from Flickr. The V3C dataset \cite{rossetto2019v3c} is a large collection of videos from Vimeo. It also provides us with the advantage that we can distribute the videos rather than links, which may not be available in the future. 

For the purpose of this task, we only selected video segments with lengths between 3 and 10 seconds. A total of 1700 video segments were annotated manually by multiple annotators for this year's task.

\begin{figure}[htbp]
  \centering
  \includegraphics[width=1.0\linewidth]{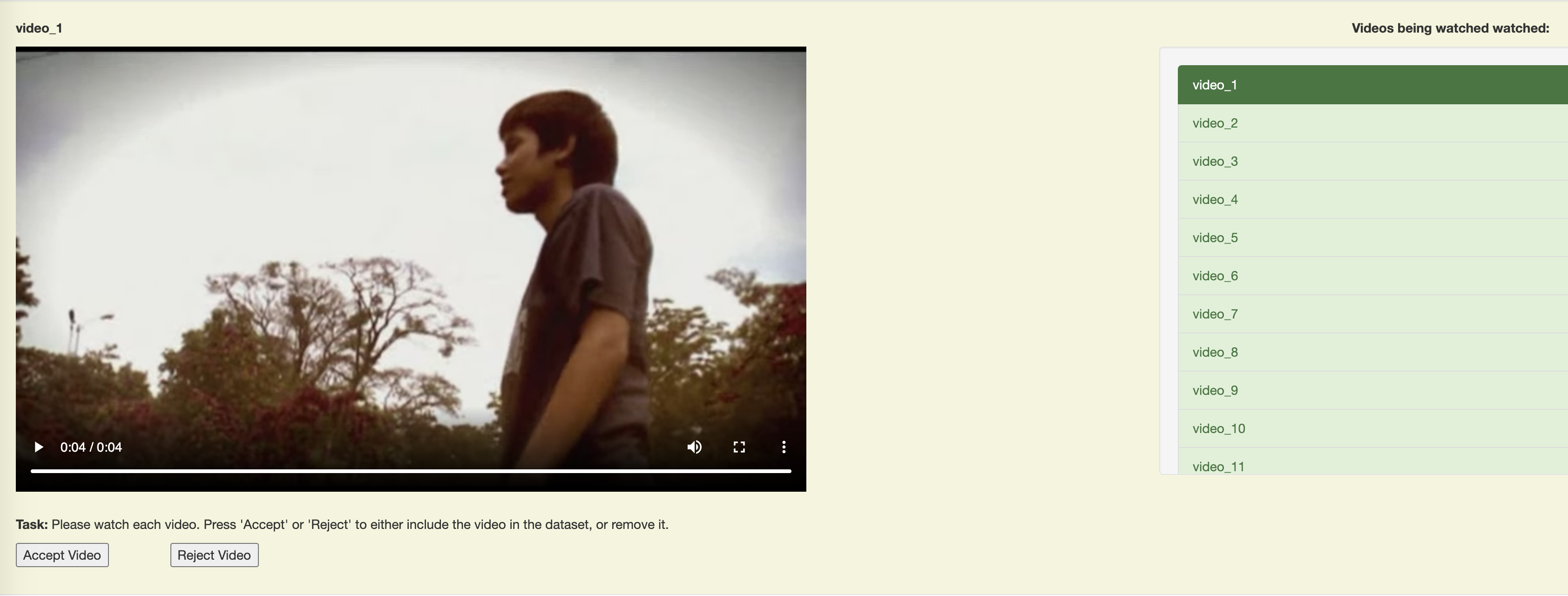}
  \caption{VTT: Screenshot of video selection tool.}
  \label{fig:vtt.video.selection}
\end{figure}

It is important for a good dataset to have a diverse set of videos. We reviewed over 8000 videos and selected 1700 videos. Figure \ref{fig:vtt.video.selection} shows a screenshot of the video selection tool that was used to decide whether a video was to be selected or not. We tried to ensure that the videos covered a large set of topics. If we came across a large number of videos that looked similar to previously selected clips, they were rejected. We also removed the following types of videos:
\begin{itemize}
    \item Videos with multiple, unrelated segments that are hard to describe, even for humans.
    \item Any animated videos.
    \item Other videos that may be considered inappropriate or offensive.
\end{itemize}



\begin{table}[htbp]
    \centering
    
    \begin{tabular}{|c|c|c|}
    \hline
    Annotator  & Avg. Length & Total Videos Watched\\
    \hline\hline
       1  & 16.60 & 825\\
       \hline
       2  & 16.65 & 875\\
       \hline
       3  & 17.67 & 1700\\
       \hline
       4  & 19.62 & 825\\
       \hline
       5  & 21.22 & 875\\
       \hline
       6  & 22.61 & 875\\
       \hline
       7  & 22.71 & 875\\
       \hline
       8  & 24.14 & 825\\
       \hline
       9  & 25.81 & 825\\
       \hline 
    \end{tabular}
    \caption{VTT: Average number of words per sentence for all the annotators. A large variation is observed between average sentence lengths for the different annotators. The table also shows the number of videos watched by each annotator. Annotator \#3 watched all 1700 videos.}
    \label{tab:avg_gt_length}
\end{table}

\paragraph {\textbf{Annotation Process}}

The videos were divided amongst 9 annotators, with each video being annotated by exactly 5 people. 

The annotators were asked to include and combine into 1 sentence, if appropriate and available, four facets of the video they are describing:

\begin{itemize}
\item{\textbf{Who} is the video showing (e.g., concrete objects and beings, kinds of persons, animals, or things)}?
\item{\textbf{What} are the objects and beings doing (generic actions, conditions/state or events)}?
\item{\textbf{Where} is the video taken (e.g., locale, site, place, geographic location, architectural)}?
\item{\textbf{When} is the video taken (e.g., time of day, season)}?
\end{itemize}

Different annotators provide varying amount of detail when describing videos. Some people try to incorporate as much information as possible about the video, whereas others may write more compact sentences. Table~\ref{tab:avg_gt_length} shows the average number of words per sentence for each of the annotators. The average sentence length varies from 16.60 words to 25.81 words, emphasizing the difference in descriptions provided by the annotators. The overall average sentence length for the dataset is 20.46 words.

Furthermore, the annotators were also asked the following questions for each video:
\begin{itemize}
\item Please rate how difficult it was to describe the video.
\begin{enumerate}
    \item Very Easy
    \item Easy
    \item Medium
    \item Hard
    \item Very Hard
\end{enumerate}
\item How likely is it that other assessors will write similar descriptions for the video?
\begin{enumerate}
    \item Not Likely
    \item Somewhat Likely
    \item Very Likely
\end{enumerate}
\end{itemize}

The average score for the first question was 2.53 (on a scale of 1 to 5), showing that in general the annotators thought the videos were on the easier side to describe. The average score for the second question was 2.24 (on a scale of 1 to 3), meaning that they thought that other people would write a similar description as them for most videos. The two scores are negatively correlated as annotators are more likely to think that other people will come up with similar descriptions for easier videos. The Pearson correlation coefficient between the two questions is -0.61.

\subsubsection{Submissions}

\begin{figure}[htbp]
\centering
\includegraphics[width=1.0\linewidth]{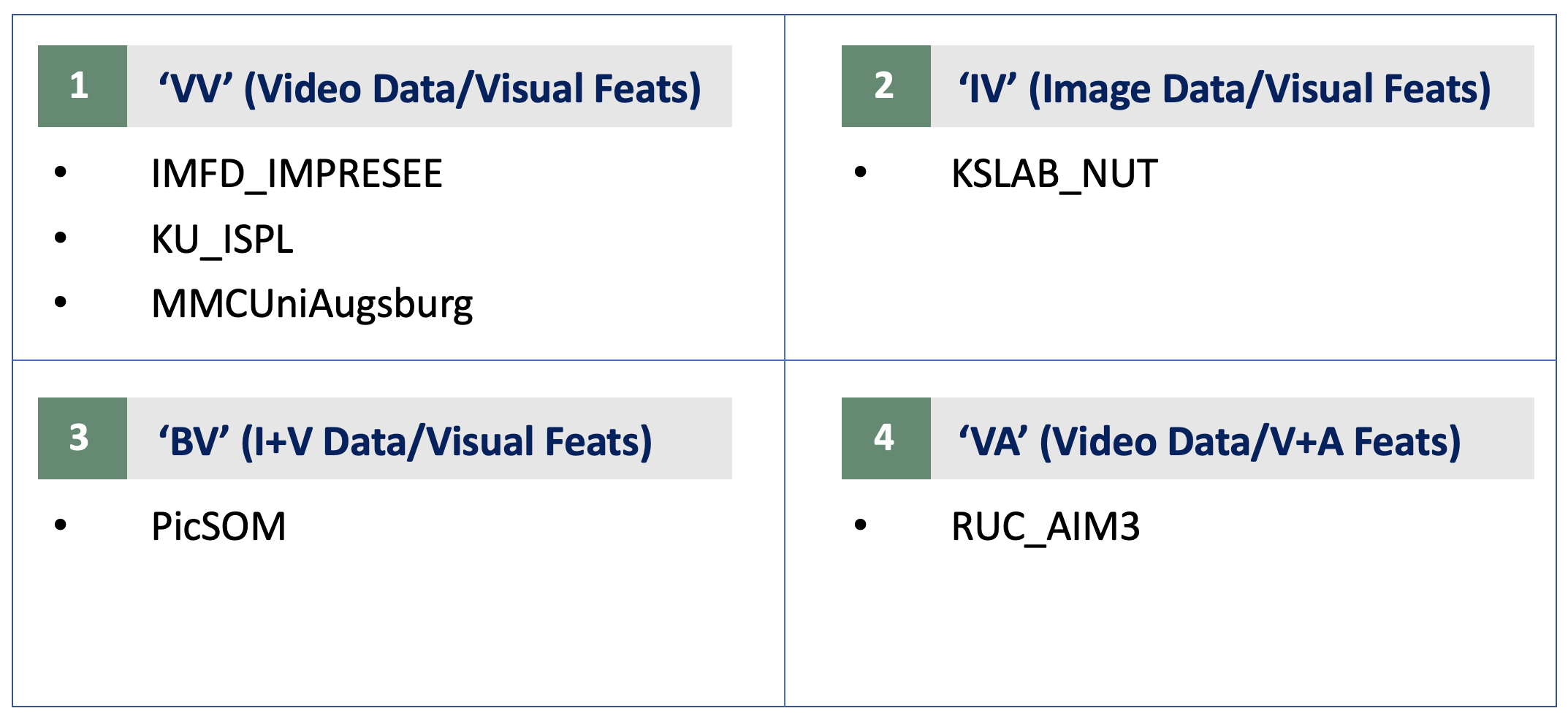}
\caption{VTT: Run types for description generation submissions.}
\label{fig:vtt.run.types}
\end{figure}

Systems were required to specify the run types based on the types of training data and features used. 

The list of training data types is as follows:
\begin{itemize}
    \item `I': Training using image captioning datasets only.
    \item `V': Training using video captioning datasets only.
    \item `B': Training using both image and video captioning datasets.
\end{itemize}

The feature types can be one of the following:
\begin{itemize}
    \item `V': Only visual features are used.
    \item `A': Both audio and visual features are used.
\end{itemize}

Figure~\ref{fig:vtt.run.types} shows the run types submitted by the teams for the description generation subtask. Only RUC\_AIM3 submitted runs for the matching and ranking subtask, and the run type was `VA'. 

\begin{figure}[htbp]
\centering
\includegraphics[width=1.0\linewidth]{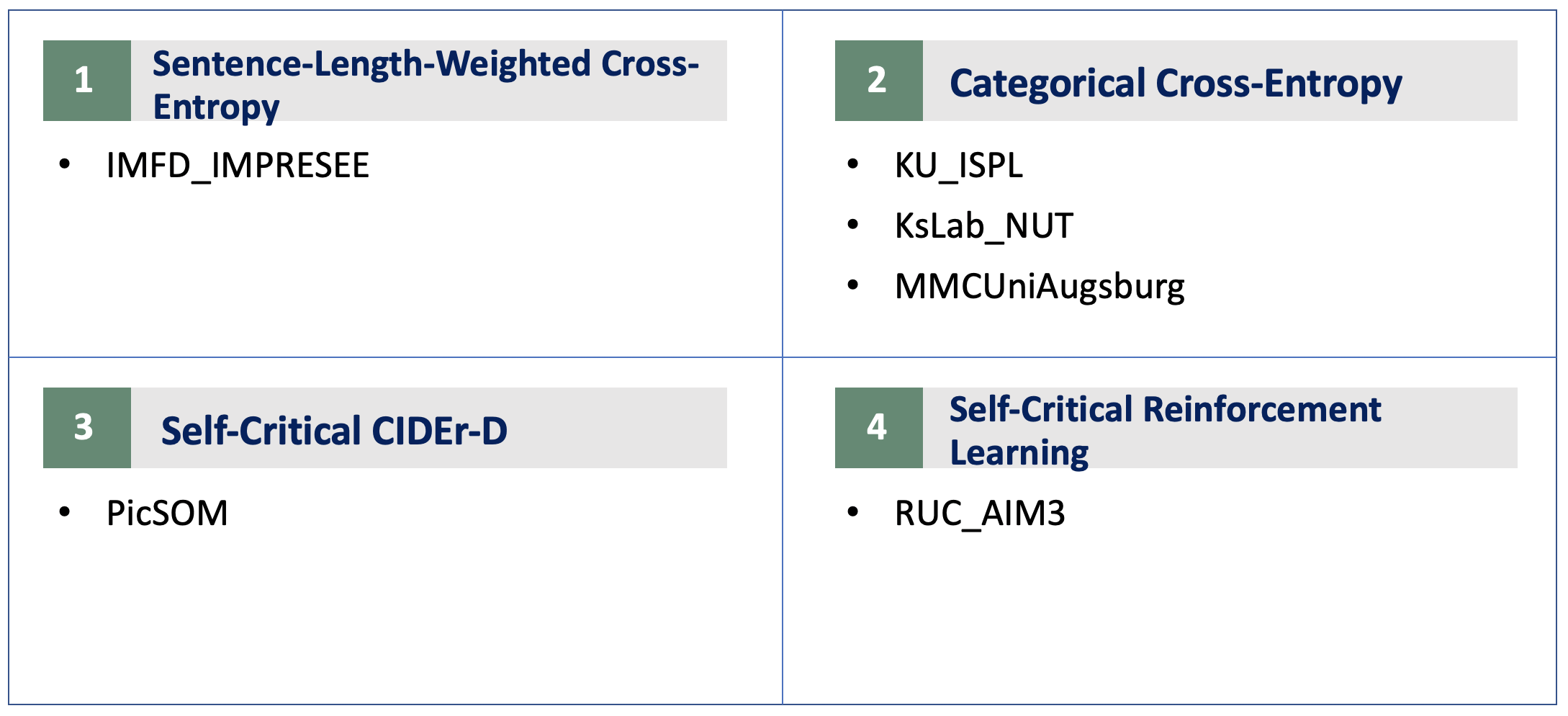}
\caption{VTT: Loss functions for description generation submissions.}
\label{fig:vtt.loss.functions}
\end{figure}

Teams were also asked to specify the loss function used for their runs, and Figure~\ref{fig:vtt.loss.functions} shows the loss functions used by the teams.

\subsubsection{Evaluation and Metrics}

The matching and ranking subtask scoring was done automatically against the ground truth using mean inverted rank at which the annotated item is found. The description generation subtask scoring was done automatically using a number of metrics. We also used a human evaluation metric on selected runs to compare with the automatic metrics. 

METEOR (Metric for Evaluation of Translation with Explicit ORdering) \cite{banerjee2005meteor} and BLEU (BiLingual Evaluation Understudy) \cite{papineni2002bleu} are standard metrics in machine translation (MT). BLEU was one of the first metrics to achieve a high correlation with human judgments of quality. It is known to perform poorly if it is used to evaluate the quality of individual sentence variations rather than sentence variations at a corpus level. In the VTT task the videos are independent and there is no corpus to work from. Thus, our expectations are lowered when it comes to evaluation by BLEU.  METEOR is based on the harmonic mean of unigram or n-gram precision and recall in terms of overlap between two input sentences. It redresses some of the shortfalls of BLEU such as better matching synonyms and stemming, though the two measures seem to be used together in evaluating MT.

The CIDEr (Consensus-based Image Description Evaluation) metric \cite{vedantam2015cider} is borrowed from image captioning. It computes TF-IDF (term frequency inverse document frequency) for each n-gram to give a sentence similarity score. The CIDEr metric has been reported to show high agreement with consensus as assessed by humans. We also report scores using CIDEr-D, which is a modification of CIDEr to prevent ``gaming the system''. 

The SPICE (Semantic Propositional Image Caption Evaluation) metric \cite{spice2016} is another metric that has gained popularity in image captioning evaluation. The metric uses scene graph similarity between generated captions and the ground truth instead of n-grams.

The STS (Semantic Textual Similarity) metric \cite{han2013umbc} was also applied to the results, as in the previous years of this task. This metric measures how semantically similar the submitted description is to one of the ground truth descriptions.

In addition to automatic metrics, the description  generation task includes human evaluation of the quality of automatically generated captions.
Recent developments in Machine Translation evaluation have seen the emergence of DA (Direct Assessment), a method shown to produce highly reliable human evaluation results for MT and Natural Language Generation \cite{DA,msr20}. 
DA now constitutes the official method of ranking in main MT benchmark evaluations \cite{WMT17,barrault-EtAl:2020:WMT1}. With respect to DA for evaluation of video captions (as opposed to MT output), human assessors are presented with a video and a single caption. After watching the video,  assessors rate how well the caption describes what took place in the video on a 0--100 rating scale \cite{graham2018evaluation}. Large numbers of ratings are collected for captions, before ratings are combined into an overall average system rating (ranging from 0 to 100\,\%). Human assessors are recruited via Amazon's Mechanical Turk (AMT) \footnote{\url{http://www.mturk.com}}, with quality control measures applied to filter out or downgrade the weightings from workers unable to demonstrate the ability to rate good captions higher than lower quality captions. This is achieved by deliberately ``polluting'' some of the manual (and correct) captions with linguistic substitutions to generate captions whose semantics are questionable. For instance, we might substitute a noun for another noun and turn the manual caption ``A man and a woman are dancing on a table" into ``A {\em horse} and a woman are dancing on a table'', where ``horse'' has been substituted for ``man''.  We expect such automatically-polluted captions to be rated poorly and when an AMT worker correctly does this, the ratings for that worker are improved.

DA was first used as an evaluation metric in TRECVID 2017. This metric has been used every year since then to rate each team's primary run, as well as 4 human systems.

\subsubsection{Overview of Approaches}
For detailed information about the approaches and results for individual teams' performance and runs, we refer the reader to the site reports \cite{tv20pubs} in the online workshop notebook proceedings. Here we present a high-level overview of the different systems. 


\paragraph{\textbf{Description Generation}}

RUC\_AIM3 outperformed the other systems on all metrics. They modeled semantic information from both temporal and spatial dimensions. Scene-level and object-level captions were generated, which were then combined via hybrid reranking. They used the TGIF, TRECVID-VTT 16-18, MSR-VTT, VATEX datasets for training, and the TRECVID-VTT 19 dataset for validation.

PicSOM also scored high consistently on all the metrics. They experimented with MS-COCO, TGIF and VATEX for training, and dropped the MSR-VTT and MSVD as they did not find these datasets useful anymore. A stacked attention captioning model was used, which is based on the Transformer model. Interestingly, they found that the addition of the VATEX dataset led to more improvement in results as compared to the new model.

IMFD\_IMPRESEE used an encoder-decoder model while focusing on syntactic representation learning to produce sentences with precise semantics and syntax. They used the TRECVID-VTT and MSR-VTT datasets to train, while using the VATEX dataset for one run.

KsLab\_NUT focused on reducing the processing time. They extracted 5 keyframes from videos --- the first and last frames, and 3 frames with largest changes in features --- and used an encoder-decoder method to caption each frame. These captions were then aggregated using extractive methods such as BERTSUM and LexRank. They used MS-COCO dataset for training.

KU\_ISPL submitted 3 different methods. Their baseline method used SA-LSTM, while they connected a Transformer and LSTM for the other runs. They only used TRECVID-VTT for training.

MMCUniAugsburg used a model based on Transformer architecture and modified it to take videos as input by adding an image embedding layer and positional encoding. They used Auto-captions on GIF, TRECVID-VTT, and MSR-VTT datasets to train, but found that using AC-GIF actually resulted in lower scores. 

\paragraph{\textbf{Matching and Ranking}}

RUC\_AIM3 was the only team that submitted runs for this subtask. They used a global matching model and a fine-grained matching model and found that combining the two gave the best results.

\subsubsection{Results}

\paragraph{\textbf{Description Generation}}

\begin{figure}[htbp]
  \centering
  \includegraphics[width=1.0\linewidth]{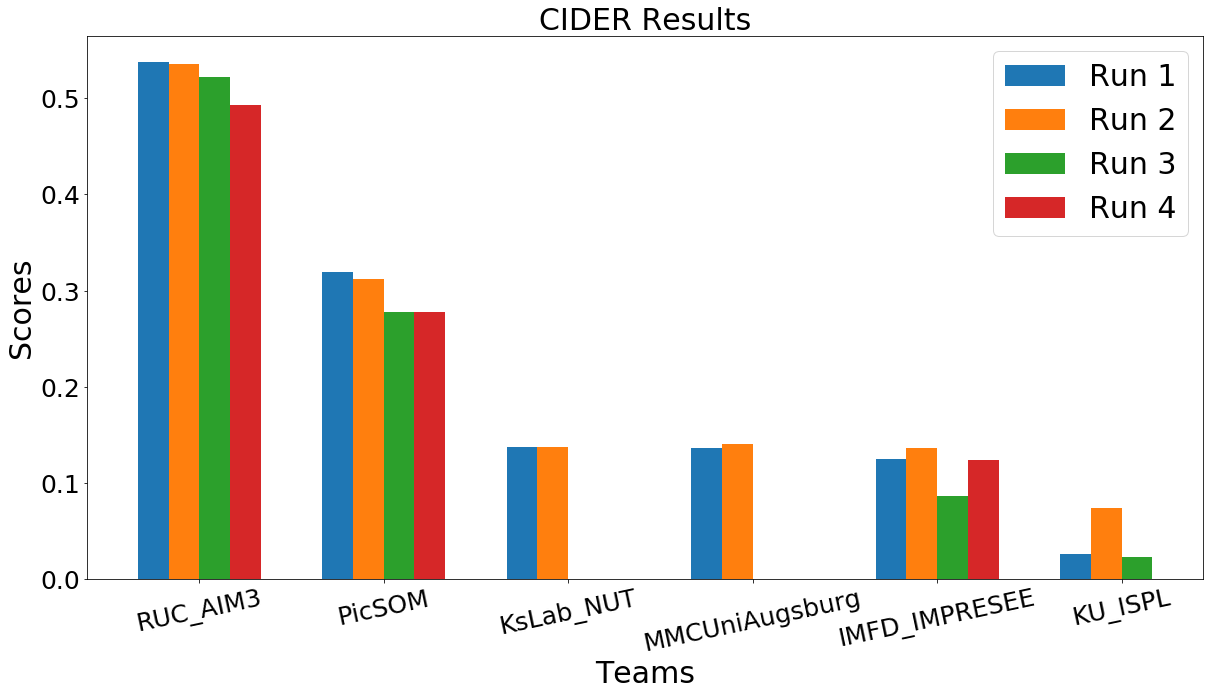}
  \caption{VTT: Comparison of all runs using the CIDEr metric.}
  \label{fig:vtt.cider.results}
\end{figure}

\begin{figure}[htbp]
  \centering
  \includegraphics[width=1.0\linewidth]{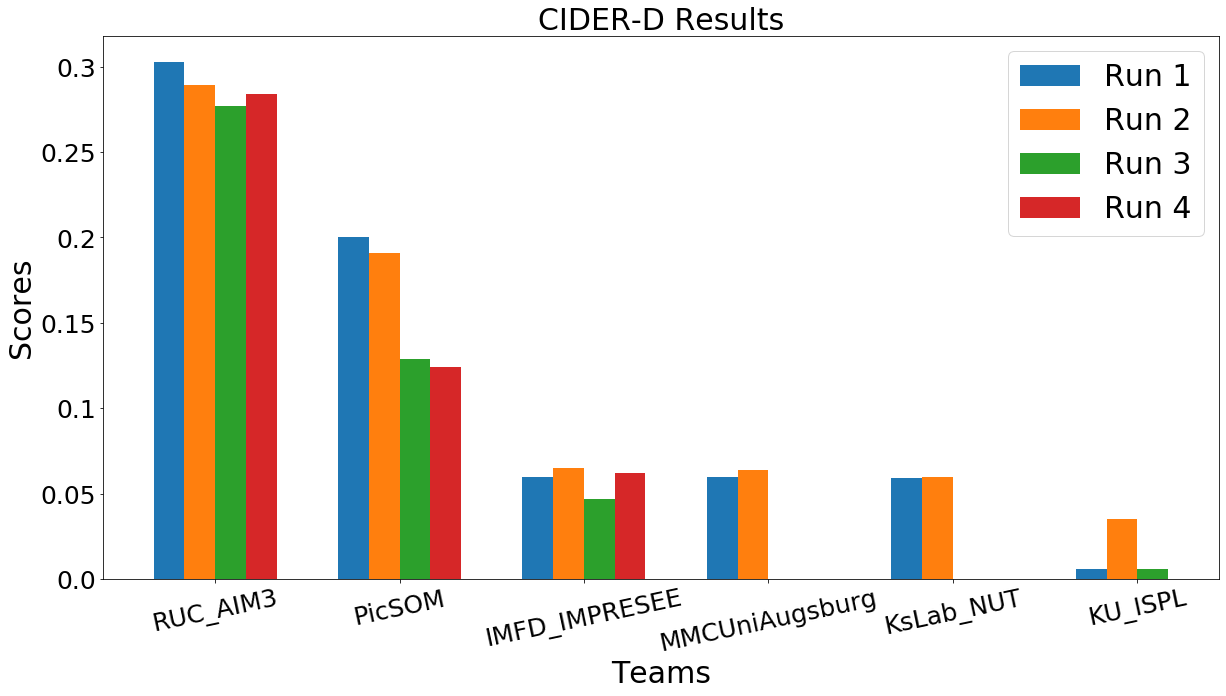}
  \caption{VTT: Comparison of all runs using the CIDEr-D metric.}
  \label{fig:vtt.ciderd.results}
\end{figure}

\begin{figure}[htbp]
  \centering
  \includegraphics[width=1.0\linewidth]{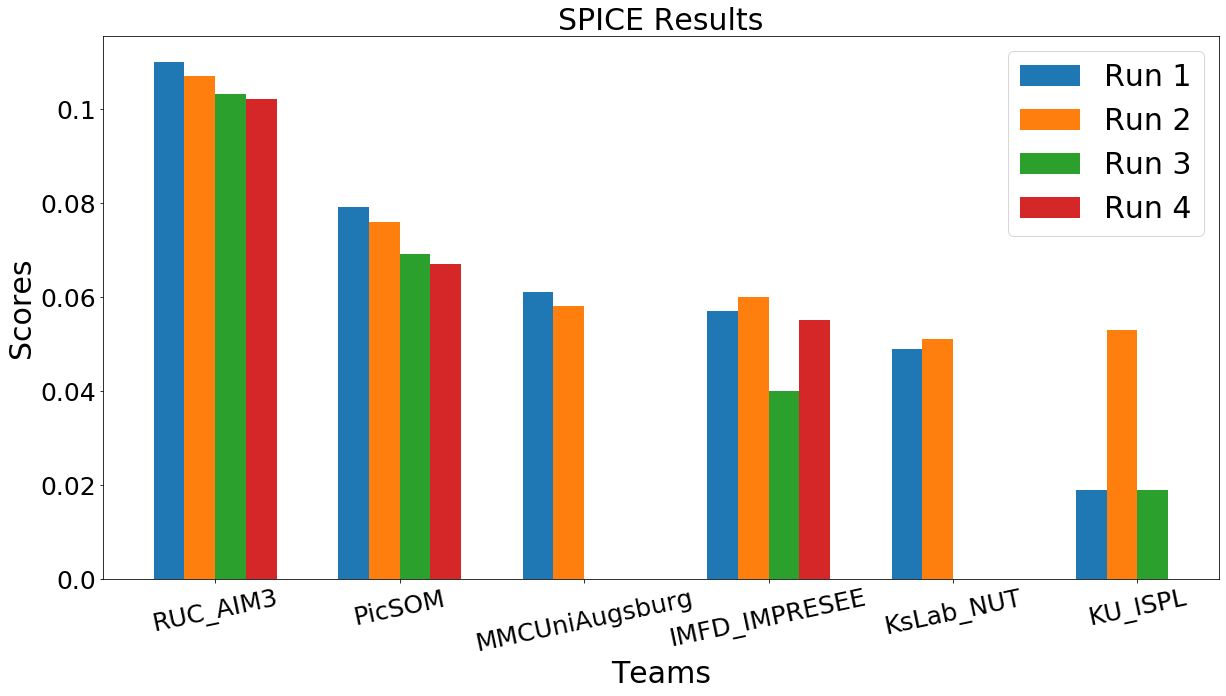}
  \caption{VTT: Comparison of all runs using the SPICE metric.}
  \label{fig:vtt.spice.results}
\end{figure}

\begin{figure}[htbp]
  \centering
  \includegraphics[width=1.0\linewidth]{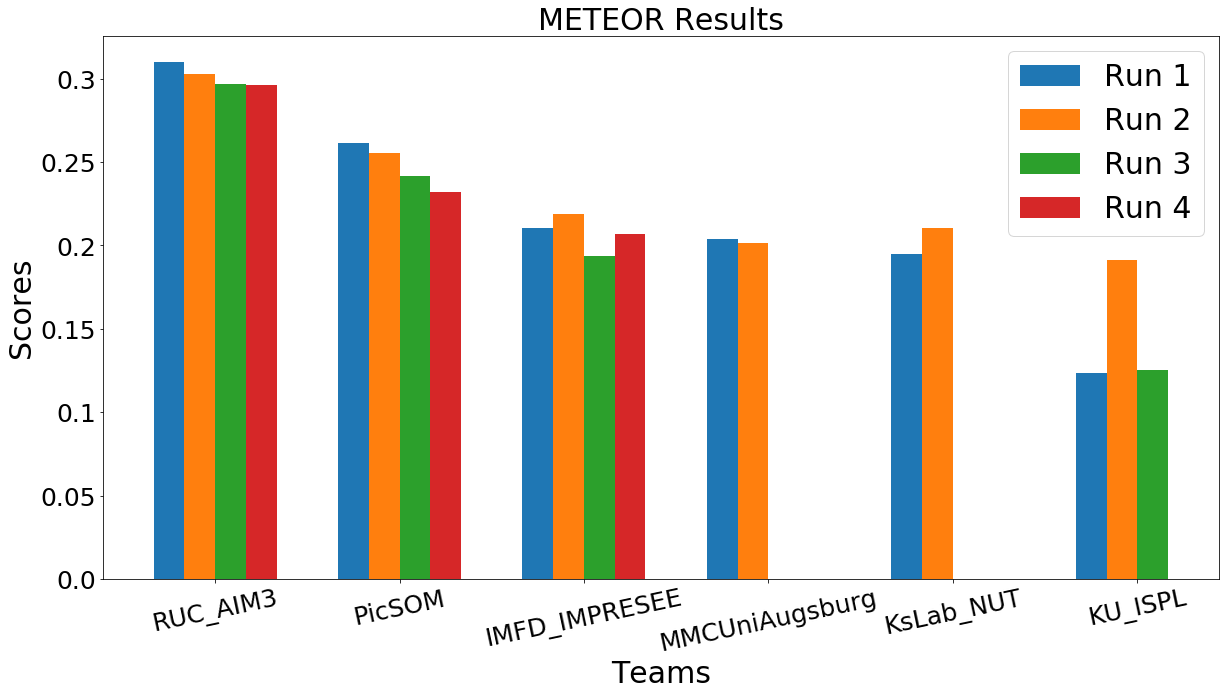}
  \caption{VTT: Comparison of all runs using the METEOR metric.}
  \label{fig:vtt.meteor.results}
\end{figure}

\begin{figure}[htbp]
  \centering
  \includegraphics[width=1.0\linewidth]{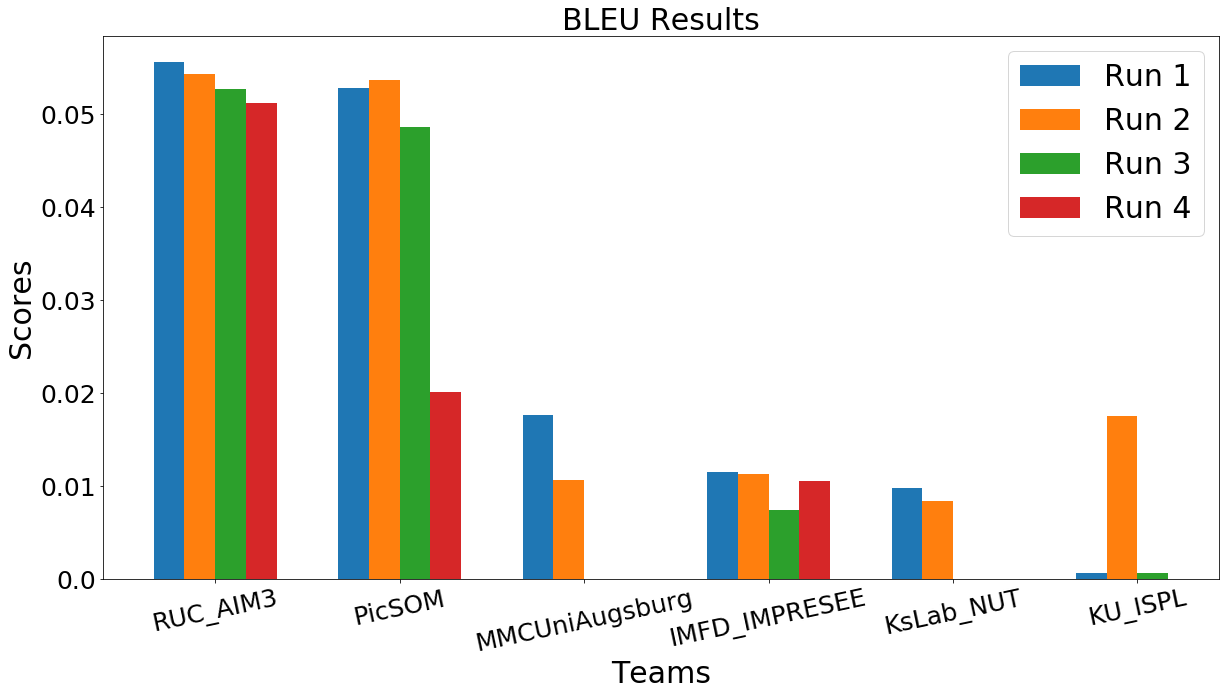}
  \caption{VTT: Comparison of all runs using the BLEU metric.}
  \label{fig:vtt.bleu.results}
\end{figure}

\begin{figure}[htbp]
  \centering
  \includegraphics[width=1.0\linewidth]{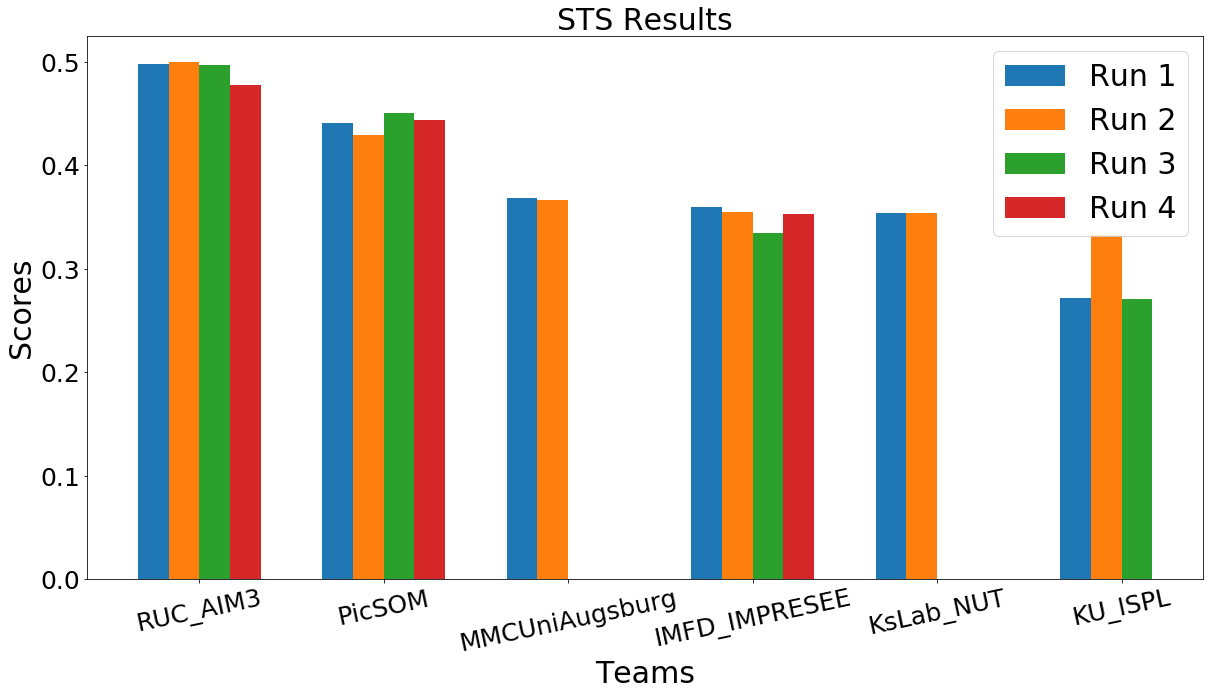}
  \caption{VTT: Comparison of all runs using the STS metric.}
  \label{fig:vtt.sts.results}
\end{figure}

\begin{table*}
\centering
\begin{tabular}{lrrrrrr}
\toprule
{} &  CIDER &  CIDER-D &  SPICE &  METEOR &  BLEU &    STS \\
\midrule
CIDER   &        1.000 &          0.992 &        0.959 &         0.948 &       0.911 &  0.961 \\
CIDER-D &        0.992 &          1.000 &        0.953 &         0.945 &       0.929 &  0.942 \\
SPICE   &        0.959 &          0.953 &        1.000 &         0.986 &       0.889 &  0.963 \\
METEOR  &        0.948 &          0.945 &        0.986 &         1.000 &       0.893 &  0.969 \\
BLEU   &        0.911 &          0.929 &        0.889 &         0.893 &       1.000 &  0.914 \\
STS           &        0.961 &          0.942 &        0.963 &         0.969 &       0.914 &  1.000 \\
\bottomrule
\end{tabular}
\caption{VTT: Correlation between overall run scores for automatic metrics.}
\label{tab:vtt.auto.metric.corr.run}
\end{table*}

\begin{table*}
\centering
\begin{tabular}{lrrrrrr}
\toprule
{} &  CIDEr &  CIDEr-D &  SPICE &  METEOR &   BLEU &    STS \\
\midrule
CIDEr  &  1.000 &   0.908 &  0.600 &   0.652 &  0.508 &  0.622 \\
CIDEr-D &  0.908 &   1.000 &  0.588 &   0.654 &  0.524 &  0.535 \\
SPICE  &  0.600 &   0.588 &  1.000 &   0.690 &  0.543 &  0.637 \\
METEOR &  0.652 &   0.654 &  0.690 &   1.000 &  0.562 &  0.682 \\
BLEU   &  0.508 &   0.524 &  0.543 &   0.562 &  1.000 &  0.458 \\
STS    &  0.622 &   0.535 &  0.637 &   0.682 &  0.458 &  1.000 \\
\bottomrule
\end{tabular}
\caption{VTT: Correlation between individual description scores for automatic metrics.}
\label{tab:vtt.auto.metric.corr.desc}
\end{table*}

\begin{table*}
\centering
\begin{tabular}{lrrrrrrr}
\toprule
{} &  CIDER &  CIDER-D &  SPICE &  METEOR &  BLEU &    STS &   DA\_Z \\
\midrule
CIDER   &        1.000 &          0.994 &        0.977 &         0.992 &       0.907 &  0.991 &  0.989 \\
CIDER-D &        0.994 &          1.000 &        0.971 &         0.999 &       0.942 &  0.994 &  0.978 \\
SPICE   &        0.977 &          0.971 &        1.000 &         0.976 &       0.918 &  0.976 &  0.947 \\
METEOR  &        0.992 &          0.999 &        0.976 &         1.000 &       0.945 &  0.991 &  0.970 \\
BLEU    &        0.907 &          0.942 &        0.918 &         0.945 &       1.000 &  0.935 &  0.866 \\
STS           &        0.991 &          0.994 &        0.976 &         0.991 &       0.935 &  1.000 &  0.984 \\
DA\_Z          &        0.989 &          0.978 &        0.947 &         0.970 &       0.866 &  0.984 &  1.000 \\
\bottomrule
\end{tabular}
\caption{VTT: Correlation between overall run scores for the primary runs.}
\label{tab:vtt.da.metric.corr}
\end{table*}

The description generation subtask scoring was done using popular automatic metrics that compare the system generation captions with ground truth captions as provided by assessors. We also continued the use of Direct Assessment, which was introduced in TRECVID 2017, to compare the submitted runs. 

The metric score for each run is calculated as the average of the metric scores for all the descriptions within that run.  
Figure~\ref{fig:vtt.cider.results} shows the performance comparison of all teams using the CIDEr metric. All runs submitted by each team are shown in the graph. Figure~\ref{fig:vtt.ciderd.results} shows the scores for the CIDEr-D metric, which is a modification of CIDEr. Figure~\ref{fig:vtt.spice.results} shows the SPICE metric scores. Figures~\ref{fig:vtt.meteor.results} and~\ref{fig:vtt.bleu.results} show the scores for METEOR and BLEU metrics respectively. The STS metric allows comparison between two sentences. For this reason, the captions are compared to a single ground truth description at a time, resulting in 5 STS scores. We will report the average of these scores as the STS score, and Figure~\ref{fig:vtt.sts.results} shows how the runs compare on this metric. 

Table~\ref{tab:vtt.auto.metric.corr.run} shows the correlation between the different metric scores for all the runs. The metrics correlate very well, which shows that they agree on the overall scoring of the runs. BLEU has comparatively weaker correlation with the other metrics, with some correlation scores being slightly less than 0.9. However, if we look at the description level metric scores, as shown in Table~\ref{tab:vtt.auto.metric.corr.desc}, we find that the metrics do not correlate as well. CIDEr and CIDEr-D correlate very well since they are based on the same method. However, the correlation scores between all other metrics range between 0.45 and 0.7. This shows that while the metrics agree on the big picture, the correlation is not as strong when it comes to the individual descriptions.

\begin{figure}[htbp]
  \centering
  \includegraphics[width=1.0\linewidth]{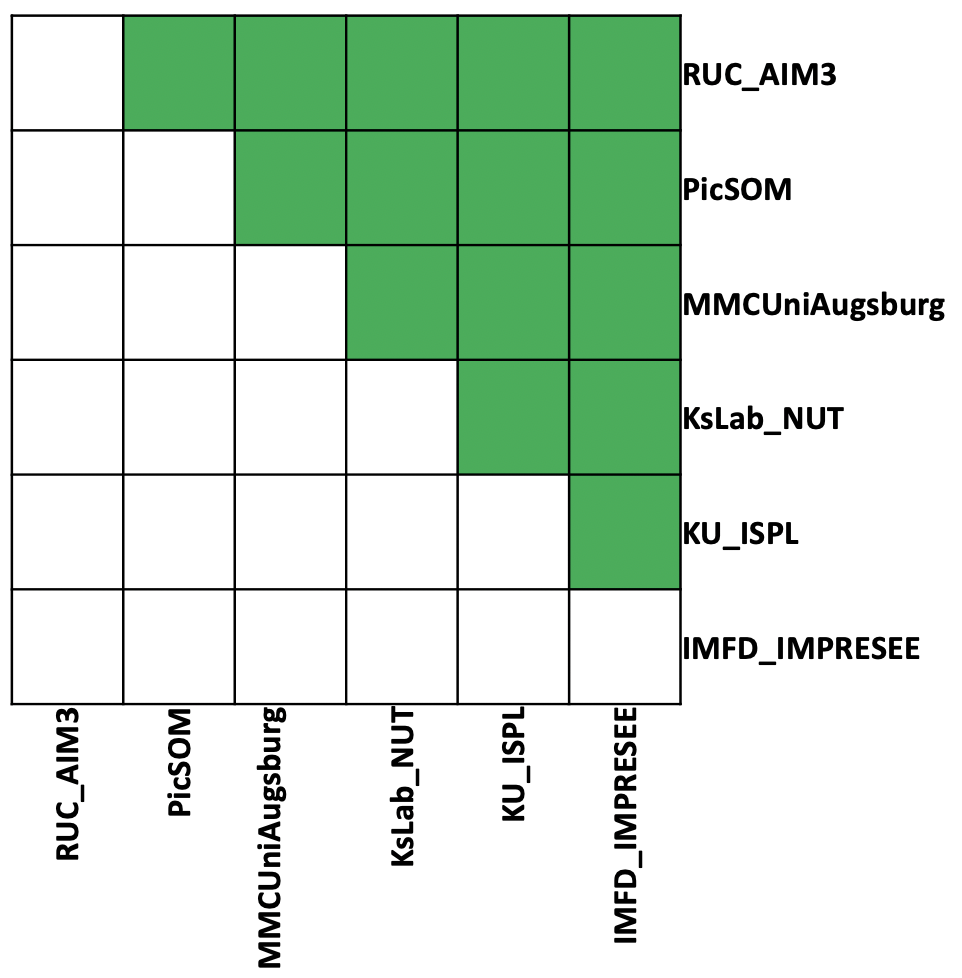}
  \caption{VTT: Comparison of the primary runs of each team with respect to the CIDEr score. Green squares indicate a significantly better result for the row over the column. }
  \label{fig:vtt.cider.significance}
\end{figure}

\begin{figure}[htbp]
  \centering
  \includegraphics[width=1.0\linewidth]{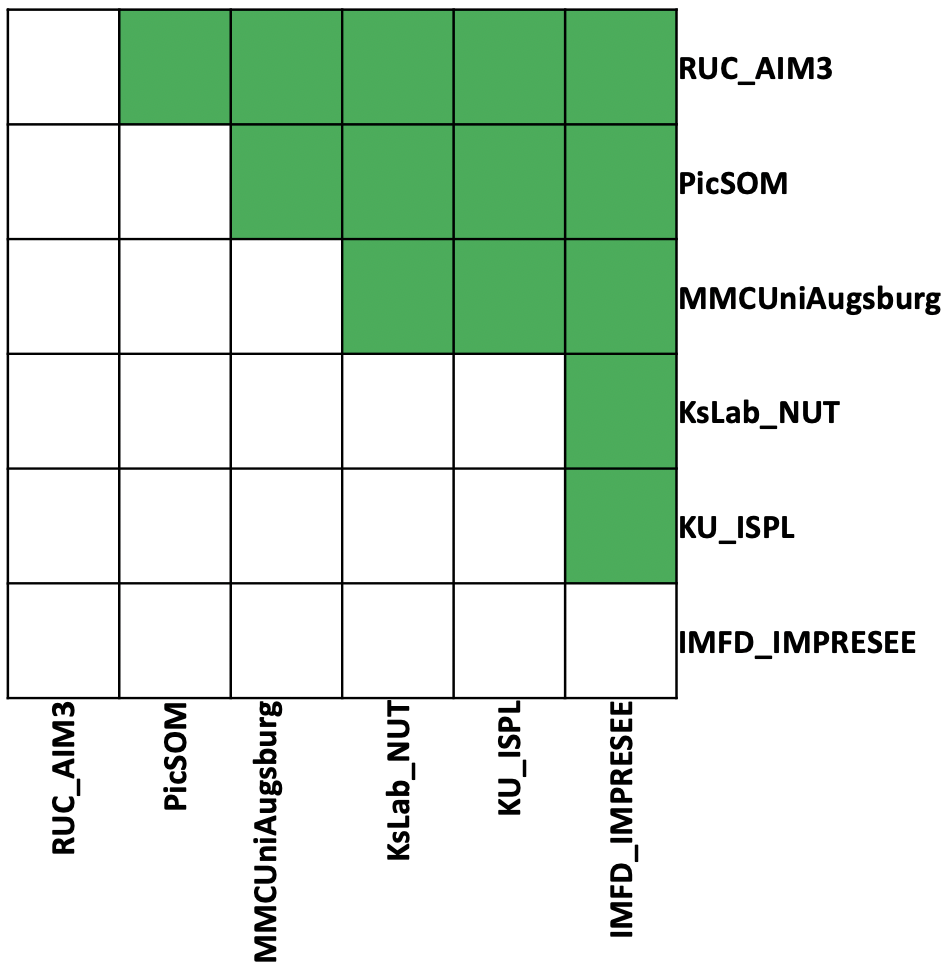}
  \caption{VTT: Comparison of the primary runs of each team with respect to the SPICE score. Green squares indicate a significantly better result for the row over the column. }
  \label{fig:vtt.spice.significance}
\end{figure}

\begin{figure}[htbp]
  \centering
  \includegraphics[width=1.0\linewidth]{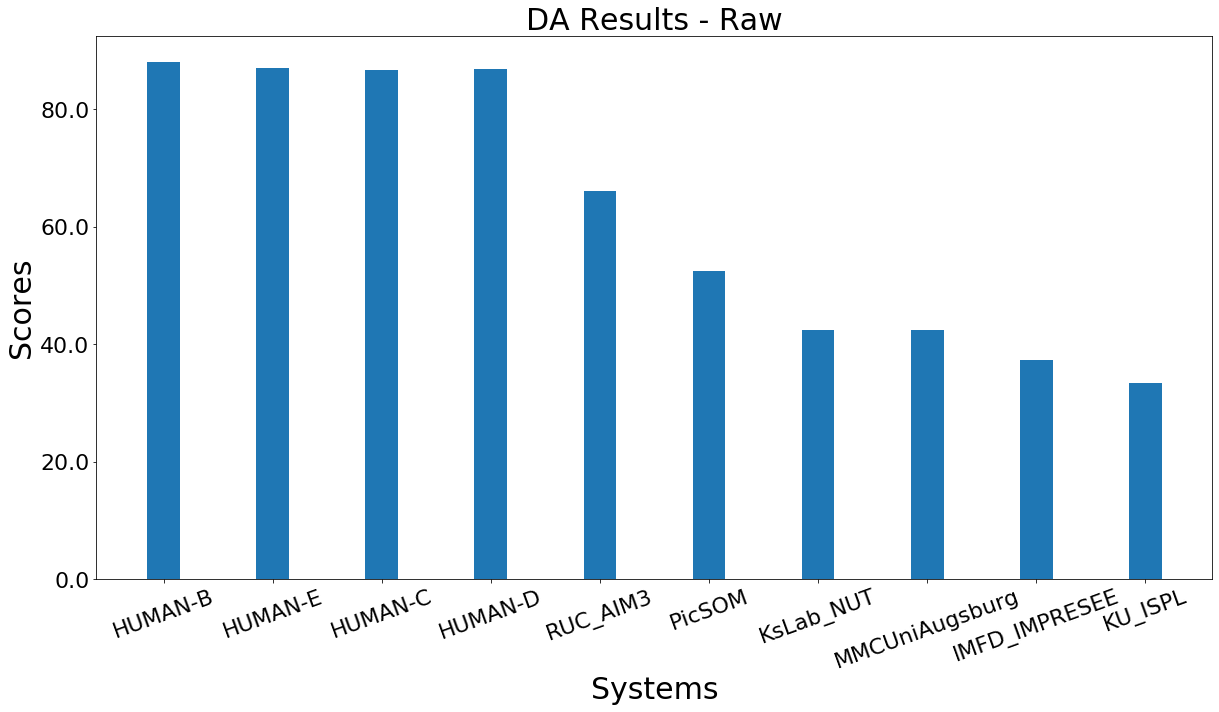}
  \caption{VTT: Average DA score for each system. The systems compared are the primary runs submitted, along with 4 manually generated system labeled as HUMAN\_B to HUMAN\_E.}
  \label{fig:vtt.da.raw.results}
\end{figure}

\begin{figure}[htbp]
  \centering
  \includegraphics[width=1.0\linewidth]{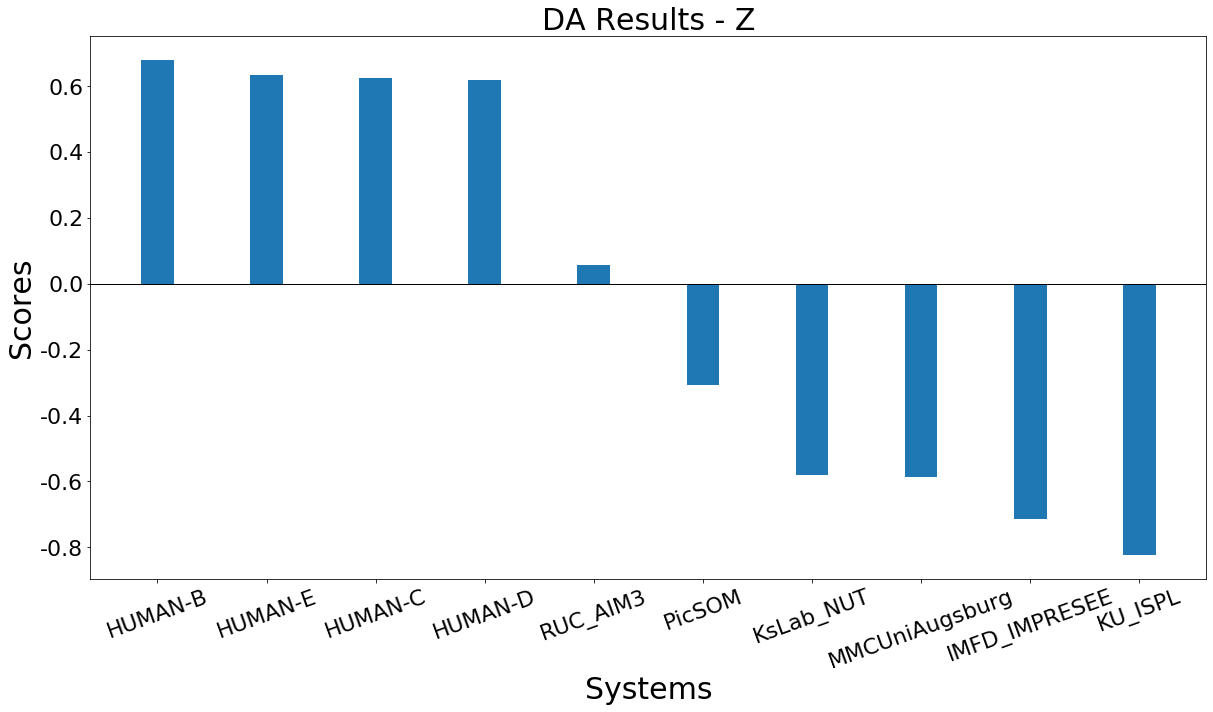}
  \caption{VTT: Average DA score per system after standardization per individual worker's mean and standard deviation score.}
  \label{fig:vtt.da.z.results}
\end{figure}

\begin{figure}[htbp]
  \centering
  \includegraphics[width=1.0\linewidth]{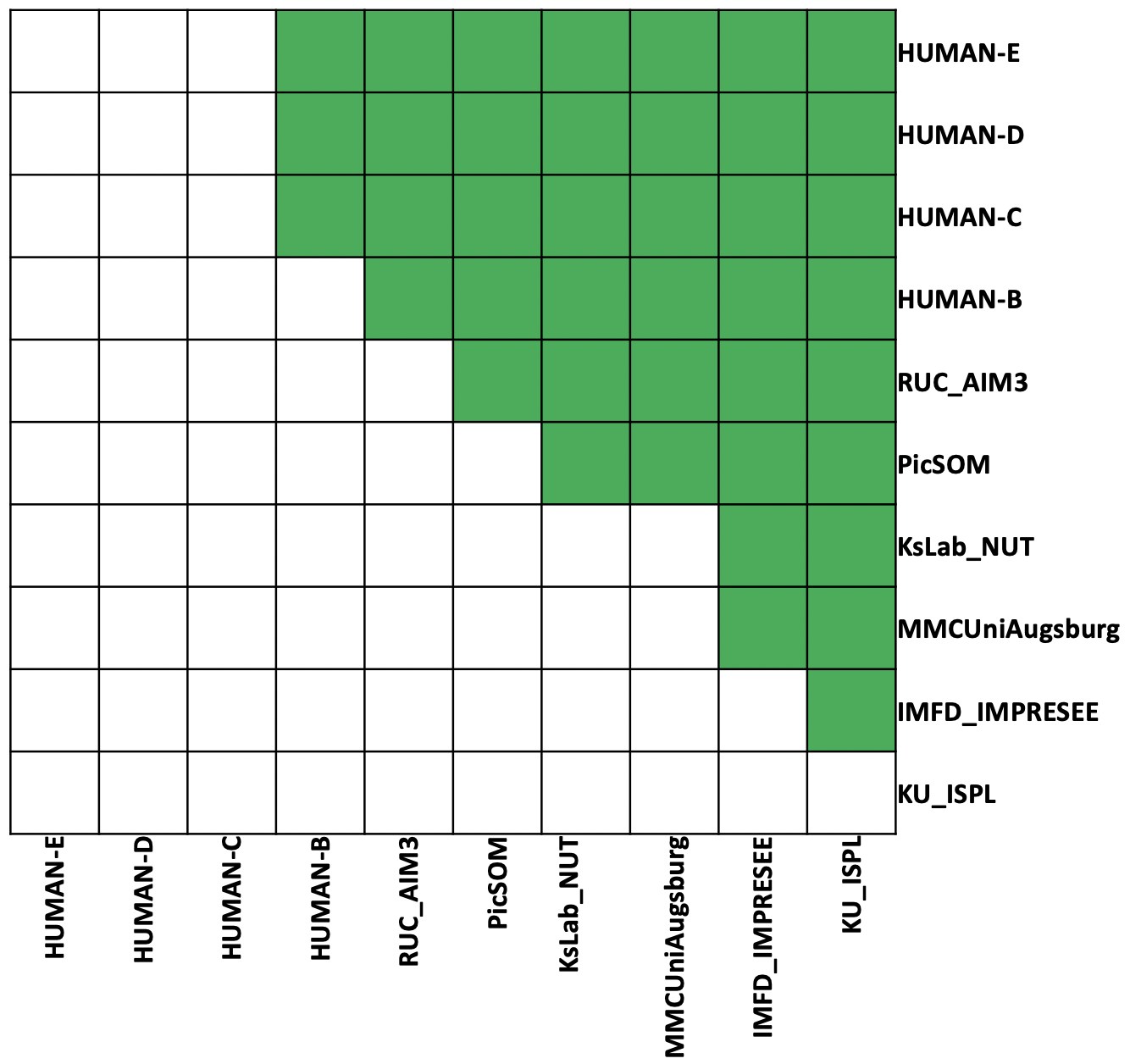}
  \caption{VTT: Comparison of the primary runs of each team with respect to the DA score. The 'HUMAN' systems are ground truth captions. Green squares indicate a significantly better result for the row over the column. }
  \label{fig:vtt.da.significance}
\end{figure}

Figure~\ref{fig:vtt.cider.significance} shows how the systems compare according to the CIDEr metric. The green squares indicate that the system in the row is significantly better (p \textless 0.05) than the system in the column. Figure~\ref{fig:vtt.spice.significance} shows the same comparison for the SPICE metric. It can be seen that for both metrics, RUC\_AIM3 is significantly better than the other metrics.



Figure~\ref{fig:vtt.da.raw.results} shows the average DA score [$0 - 100$] for each system. The score is micro-averaged per caption, and then averaged over all videos. Figure~\ref{fig:vtt.da.z.results} shows the average DA score per system after it is standardized per individual AMT worker's mean and standard deviation score. The HUMAN systems represent manual captions provided by assessors. As expected, captions written by assessors outperform the automatic systems. Figure~\ref{fig:vtt.da.significance} shows how the systems compare according to DA. The green squares indicate that the system in the row is significantly better than the system shown in the column (p \textless 0.05). The figure shows that no system reaches the level of the human performance. Among the systems, RUC\_AIM3 outperforms the rest and PicSOM is firmly in the second place.  

Table~\ref{tab:vtt.da.metric.corr} shows the correlation between different overall metric scores for the primary runs of all teams. The `DA\_Z' metric is the score generated by humans. The score correlates very well with the other metrics. As noted previously, BLEU has the weakest correlation with all the other metrics, including DA. 

\begin{figure}[htbp]
  \centering
  \includegraphics[width=1.0\linewidth]{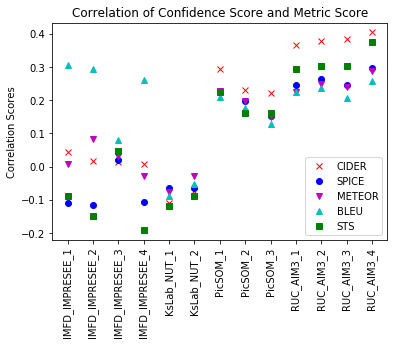}
  \caption{VTT: Correlation of system reported sentence confidence scores and the various metric scores. IMFD\_IMPRESEE and KsLab\_NUT do not correlate well with the metrics. However, the two top performing teams, RUC\_AIM3 and PicSOM, show a better correlation with most metrics.}
  \label{fig:vtt.conf.metric.corr}
\end{figure}

Teams were asked to provide a confidence score for each generated sentence. We expected these confidence scores to have a positive correlation with the metric scores. Figure~\ref{fig:vtt.conf.metric.corr} shows the correlation of the sentence confidence scores reported by the systems and the various metric scores. IMFD\_IMPRESEE shows a correlation of 0.3 with BLEU, but not with the other metrics. KsLab\_NUT does not show any correlation. On the other hand, the two top performing teams, RUC\_AIM3 and PicSOM, show a better correlation with most metrics, especially CIDEr. 

\begin{table*}
\centering
\begin{tabular}{llll}
\toprule
{} &               Run & Avg Length & Avg Length (Unique)\\
\midrule
0  &         KU\_ISPL\_1 &    6.08235 &         5.57529 \\
1  &         KU\_ISPL\_3 &    6.32235 &         5.69176 \\
2  &          PicSOM\_4 &    9.57882 &         8.03353 \\
3  &          PicSOM\_3 &    10.3335 &         9.25941 \\
4  &       KsLab\_NUT\_1 &    10.9329 &            8.67 \\
5  &       KsLab\_NUT\_2 &    11.4212 &         9.45059 \\
6  &         KU\_ISPL\_2 &    12.3659 &         10.8347 \\
7  &  MMCUniAugsburg\_2 &    12.5259 &         10.0029 \\
8  &        RUC\_AIM3\_3 &    12.9176 &         10.4318 \\
9  &          PicSOM\_2 &    13.4447 &           11.55 \\
10 &        RUC\_AIM3\_2 &    13.4447 &         10.7047 \\
11 &          PicSOM\_1 &    13.9618 &         11.8335 \\
12 &        RUC\_AIM3\_4 &    14.6888 &         11.4735 \\
13 &        RUC\_AIM3\_1 &    14.6971 &         11.2924 \\
14 &   IMFD\_IMPRESEE\_3 &      14.73 &         11.5288 \\
15 &   IMFD\_IMPRESEE\_4 &    17.3076 &         12.8582 \\
16 &   IMFD\_IMPRESEE\_2 &    18.6694 &         12.6165 \\
17 &   IMFD\_IMPRESEE\_1 &    18.9147 &         13.5635 \\
18 &  MMCUniAugsburg\_1 &    19.0447 &         12.0982 \\
\bottomrule
\end{tabular}
\caption{VTT: The table shows the average length of sentences for each run. The Avg Length (Unique) column shows the average length when only unique words are counted. }
\label{tab:vtt.run.avg.length}
\end{table*}

\begin{figure}[htbp]
  \centering
  \includegraphics[width=1.0\linewidth]{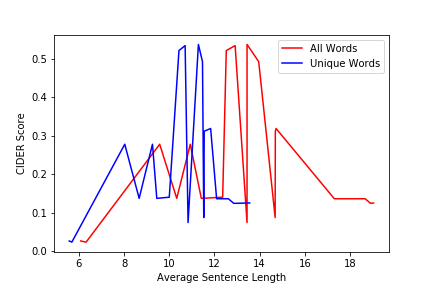}
  \caption{VTT: Correlation of average sentence length and the CIDEr score.}
  \label{fig:vtt.length.v.ciderscore}
\end{figure}

Table~\ref{tab:vtt.run.avg.length} shows the average length of sentences for each run. The table also shows the average length when only unique words are counted. These lengths can vary significantly for certain runs where words and phrases repeat often. Figure~\ref{fig:vtt.length.v.ciderscore} shows how the average sentence length correlates with the CIDEr score. There does not seem to be any pattern to suggest that longer sentences score better.

\begin{figure*}[htb]
  \centering
  \includegraphics[width=1\linewidth]{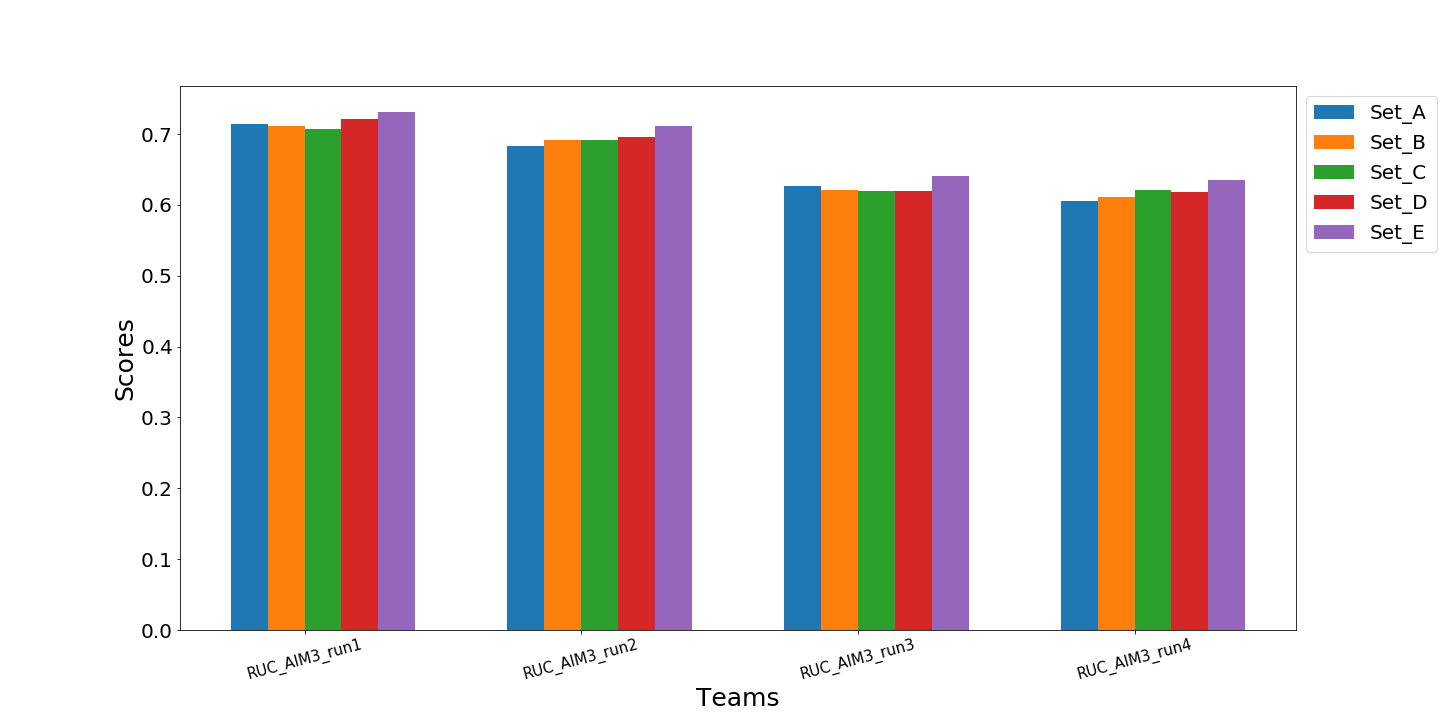}
  \caption{VTT: Matching and Ranking results across all runs for all sets.}
  \label{fig:vtt.match.rank.all.sets}
\end{figure*}


 
  




\paragraph{\textbf{Matching and Ranking}}

Only one team participated in this optional subtask and provided four runs. 
The results for the subtask are shown for each of the 5 sets (A-E) in Figure~\ref{fig:vtt.match.rank.all.sets}. The graph shows the mean inverted rank scores for all runs submitted for each of the description sets. The maximum mean inverted rank score is 0.73 and is comparable to last year's best performer. However, the testing dataset is different. 

This year, we included some fake sentences, i.e. sentences that did not correspond to any of the videos, in the descriptions sets provided for matching and ranking. We wanted to check how these sentences would be ranked by the systems. There were a total of 100 such fake sentences, so that each set had 20 fake sentences in addition to the actual ground truth of 1700 sentences. The fake sentences were of varying lengths and fell into two broad categories:
\begin{enumerate}
    \item Grammatically correct sentences that make no logical sense. These sentences were often so ridiculous in meaning that it is hard to imagine they could correspond to any real-world video. 
    \item Grammatically incorrect sentences. These were often random words strung together. 
\end{enumerate}

It was expected that these sentences would be ranked low by the systems. We found that their median rank was 461 out of 1720. 13.5\% of the fake sentences ranked in the top 100, while 53\% ranked in the top 500. These numbers were much higher than our expectation, and show that there is still a lot of room for improvement in bridging the video and language domains.



\subsubsection{Conclusion and Future Work}

The VTT task continues to have healthy participation. Given the challenging nature of the task, and the increasing interest in video captioning in the computer vision community, we hope to see improvements in performance. 

This year we used the new V3C2 dataset and plan to continue with this dataset for the next year. With increasing interest in video captioning, participants have a number of open datasets available to train their systems.

We are proposing some major changes for the VTT task next year:
\begin{enumerate}
    \item Introduction of a progress task: In a similar vein to AVS and INS, we will use a subset of the dataset to measure the progress of systems over a period of 3 years. We will withhold the ground truth for a selected number of videos in 2021. Then, systems will be able to compare their systems in 2022 and 2023 on these same videos as testing data to compare their progress. The progress subset will be selected to ensure that the videos are of appropriate difficulty level and are diverse.
    \item Removal of the matching and ranking subtask: The matching and ranking subtask seems to have reached the end of its usefulness. We will not be continuing it in the future.
    \item Introduction of new subtask: We intend to introduce a new 'Fill in the Blank' subtask. This is a variation of the common visual question answering (VQA). For this subtask, participants will be given a video and an incomplete sentence, and the systems will return the most suitable word(s) to complete the sentence. The goal is to test VTT systems to see how well they understand the video content and the textual representation. The task may be manually evaluated.
\end{enumerate}


\subsection{Activities in Extended Video}

This year we continued with the ActEV task with 35 target activities that we had started in 2018.  NIST TRECVID Activities in Extended Video (ActEV) series was initiated in 2018 to support the Intelligence Advanced Research Projects Activity (IARPA) Deep Intermodal Video Analytics (DIVA) Program.
The Activities in Extended Video (ActEV) evaluation series is designed to accelerate development of robust, multi-camera, automatic human activity detection systems for forensic and real-time alerting
applications. In this evaluation, an activity is defined as \say {one or more people performing a specified movement or interacting with an object or group of objects (including driving and flying)}, while an instance indicates an occurrence (time span of the start and end frames) associated with the activity.

\par ActEV began with the Summer 2018 Blind and Leaderboard evaluations and has currently progressed to the running of two concurrent evaluations: 1) the ActEV Sequestered Data
Leaderboard (ActEV SDL) based on the Multiview Extended Video (MEVA) dataset \cite{MEVAdata} with 37
activities. 2) The TRECVID 2020 ActEV TRECVID self-reported leaderboard based on
the VIRAT V1 and V2 datasets \cite{oh2011large} with 35 activities.

The  TRECVID 2018 ActEV (ActEV18) evaluated system detection performance on 12 activities for the self-reported evaluation and 19 activities for the leaderboard evaluation using the VIRAT V1 and V2 datasets \cite{TrecVIDActev18}. 
For the self-reported evaluation, the participants ran their software on their hardware and configurations and submitted the system outputs with the defined format to the NIST scoring server. For the leaderboard evaluation, the participants submitted their runnable systems to the NIST scoring server, which was independently evaluated on the sequestered data using the NIST hardware.
\par The ActEV18 evaluation addressed two different tasks: 1) identify a target activity along with the time span of the activity (AD: activity detection), 2) detect objects associated with the activity occurrence (AOD: activity and object detection). 
\par For the TRECVID 2019 ActEV (ActEV19) evaluation, we primarily focused on  18 activities and increased the number of instances for each activity. ActEV19 included the test set from both VIRAT V1 and V2 datasets and the systems were evaluated on the activity detection (AD) task only. 
\par The TRECVID 2020 ActEV (ActEV20) self-reported leaderboard is based on the VIRAT V1 and V2 datasets  with 35 activities with updated names to make it easier to use the MEVA dataset to train systems for TRECVID ActEV leaderboard.
\par
\begin{figure}[htb]
\begin{centering}
\includegraphics[width=3.16667in,height=2.10000in]{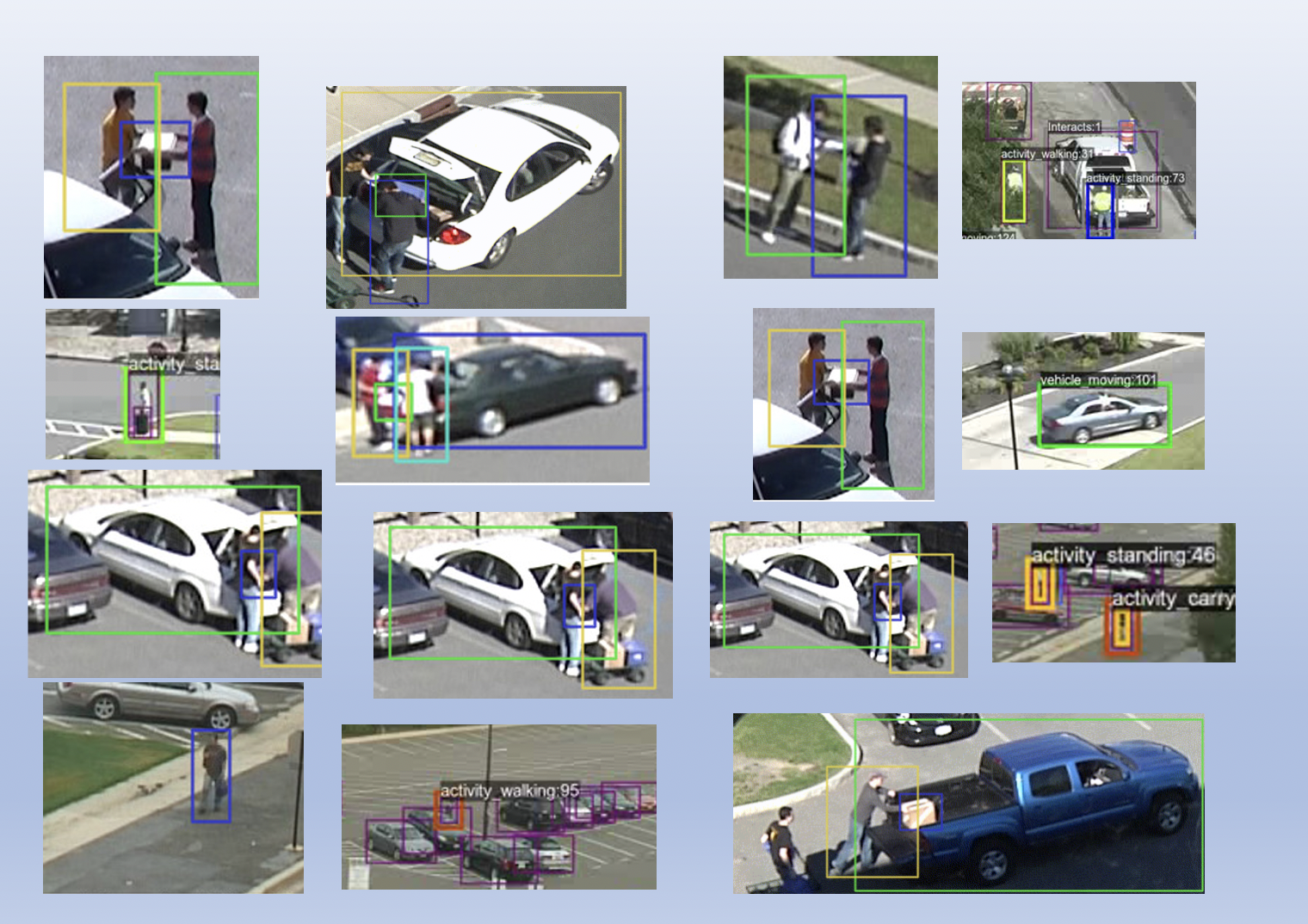}
\caption{Example of activities for ActEV series. IRB (Institutional Review Board): 00000755}
\label{fig_actev:7}
\end{centering}
\end{figure}

 Figure \ref{fig_actev:7} illustrates an example of representative activities that were used in the TRECVID 2020 ActEV. The evaluation primarily targeted on the forensic analysis that processes an entire corpus prior to returning a list of detected activity instances. A total of 7 different organizations participated in this year evaluation (ActEV20) and over 118 different runs were submitted.

\par In this section, we first discuss the task and datasets used and introduce a new metric to evaluate algorithm performance. In addition, we present the results for the TRECVID20 ActEV submissions and discuss observations and conclusions.

\subsubsection{Task and Dataset}
\par In the ActEV20 leaderboard evaluation, we addressed activity detection (AD) task for detecting and localizing activities; a systems was required to automatically detect and localize all instances of the activity. For a system-identified activity instance to be evaluated as correct, the type of activity should be correct, and the temporal overlap should fall within a minimal requirement. The ActEV20 was an open leaderboard evaluation. The challenge participants were required to run their systems locally and submit the outputs in a pre-specified format to the NIST scoring server. The systems were supposed to detect target activities that visibly occurred  in a single-camera video, as well as the frame span (the start and end frames) of the detected activity instance along with a confidence score indicating the likelihood of the presence of the activity within the frame boundaries.

\begin{table*}[htbp]
  \centering
  \caption{A list of activity names for TRECVID ActEV, for ActEV19  there were 18 activities and for ActEV20 there were 35 activities based on the VIRAT dataset and their associated number of instances for the training and validation sets are also listed.}
     \begin{tabular}{m{5.6 cm} | m{5.5 cm} | m{0.7 cm}| m{.85 cm} } 
    \hline
    VIRAT19 (18 Activities) &  VIRAT20 (35 Activities) & Train & Validate \\ \hline
   
    Closing & person\_closes\_facility\_or\_vehicle\_door & 141   & 130 \\
   
    Closing\_Trunk & person\_closes\_trunk & 21    & 31 \\
    
          x & vehicle\_drops\_off\_person & 0     & 4 \\
    
    Entering & person\_enters\_facility\_or\_vehicle & 77    & 70 \\
    
    Exiting & person\_exits\_facility\_or\_vehicle & 66    & 72 \\
    
          x & person\_interacts\_object & 101   & 88 \\
   
    Loading & person\_loads\_vehicle & 38    & 38 \\
    
    Open\_Trunk & person\_opens\_trunk & 22    & 35 \\
    
    Opening & person\_opens\_facility\_or\_vehicle\_door & 137   & 128 \\
    
          x & person\_person\_interaction & 11    & 17 \\
    
          x & person\_pickups\_object & 19    & 12 \\
    
          x & vehicle\_picks\_up\_person & 9     & 5 \\
    
    Pull  & person\_pulls\_object & 23    & 43 \\
    
          & person\_pushs\_object & 4     & 6 \\
    
    Riding & person\_rides\_bicycle & 22    & 21 \\
    
          x & person\_sets\_down\_object & 12    & 11 \\
    
    Talking & person\_talks\_to\_person & 41    & 67 \\
   
    Transport\_HeavyCarry & person\_carries\_heavy\_object & 31    & 44 \\
    
    Unloading & person\_unloads\_vehicle & 32    & 44 \\
   
    activity\_carrying & person\_carries\_object & 237   & 364 \\
   
          x & person\_crouches & 1     & 9 \\
    
          x & person\_gestures & 82    & 148 \\
    
          x & person\_runs & 14    & 18 \\
    
          x & person\_sits & 21    & 11 \\
    
          x & person\_stands & 398   & 819 \\
    
          x & person\_walks & 761   & 901 \\
   
    specialized\_talking\_phone & person\_talks\_on\_phone & 17    & 16 \\
    
    specialized\_texting\_phone & person\_texts\_on\_phone & 5     & 20 \\
  
          x & person\_uses\_tool & 7     & 11 \\
    
          x & vehicle\_moves & 718   & 797 \\
   
          x & vehicle\_starts & 259   & 239 \\
 
          x & vehicle\_stops & 292   & 295 \\

    vehicle\_turning\_left & vehicle\_turns\_left & 152   & 176 \\
    
    vehicle\_turning\_right & vehicle\_turns\_right & 149   & 172 \\
    
    vehicle\_u\_turn & vehicle\_makes\_u\_turn & 9     & 13 \\
    
  \end{tabular}%
  \label{table_actev:1}%
\end{table*}%

\par For this evaluation, we used 35 activities from the VIRAT dataset and the activities were annotated by Kitware, Inc. The VIRAT dataset consists of 29 hours of video  and more than 43 activity types. A total of 10  hours of video were annotated for the test set across 35 activities. The detailed definition of each activity and evaluation requirements are described in the evaluation plan \cite{TrecVIDActev20}. 
\par Table \ref{table_actev:1} lists the number of instances for each activity for the training and validation sets. Due to ongoing evaluations, the information about the test sets are not included in the table. The frequency of instances are not balanced across activities, which may affect the system performance results. 

\subsubsection{Measures}

 Activity detection in extended video is not a discrete detection task unlike speaker recognition \cite{greenberg2020two} and fingerprint identification \cite{karu1996fingerprint},  it is a streaming detection task where multiple activity instances can overlap temporally or spatially and is similar to keyword spotting in audio \cite{le2014developing}. From a metrology perspective the difference between discrete and streaming detection tasks is that non-target trials (i.e., test probes not belonging to the class) are not countable for streaming detection because the number of unique temporal/spatial instances are near infinite.  To account for this difference, the ActEV evaluations used two methods to normalize the measured false alarm performance.  The first, \say{Rate of False Alarms}, is an instance-based false alarm measure that uses the number of video minutes as an estimate of the number of non-target trials as the false alarm denominator.  The second, \say{Time-based False Alarms}, is a time-based false alarm measure that used the sum of non-target time as the denominator.  The two variations correspond to two views concerning the impact false alarms  have on a user reviewing detection. The former is instance-based which implies the user effort would scale linearly with the detected instances and the latter time-based which implies the user effort would scale linearly with the duration of video reviewed.

The primary measure of performance for TRECVID ActEV20 is the normalized, partial Area Under the DET Curve ($nAUDC$) from 0 to a fixed, Time-based False Alarm ($T_{fa}$) nAUDC $T_{FA}$ value $a$ , denoted $nAUDC_a$, which is the same as the metric used for the TRECVID ActEV19 evaluation.  All ActEV performance measurements were on a per-activity basis and then performance was aggregated by averaging over activities.  While presence confidences scores where used to compute performance, cross-activity presence confidences score normalization was not required nor evaluated.

\begin{figure*}[htbp]
\begin{centering}
\includegraphics[width=5.56667in,height=1.2000in]{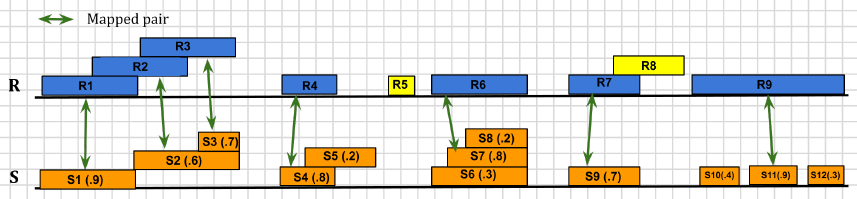}
\caption{Illustration of activity instance alignment and \(P_{miss}\) calculation (\(R\) is the reference instances and \(S\) is the system instances. In \(S\), the first number indicates instance id and the second indicates presence confidence score. For example, \(S1 (.9)\) represents the instance \(S1\) with corresponding confidence score (.9). Green arrows indicate aligned instances between \(R\) and \(S\))}
\label{fig_actev:1}
\end{centering}
\end{figure*}

For TRECVID ActEV18, the primary metric was instance-based measures for both missed detections and  false alarms (as illustrated in Figure \ref{fig_actev:1}. The metric evaluates how accurately a system detects instance occurrences of the activity. 

 As shown in Figure \ref{fig_actev:1}, the detection confusion matrix is calculated with alignment between reference and system output on the target activity instances; Correct Detection (\(CD\)) indicates that the reference and system output instances are correctly mapped (instances marked in blue). Missed Detection (\(MD\)) indicates that an instance in the reference has no correspondence in the system output (instances marked in yellow) while False Alarm (\(FA\)) indicates that an instance in the system output has no correspondence in the reference (instances marked in red). After calculating the confusion matrix, we summarize system performance: for each instance, a system output provides a confidence score that indicates how likely the instance is associated with the target activity. The confidence scores are not used as a decision threshold. Rather, a decision threshold is applied on the scores to determine the error counts ($N_{FA}$ and $N_{miss}$).
 \par In the ActEV20 evaluation (same as for ActEV19 evaluation), a probability of missed detections (\(P_{\text{miss}}\)) and a rate of false alarms (\(R_{\text{FA}})\) were used and computed at a given decision threshold:
\[P_{\text{miss}}(\tau)\  = \frac{N_{\text{MD}}(\tau)}{N_{\text{TrueInstance}}}\]
\[\text{R}_{\text{FA}}(\tau)\  = \frac{N_{\text{FA}}(\tau)}{\text{VideoDurInMinutes}}\]

\noindent where \(N_{\text{MD\ }}(\tau)\) is the number of missed detections at the threshold   
\(\tau\), \(N_{\text{FA}}(\tau)\) is the number of false alarms, and \emph{VideoDurInMinutes} is the video duration in minutes. \(N_{\text{TrueInstance}}\) is the number of reference instances annotated in the sequence per activity. Lastly, the Detection Error Tradeoff (DET) curve \cite{martinDET} is used to visualize system performance. For the TRECVID ActEV18 challenge, we evaluated algorithm performance for two operating points:
\(P_{\text{miss}}\text{\ at\ }R_{\text{FA}} = 0.15\) and
\(P_{\text{miss}}\text{\ at\ }R_{\text{FA}} = 1\).

To understand system performance better and to be more relevant to the user cases, for ActEV20, we used the normalized, partial area under the DET curve (\(nAUDC\)) from 0 to a fixed time-based false alarm (\(T_{fa}\)) to evaluate algorithm performance.
The partial area under DET curve is computed separately for each activity over all videos in the test collection and then is normalized to the range [0, 1] by dividing by the maximum partial area. \(nAUDC_a=0\) is a perfect score. The \(nAUDC_a\) is defined as:

\[nAUDC_{a} = \frac{1}{a}\int_{x=0}^{a} P_{miss}(x)dx,  x=T_{fa}\]

\noindent where \(x\) is integrated over the set of \(T_{fa}\) values. The instance-based probability of missed detections \(P_{miss}\) is defined as:

\[P_{miss} (x) = \frac{N_{md}(x)}{N_{TrueInstance}}\]

\noindent where \(N_{md}(x)\) is the number of missed detections at the presence confidence threshold that result in \(T_{fa}=x\) (see the below equation for the details). \(N_{TrueInstance}\) is the number of true instances in the sequence of reference.
\par
\begin{figure*}[htbp]
\begin{centering}
\includegraphics[width=5.56667in,height=2.70000in]{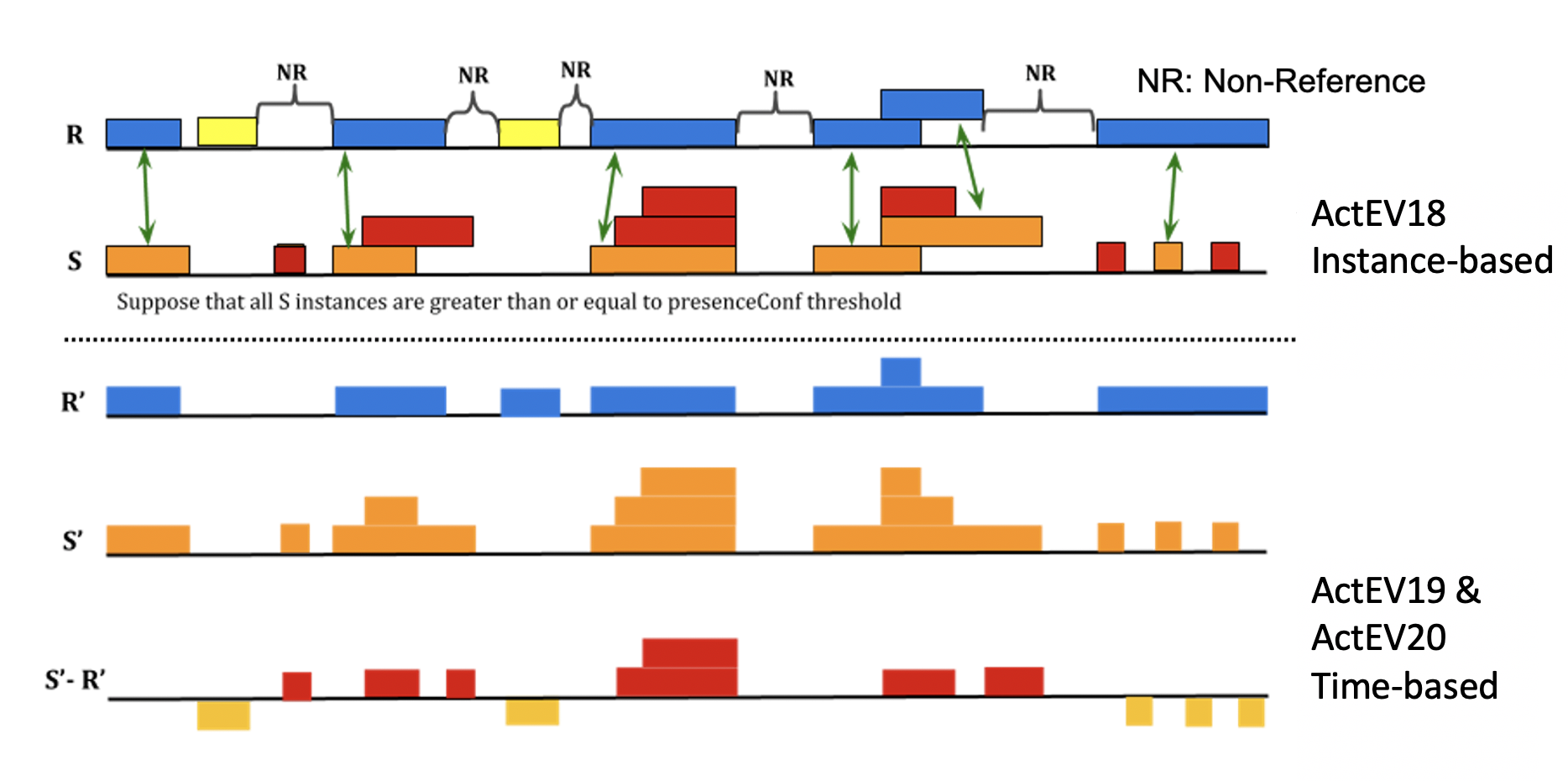}
\caption{Comparison of instance-based and time-based false alarms. \(R\) is the reference instances and \(S\) is the system instances. \(R^\prime\) is the histogram of the count of reference instances and \(S^\prime\) is the histogram of the count of system instances for the target activity. \(S\) shows a depiction of instance-based false alarms while \({S^\prime}-{R^\prime}\) illustrates time-based false alarms as marked in red.}
\label{fig_actev:2}
\end{centering}
\end{figure*}

 The time-based false alarm \(T_{fa}\) is defined as: 
\[T_{fa} = \frac{1}{NR} {\sum_{i=1}^{N_{frames}}} {\max(0, {S_i^\prime}-{R_i^\prime})}\]

\noindent where \(N_{frames}\) is the duration of the video and \(NR\) is the non-reference duration; the duration of the video without the target activity occurring. \({S_i^\prime}\) is the total count of system instances for frame \(i\) while \({R_i^\prime}\) is the total count of reference instances for frame \(i\). The detailed calculation of \(T_{fa}\) is illustrated in Figure \ref{fig_actev:2}.
\par The non-reference duration (NR) of the video where no target activities occur is computed by constructing a time signal composed of the complement of the union of the reference instances duration. \(R\) is the reference instances and \(S\) is the system instances. \(R^\prime\) is the histogram of the count of reference instances and \(S^\prime\) is the histogram of the count of system instances for the target activity. \(R^\prime\) and \(S^\prime\) both have \(N_{frames}\) bins, thus \(R_i^\prime\) is the value of the \(i^{th}\) bin \(R^\prime\) while \(S_i^\prime\) is the value of the \(i^{th}\) bin \(S^\prime\).
\(S^\prime\) is the total count of system instances in frame \(i\) and \(R^\prime\) is the total count of reference instances in frame \(i\).  
False alarm time is computed by summing over positive difference of \({S^\prime}-{R^\prime}\)(shown in red in Figure \ref{fig_actev:2}); that is the duration of falsely detected system instances. This value is normalized by the non-reference duration of the video to provide the \(T_fa\) value in Equation above.

\begin{figure}[hbt]
\begin{centering}
\includegraphics[width=3.21667in,height=1.50000in]{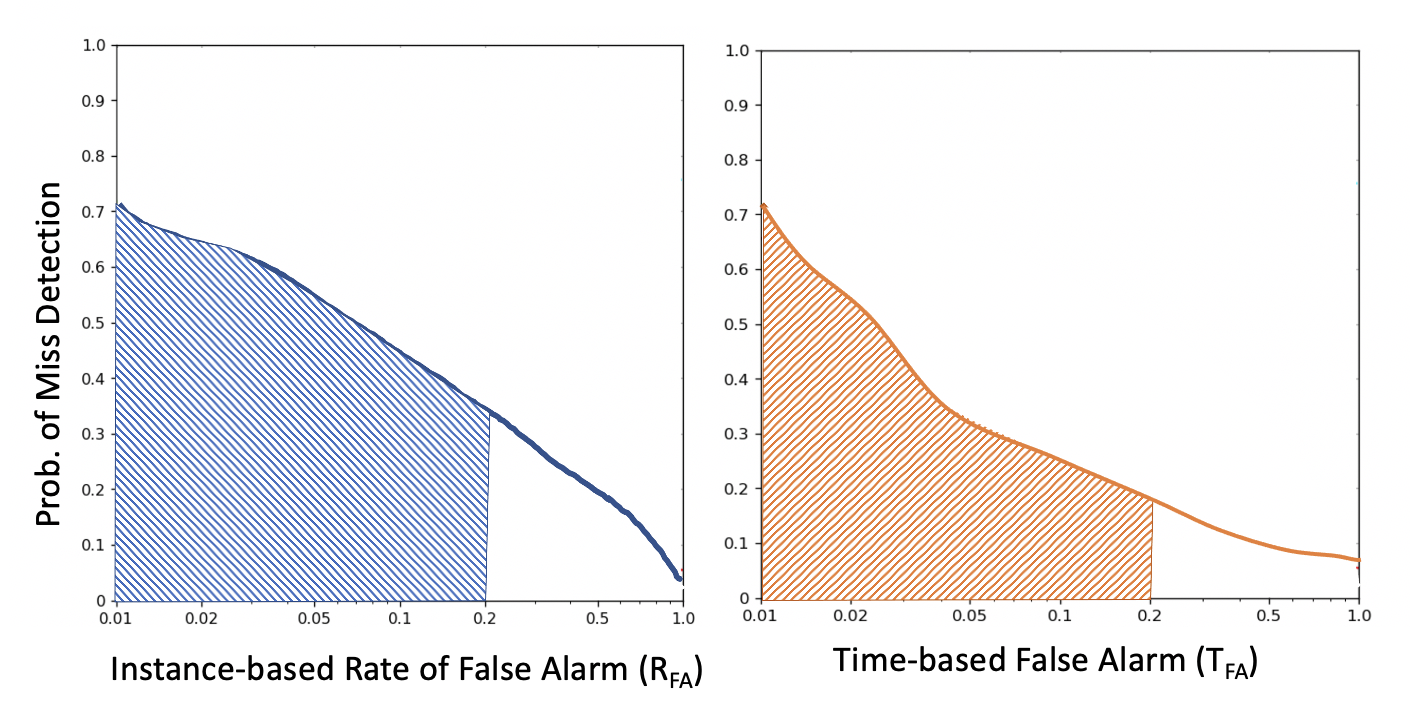}
\caption{Comparison of ActEV18 (\(R_{FA}\)) and ActEV20 (\(T_{FA}\)) measures using the Detection Error Tradeoff (DET) curves}
\label{fig_actev:3}
\end{centering}
\end{figure}

Figure \ref{fig_actev:3} visualizes the major differences between the ActEV18 and ActEV19/ActEV20 metrics. For the ActEV18 metric, we used Instance-based Rate of false alarms and system performance was evaluated at the specific operating point as illustrated in the left DET. For the ActEV19/ActEV20 metric, we used Time-based false alarms and calculated \(nAUDC\) from \(T_{FA}\) 0 to 0.2.

\subsubsection{ActEV Results}

A total of 7 teams from academia and industry from 4 countries participated in the ActEV20 evaluation. Each participant was allowed to submit multiple system outputs and a total of 118 submissions were received. Table \ref{table_actev:4}  lists the participating teams along with nAUDC scores for the best performing system per team. The best performance on activity detection is by INF-CMU at 42.3\% followed by BUPT-MCPRL at 55.5\%.

To be able to compare system performance with that from previous TRECVID ActEV evaluations, Table \ref{table_actev:5} presents results ordered by $nAUDC$ along with $mean\_P_{miss}@T_{FA}.15$, $mean\_w\_P_{miss}@R_{FA}.15$ and $mean\_P_{miss}@R_{FA}.15$.

\begin{table*}[htbp]
 
  \caption{Summary of participants information and their ($nAUDC$) values. Each team was allowed to have multiple submissions.}
  \label{table_actev:4} 
  \begin{centering}
   \begin{tabular}{ m{3cm} | m{10cm}| m{1.2cm} } 
    \hline
    Team  & Organization & $nAUDC$ \\
    \hline
    INF-CMU & Carnegie Mellon University, USA & 0.423 \\
    
    BUPT-MCPRL & Beijing University of Posts and Telecommunications, China & 0.555 \\
    
    UCF   & University of Central Florida, USA & 0.585 \\
    
    TokyoTech\_AIST & Tokyo Institute of Technology, Japan & 0.798 \\
    
    CERTH-ITI & Information Technologies Institute, Greece & 0.866 \\
    
    Team UEC & The University of Electro-Communications,  Japan & 0.952 \\
    
    Kindai\_Kobe & Kindai University and Osaka Gakuin University, Japan & 0.968 \\
    \hline
    \end{tabular}%
  \end{centering}
\end{table*}%

\begin{table*}[htbp]

  \centering
  \caption{The ranked list of system performance (ordered by $nAUDC$), we also present   $mean\_P_{miss}@T_{FA}.15$, $mean\_w\_P_{miss}@R_{FA}.15$ and $mean\_P_{miss}@R_{FA}.15$ }
 \label{table_actev:5}
    \begin{tabular}{l|r|r|r|r}
    \hline
Team  & $nAUDC$ & $mean\_P_{miss}@T_{FA}.15$ & $mean\_w\_P_{miss}@R_{FA}.15$ & $mean\_P_{miss}@R_{FA}.15$ \\
\hline

  INF-CMU & 0.423 & 0.332 & 0.810 & 0.805 \\
  BUPT-MCPRL & 0.555 & 0.488 & 0.845 & 0.846 \\
 UCF   & 0.585 & 0.547 & 0.835 & 0.834 \\
 TokyoTech\_AIST & 0.798 & 0.755 & 0.879 & 0.879 \\
  CERTH-ITI & 0.866 & 0.845 & 0.882 & 0.895 \\
  Team UEC & 0.952 & 0.953 & 0.983 & 0.987 \\
  Kindai\_Kobe & 0.968 & 0.964 & 0.957 & 0.969 \\
\hline
\end{tabular}%
 
\end{table*}%

\begin{figure}[htb]
\begin{centering}
\includegraphics[width=3.16667in,height=2.00000in]{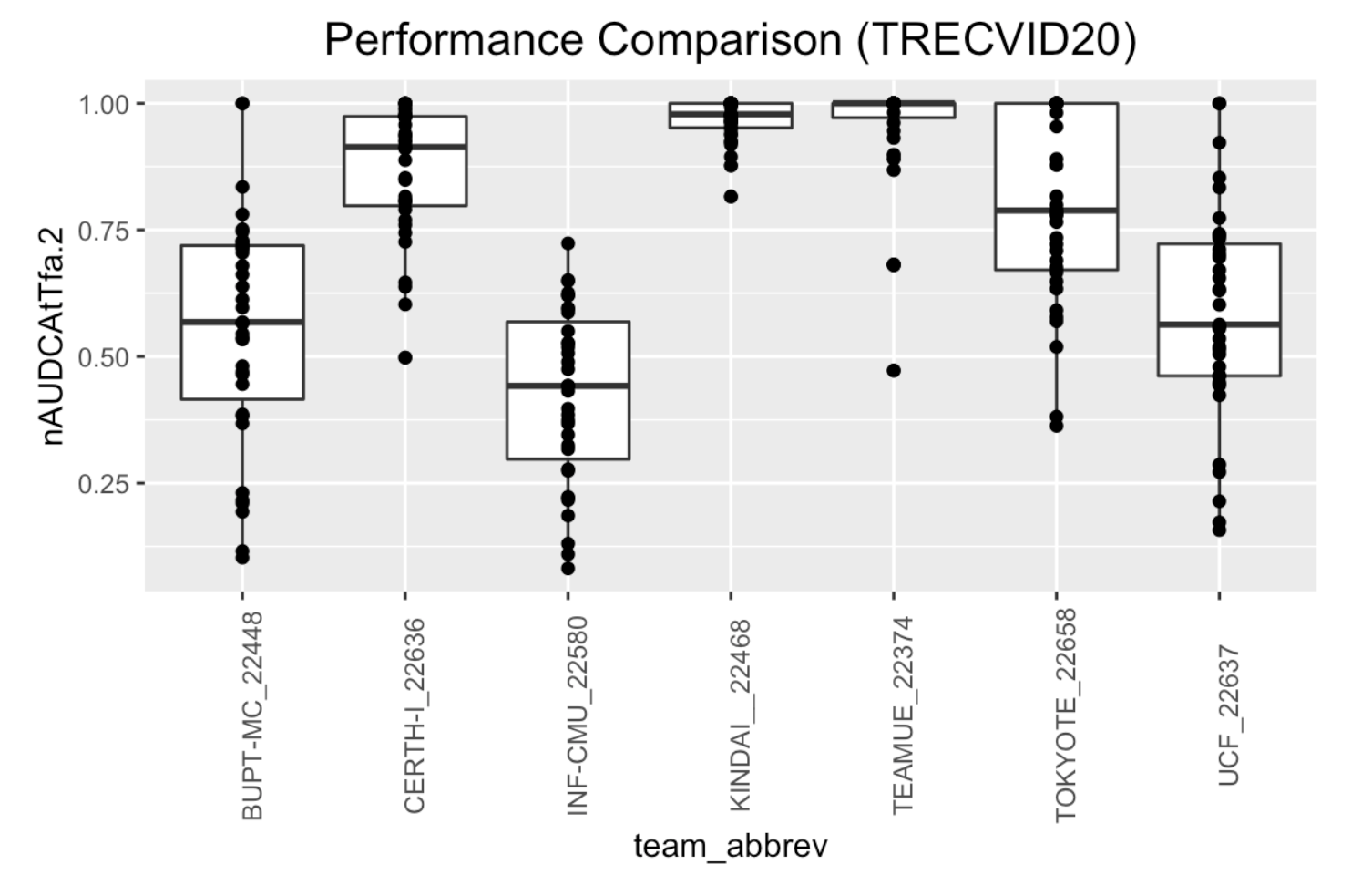}
\caption{The performance ranking of the 7 teams. The x-axis is the team name and the y-axis is $nAUDC$ value. A lower value is considered  a better performance.}
\label{fig_actev:16}
\end{centering}
\end{figure}

Figure \ref{fig_actev:16} shows the ranking of the 7 systems ordered by $nAUDC$.
The x-axis is the team name and the y-axis is the metric value $nAUDC$. Since this is error rate, a lower value indicates a better performance. 
The dots within the box plots represent the performance of the 35 different activities.
The black center bars indicates the mean values across 35 activities and the outer bars represent the standard deviation. We observe a large variance of $nAUDC$ for the 35 activities across systems chosen as the best performing systems per team.

\begin{figure}[htb]
\begin{centering}
\includegraphics[width=3.16667in,height=2.80000in]{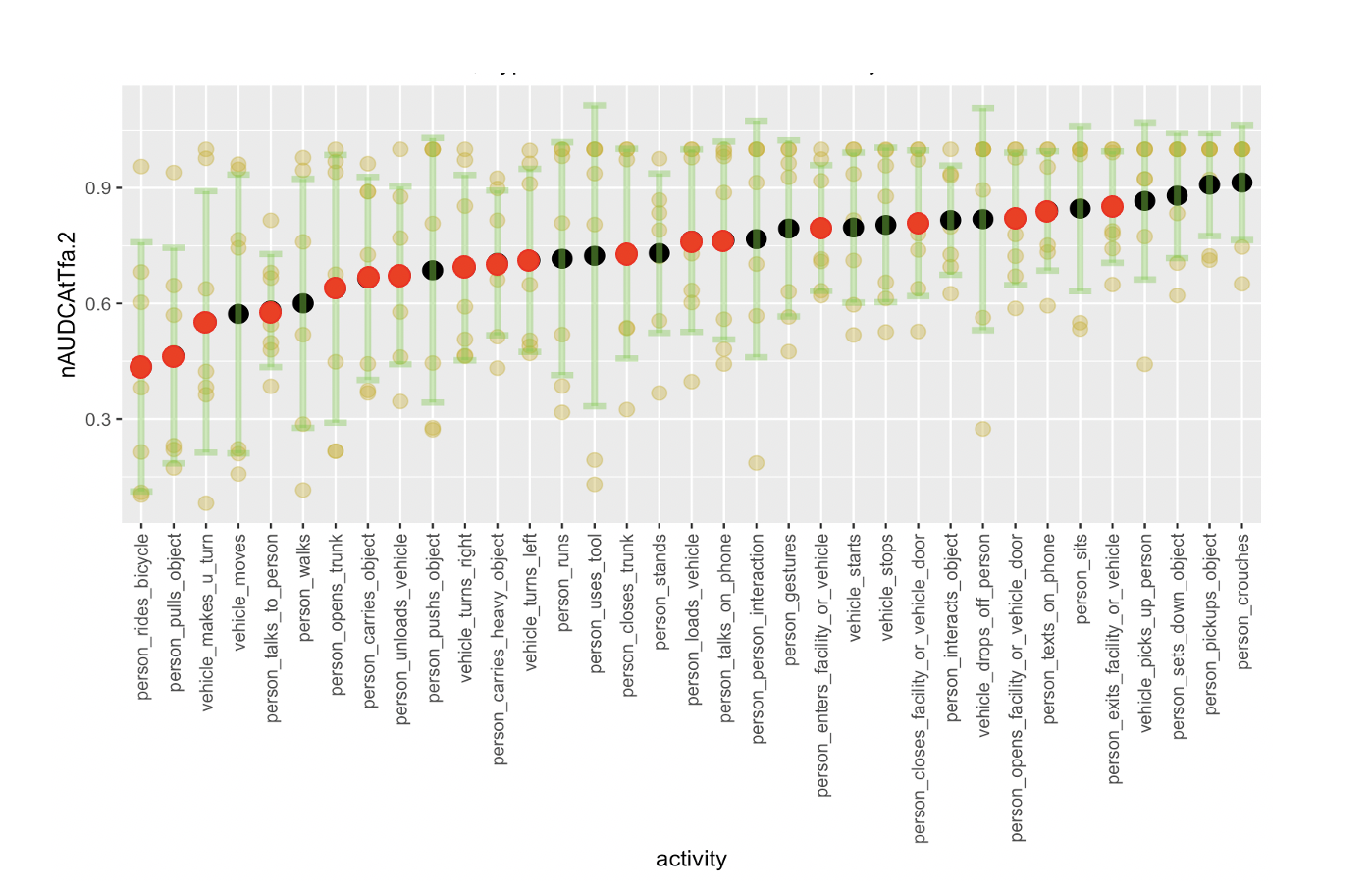}
\caption{Activity Ranking of the 7 systems which are ordered by $nAUDC$.
The x-axis shows the 35 activities. The y-axis shows the $nAUDC$. The red dots are the mean $nAUDC$ value of activities for 2019 and the black dots are for activities added for 2020.}
\label{fig_actev:4}
\end{centering}
\end{figure}

Figure \ref{fig_actev:4} illustrates the ranking of the activities across systems. The x-axis is the activity type and the y-axis is the metric $nAUDC$. The points marked in red and black indicate a mean values across different systems and the green error bars indicate the standard deviations.  The black and red dots represent the average performance of the 35 different activities. The red dots denote the 18 activities from  2019 and the black dot indicates new activities for 2020. The green points show the  $nAUDC$ values of the seven teams. The figure also shows that “person\_rides\_bicycle” is the easiest activity to detect while “person\_crouches” is the hardest.

\begin{figure}[htb]
\begin{centering}
\includegraphics[width=3.16667in,height=2.80000in]{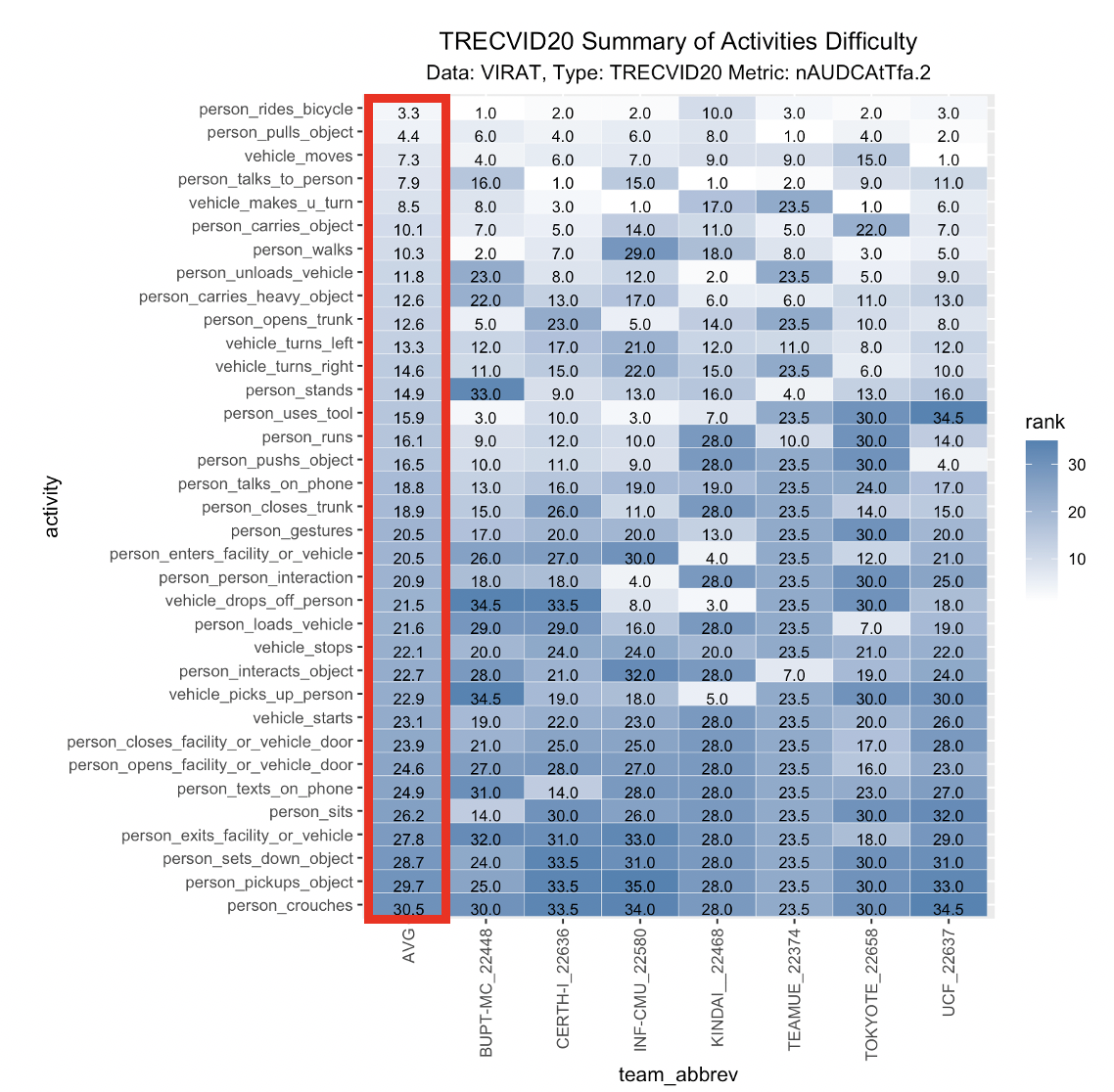}
\caption{Which activities are easier or more difficult to detect?}
\label{fig_actev:45}
\end{centering}
\end{figure}

Figure \ref{fig_actev:45} shows the activities that are ordered by the level of difficulty for each team. The x-axis shows the team names and average activity ranking (AVG), the y-axis, shows the 35 activities, and the numbers in the matrix show the rankings of 35 activities per system. The activity class was characterized by systems and baseline performance. The main observation is that the person\_rides\_bicycle, person\_pull\_object, and vehicle\_moves activities seem to be easier to detect compared to the rest of activities considered in this evaluation.

To examine the performance improvement from ActEV18 through ActEV20, Table \ref{table_actev:6} summarizes the leaderboard evaluation results from ActEV18, ActEV19 and ActEV20. Out of the 7 teams in current year participants, only 5 teams participated in either ActEV18 or ActEV19 evaluations. Note that, for this comparison, we had a slightly different dataset and number of activities, while using the same scoring protocol and performance measure (namely, PR.15: $P_{miss}$ at $R_{FA}$ = 0.15). System performance on 18 activities improved from ActEV19 to ActEV20 for CMU and  ITI\_CERTH.
We had an extra activity \say{interact} in the ActEV18 evaluation compared to ActEV19 and ActEV20.

\begin{table*}[htbp]
  \centering
  \caption {Comparison of ActEV18, ActEV19 and ActEV20 results for 18 activities. Since $P_{miss}$ at $R_{FA}$ = 0.15 was a primary measure for ActEV18, the ActEV19 column lists both $P_{miss}$ at $R_{FA}$ = 0.15 (PR.15$\downarrow$) and $nAUDC$ for comparison purpose.  }
\begin{tabular}{|l|c|c|c|c|c|c|}
\hline

\multicolumn{1}{|c|}{Team} & \multicolumn{2}{c|}{ActEV18 (LB19)} & \multicolumn{2}{c|}{ActEV19 (LB18)} & \multicolumn{2}{c|}{ActEV20 (LB18)}\\
\cline{2-7}
\multicolumn{1}{|c|}{} & PR.15$\downarrow$ & nAUDC & PR.15$\downarrow$ & nAUDC & PR.15$\downarrow$ & nAUDC \\
\hline

    UMD   & x     & x     & {x} & {x} & x     & x \\
    SeuGraph & x & x     & x &  x  & x     & x \\
    Team\_Vision & 0.709 & x     & {x} & {x} & x     & x \\
    UCF   & 0.733 & x     & 0.68  & 0.491 & 0.817 & 0.5188 \\
    STR-DIVA Team & x     & x     & {x} & {x} & x     & x \\
    JHUDIVATeam & x     & x     & {x} & {x} & x     & x \\
    CMU/INF & 0.844 & x     & 0.789 & 0.484 & 0.788 & 0.405 \\
    SRI   & x     & x     & {x} & {x} & x     & x \\
    VANT  & 0.882 & x     & {x} & {x} & x     & x \\
    HSMW\_TUC & x     & x     & 0.951 & 0.941 & x     & x \\
    BUPT-MCPRL & {0.749} & x     & 0.736 & 0.524 & {0.807} & 0.526 \\
    USF Bulls & {0.934} & x     & {x} & {x} & x     & x \\
    MKLab (ITI\_CERTH)  & x     & x     & 0.968 & 0.964 & 0.867 & 0.833 \\
    UTS-CETC & 0.925 & x     & {x} & {x} & 0.976 & 0.923 \\
    NII\_Hitachi\_UIT & 0.925 & x     & 0.819 & 0.599 & x     & x \\
    Fraunhofer IOSB & x     & x     & 0.849 & 0.827 & x     & x \\
    NTT\_CQUPT & x     & x     & 0.878 & 0.601 & x     & x \\
    vireoJD-MM & x     & x     & 0.714 & 0.601 & x     & x \\
    TokyoTech\_AIST & x     & x     & x     & x     & 0.821 & 0.689 \\
    Kindai\_Kobe & x     & x     & x     & x     & 0.950 & 0.959 \\
    \hline
    \end{tabular}%
  \label{table_actev:6}%
\end{table*}%

\subsubsection{Summary}\label{ActEV_Conclusion}
In this section, we presented the TRECVID ActEV20 evaluation task, the  performance metric and results for human activity detection. We primarily focused on the activity detection task only and the time-based false alarms were used to have a better understanding of system's behavior and to be more relevant to the use cases. The proposed metric was compared to the metric that incorporated instance-based false alarms and was used in ActEV18 evaluation. 
The ActEV20 activity names were made more consistent with the MEVA \cite{MEVAdata} dataset names and we added 17 more activities compared to prior year evaluations (35 target activities in total). Seven teams from 4 countries participated in the ActEV20 evaluation and made a total of 118 submissions. Given the test set and the 35 activities, overall, the person\_rides\_bicycle, person\_pull\_object, and vehicle\_moves activities seem to be easier to detect compared to the rest of activities considered in this evaluation. We provided a ranked list of system performances and examined the level of activity difficulty in detection using the submissions selected as the best performing system per team.
The TRECVID ActEV20 evaluation provided researchers an opportunity to evaluate their activity detection algorithms on a self-reported leaderboard. The competition also resulted in some progress in improving activity detection accuracy. We hope the TRECVID ActEV20 evaluation, and the associated datasets will facilitate the development of activity detection algorithms. This will in turn provide an impetus for more research worldwide in the field of activity detection in videos.

\subsection{Video Summarization}

An important need in many situations involving video collections (archive video search/reuse, personal video organization/search, movies, tv shows, etc.) is to summarize the video in order to reduce the size and concentrate the amount of high value information in the video track. In 2020 we introduced a new video summarization track in TRECVID in which the task was to summarize the major life events of specific characters over a number of weeks of programming on the BBC Eastenders TV series. The plan is to, every year, choose a few characters from a specific period of the show, and to ask participating teams to produce summaries for the character's major life events in that period.

The use case for this task is to generate an automatic summary, using a predefined maximum number of unique shots, of the significant life events of a given character from the Eastenders series over a given number of episodes. The generated summaries should be enough to gain a clear and concise overview of that characters major life events over the course of 8 - 12 weeks of programming in the series, and to see how they intertwine with the major life events of other specified character's in that time frame of the series.

\subsubsection{Video Summarization Data}
In 2020 this task embarked on a multi-year effort using 464 h of the BBC soap opera EastEnders. 244 weekly ``omnibus'' files were divided by the BBC into 471\,523 video clips to be used as the unit of retrieval. The videos present a ``small world'' with a slowly changing set of recurring people (several dozen), locales (homes, workplaces, pubs, cafes, restaurants, open-air market, clubs, etc.), objects (clothes, cars, household goods, personal possessions, pets, etc.), and views (various camera positions, times of year, times of day).

\subsubsection{System task}
The primary task for this track was, given a collection of BBC Eastenders videos, a master shot boundary reference, a list of characters from the series, and a time frame of the series, summarize the major life events of each character within the specified time frame of the series. Some examples of major life events were as follows: The birth of a child rather than a short illness, A divorce rather than an argument with a loved one, the passing of a loved one rather than the passing of someone loosely known to you. Summaries were limited to a maximum number of unique shots, thus the main challenge was to select those shots most likely to be considered a major life event by human assessors. 

Each topic consisted of a set of 4 example frame images in bitmap (bmp) file format drawn from test videos containing the person of interest in a variety of different appearances to the extent possible.

For each frame image (of a target person) there was a binary mask of the region of interest (ROI), as bounded by a single polygon and the ID from the master shot reference of the shot from which the image example was taken. In creating the masks (in place of a real searcher), we assumed the searcher wanted to keep the process simple. So, the ROI could contain non-target pixels, e.g., non-target regions visible through the target or occluding regions. 

\subsubsection{Topics}
By analyzing meta-data of the full set of BBC Eastenders omnibus episodes, NIST selected queries of three characters who were shown to play a big part in the series over a ten week period. The following three characters were selected:

\begin{itemize}
    \item Janine
    \item Ryan
    \item Stacey
\end{itemize}

In addition to specifying this year's query characters, the time frame of the series (Start Shot \# and End Shot \#), links to images of the query characters, and the maximum length and number of shots for each run were also disseminated to participating teams. These are indicated in Table \ref{vsumqueries}.

\begin{table*}[h]
\centering
\caption{Video Summarization Queries and Specifics}
\label{vsumqueries}
 \begin{tabular}{|c||c|c|c|} 
 \hline
 \textbf{Character} & Janine & Ryan & Stacey \\
 \hline
 \textbf{Start Shot \#} & shot175\_1 & shot175\_1 & shot175\_1 \\ 
 \hline
 \textbf{End Shot \#} & shot185\_1736 & shot185\_1736 & shot185\_1736 \\
 \hline
 \textbf{Max \# Shots Run 1} & 5 & 5 & 5 \\
 \hline
 \textbf{Max Summary Length Run 1} & 150 seconds & 150 seconds & 150 seconds \\
 \hline
 \textbf{Max \# Shots Run 2} & 10 & 10 & 10 \\
 \hline
 \textbf{Max Summary Length Run 2} & 300 seconds & 300 seconds & 300 seconds \\
 \hline
 \textbf{Max \# Shots Run 3} & 15 & 15 & 15 \\
 \hline
 \textbf{Max Summary Length Run 3} & 450 seconds & 450 seconds & 450 seconds \\
 \hline
 \textbf{Max \# Shots Run 4} & 20 & 20 & 20 \\
 \hline
 \textbf{Max Summary Length Run 4} & 600 seconds & 600 seconds & 600 seconds \\
 \hline
\end{tabular}
\end{table*}

\subsubsection{Evaluation}
Each team was asked to submit 4 runs, with the maximum number of shots and maximum summary length as specified in Table \ref{vsumqueries}. In total, 2 groups submitted 8 runs, with each run containing video summaries for each of the 3 specified queries, giving a total of 24 video summaries to be evaluated.

Submissions were evaluated by the TRECVID team at NIST, with one person responsible for evaluating summaries for a single query. Assessors answered 5 content based questions for each of the 8 video summaries they had been asked to evaluate. Content questions were created by the TRECVID team after watching each episode of the specified time frame of the series, marking those scenes they considered to be important, reducing these to 5 specific scenes based on what they considered to be the 5 most important scenes for each query, and finally voting on these as a group to establish the final 5 most important scenes for each character. From each of these, a question was worded to ask if the submitted video summary \textit{could be said} to have answered that question. The content questions for each character are specified below:

\begin{itemize}
    \item Janine
        \begin{enumerate}
            \item What is causing Ryan to be sick in bed?
            \item How does Janine attempt to kill Ryan while in the hospital?
            \item What happens when Janine attempts to play recording of Stacey?
            \item Who stabbed Janine?
            \item Who gives Janine the recording of Stacey?
        \end{enumerate}

    \item Ryan
        \begin{enumerate}
            \item How does Janine attempt to kill Ryan in the hospital?
            \item What does Ryan do when Janine is lying in the hospital?
            \item Where is Ryan trapped?
            \item What does Ryan tell Phil he can do for him?
            \item Who is Ryan with when going to put his name on the baby's birth cert?
        \end{enumerate}

    \item Stacey
        \begin{enumerate}
            \item Who climbs up to the roof to talk Stacey out of jumping off?
            \item What does Stacey reveal when in a cell with Janine, Kat, and Pat?
            \item What does Stacey admit to her mum in bedroom when mum is upset?
            \item Who confronts Stacey in restroom where Stacey finally admits to killing Archie?
            \item Who calls to Stacey's door to tell her to get her stuff and go, after Stacey's mum had called the police?
        \end{enumerate}
\end{itemize}

Assessors also marked video summaries on the subjective metrics of tempo/rhythm, contextuality, and redundancy, on a 7-point Likert-scale, with the following definitions. 
\textbf{Tempo/Rhythm} was defined as: \textit{How well do the video shots flow together? Do shots cut mid-sentence (indicating poor tempo/rhythm)? Do they flow together nicely so it wouldn't be obvious that this is an automatically generated summary (high tempo/rhythm)? (High is best)}. 
\textbf{Contextuality} was defined as: \textit{Does the content provide the circumstances that form the setting for an event, statement, or idea, and in terms of which it can be fully understood and assessed? (High is best)}. 
\textbf{Redundancy} was defined as: \textit{Does the video contain content considered to be unnecessary or superfluous? (Low is best)}.

\subsubsection{Metrics}
Scores were calculated as a percentage using marks for the 5 content based questions and the 3 subjective quality based questions. Base Likert-scale scores for Tempo/Rhythm and Contextuality were taken as assessed by human annotators. Scores for Redundancy, where a lower score is better, were flipped. This gave a total of 21 possible marks available for subjective quality scores. The remainder was calculated by taking the remaining 79 possible marks and dividing by the 5 content based questions, giving a total of 15.8 possible marks for each correct content based question which was to be rounded to the nearest integer. This would give a perfect summary 100 points. A summary with no relevant content but all perfect scores for the other factors would get 21 points. Overall this gave summaries a maximum score of 100 down to a minimum score of 3.

\subsubsection{Results}
Table \ref{vsum.results} shows the individual results for each submission query and run on all metrics and content questions. Team Me\_MAD consistently achieves the best results.

\begin{figure}[htbp]
\begin{center}
\includegraphics[height=2.5in,width=3in,angle=0]{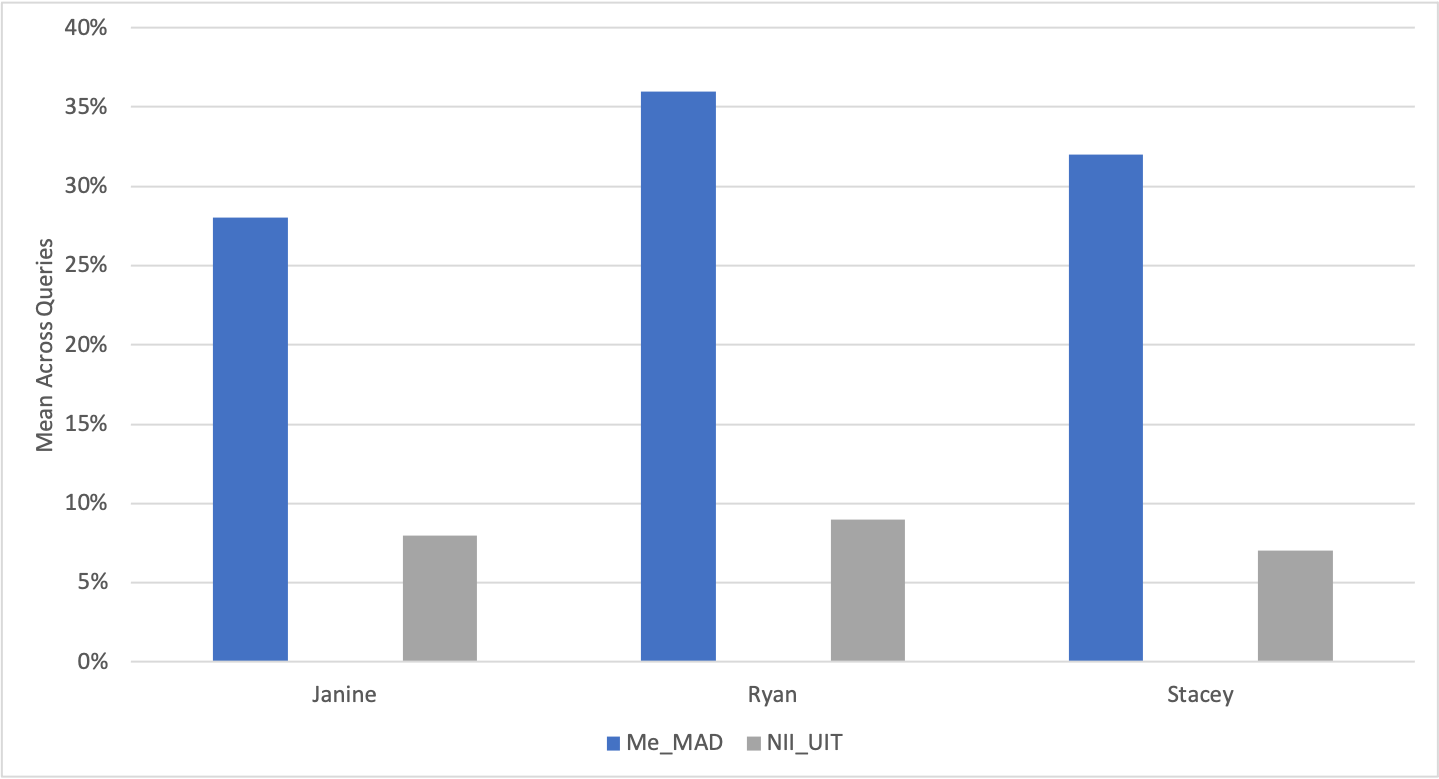}
\caption{VSUM: Average scores by Character}
\label{vsum.across.query}
\end{center}
\end{figure}

Figure \ref{vsum.across.query} shows the average scores
for each target query by team. Scores are averaged across all runs. \textbf{Ryan} is seen to be the easiest character to summarize, with \textbf{Janine} the most difficult. However the only differences between \textbf{Janine} and \textbf{Stacey} are in the subjective measures of Tempo, Contextuality and Redundancy.

\begin{figure}[htbp]
\begin{center}
\includegraphics[height=2.5in,width=3in,angle=0]{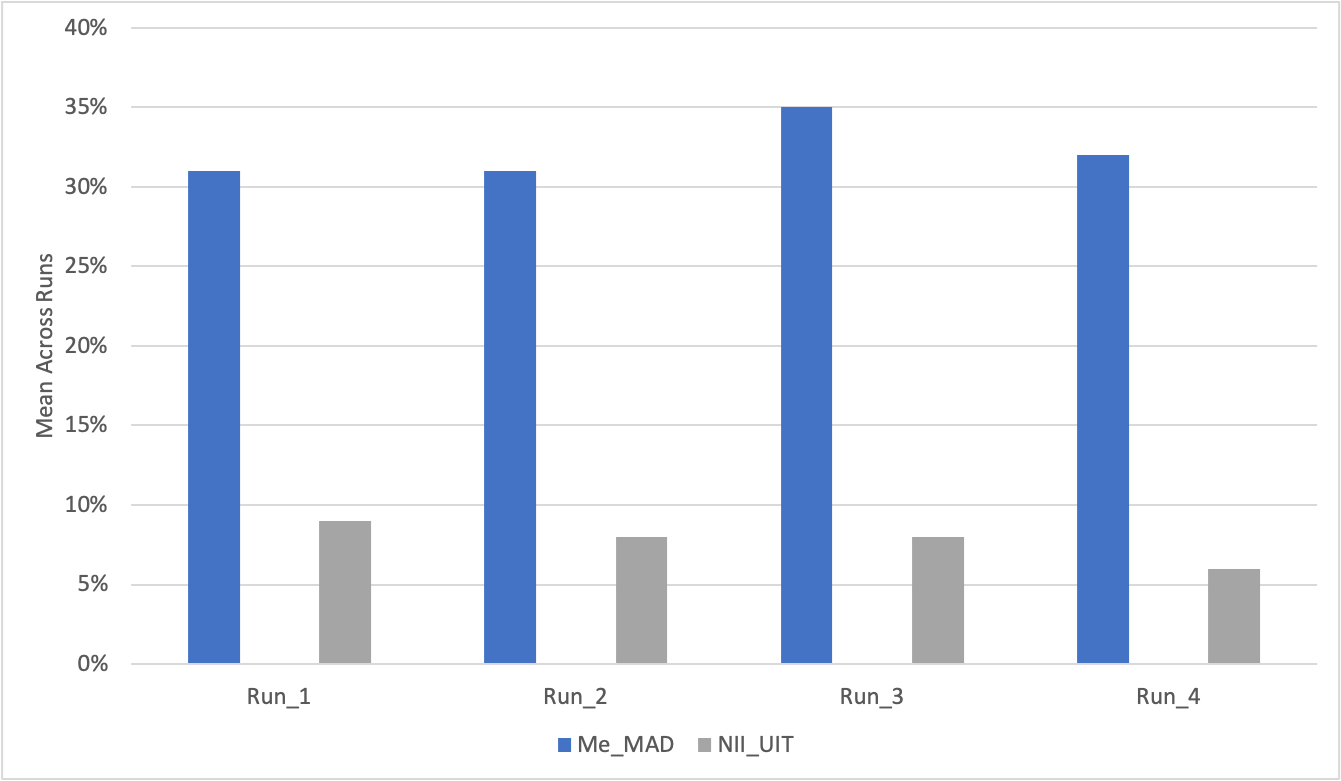}
\caption{VSUM: Average scores for each run}
\label{vsum.across.run}
\end{center}
\end{figure}

Figure \ref{vsum.across.run} shows the average scores
for each target query by team. Scores are averaged across all target queries. Run 3, containing 15 shots and up to a maximum of 450 seconds in length scores higher than other runs. This is most affected by target query \textbf{Ryan} who scored one extra content question by one team for runs 3 and 4. Run 4 average scores were reduced by lost marks in Tempo, Contextuality and Redundancy.

\begin{figure}[htbp]
\begin{center}
\includegraphics[height=2.5in,width=3in,angle=0]{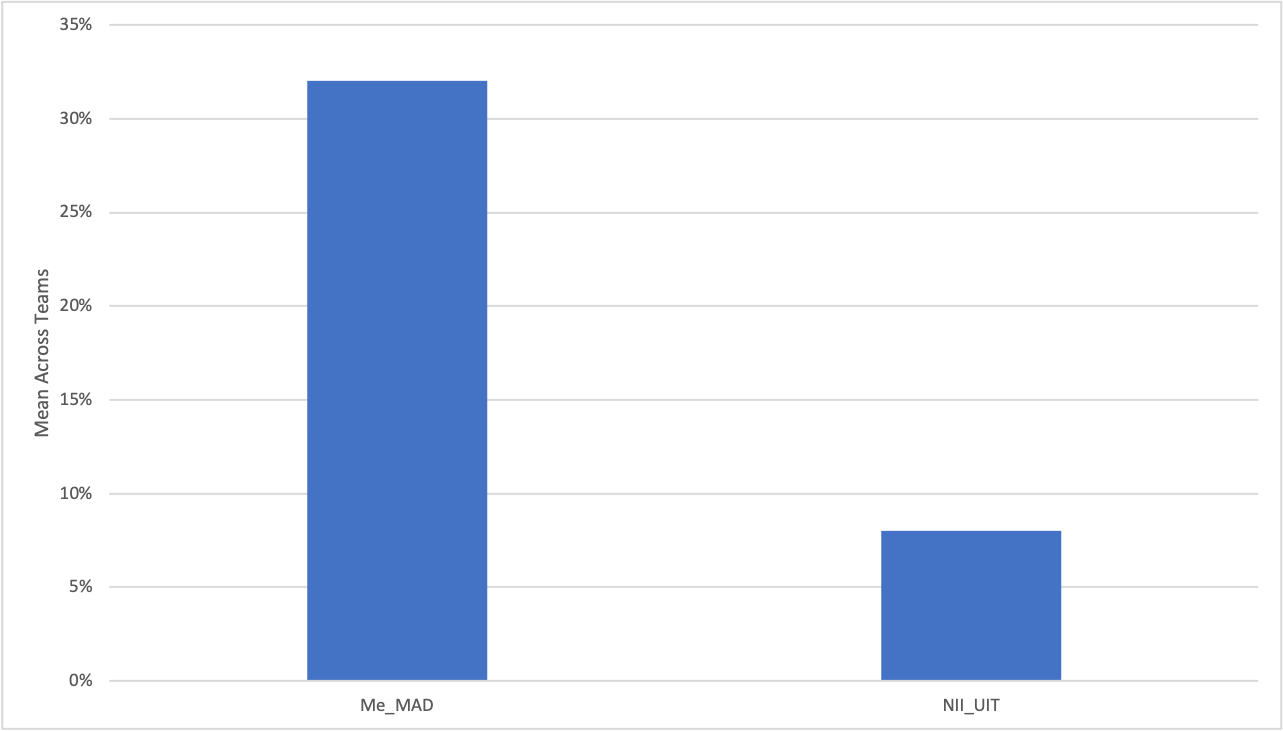}
\caption{VSUM: Average scores by team}
\label{vsum.across.team}
\end{center}
\end{figure}

Figure \ref{vsum.across.team} shows the average scores
for each team. Scores are averaged across all runs and target queries. Team Me\_MAD achieves consistently higher marks, mostly due to summaries containing at least one correct content question.

\begin{figure}[htbp]
\begin{center}
\includegraphics[height=2.5in,width=3in,angle=0]{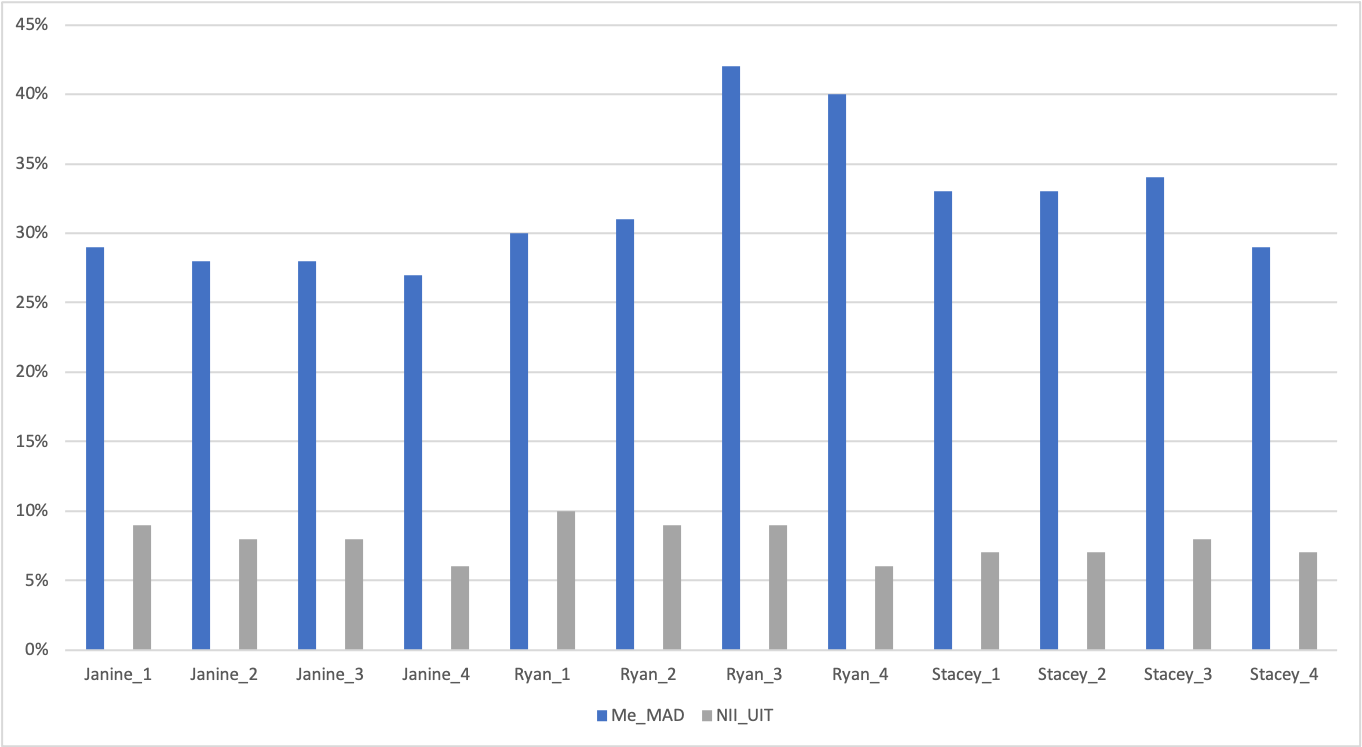}
\caption{VSUM: Individual scores}
\label{vsum.individual}
\end{center}
\end{figure}

Figure \ref{vsum.individual} shows the individual scores
for all teams, runs and target queries. This chart visualizes the final results shown in table \ref{vsum.results}, from which it can be seen that team Me\_MAD scores higher for \textbf{Ryan} run 3 and \textbf{Ryan} run 4 than for all other submitted summaries.

\begin{table*}[h]
\centering
\caption{Video Summarization Queries and Specifics}
\label{vsum.results}
 \begin{tabular}{|c||c|c|c|c|c|c|c|c|c|} 
 \hline
 \textbf{Team\_Run\_Query} & \textbf{Tempo} & \textbf{Contextuality} & \textbf{Redundancy} & \textbf{Q1} & \textbf{Q2} & \textbf{Q3} & \textbf{Q4} & \textbf{Q5} & \textbf{Score}\\
 \hline \hline
 Me\_MAD\_1\_Janine & 6 & 4 & 5 & No & No & No & No & Yes & 29\% \\ 
 \hline
 Me\_MAD\_2\_Janine & 5 & 5 & 6 & No & No & No & No & Yes & 28\% \\
 \hline
 Me\_MAD\_3\_Janine & 5 & 5 & 6 & No & No & No & No & Yes & 28\% \\
 \hline
 Me\_MAD\_4\_Janine & 5 & 5 & 7 & No & No & No & No & Yes & 27\% \\
 \hline
 NII\_UIT\_1\_Janine & 5 & 3 & 7 & No & No & No & No & No & 9\% \\
 \hline
 NII\_UIT\_2\_Janine & 4 & 3 & 7 & No & No & No & No & No & 8\% \\
 \hline
 NII\_UIT\_3\_Janine & 4 & 3 & 7 & No & No & No & No & No & 8\% \\
 \hline
 NII\_UIT\_4\_Janine & 2 & 3 & 7 & No & No & No & No & No & 6\% \\
 \hline
 Me\_MAD\_1\_Ryan & 4 & 5 & 3 & No & No & No & No & Yes & 30\% \\
 \hline
 Me\_MAD\_2\_Ryan & 5 & 5 & 3 & No & No & No & No & Yes & 31\% \\
 \hline
 Me\_MAD\_3\_Ryan & 3 & 4 & 5 & No & No & No & Yes & Yes & 42\% \\
 \hline
 Me\_MAD\_4\_Ryan & 2 & 3 & 5 & No & No & No & Yes & Yes & 40\% \\
 \hline
 NII\_UIT\_1\_Ryan & 4 & 3 & 5 & No & No & No & No & No & 10\% \\
 \hline
 NII\_UIT\_2\_Ryan & 3 & 3 & 5 & No & No & No & No & No & 9\% \\
 \hline
 NII\_UIT\_3\_Ryan & 2 & 4 & 5 & No & No & No & No & No & 9\% \\
 \hline
 NII\_UIT\_4\_Ryan & 2 & 2 & 6 & No & No & No & No & No & 6\% \\
 \hline
 Me\_MAD\_1\_Stacey & 6 & 5 & 2 & No & Yes & No & No & No & 33\% \\ 
 \hline
 Me\_MAD\_2\_Stacey & 6 & 5 & 2 & No & Yes & No & No & No & 33\% \\
 \hline
 Me\_MAD\_3\_Stacey & 6 & 6 & 2 & No & Yes & No & No & No & 34\% \\
 \hline
 Me\_MAD\_4\_Stacey & 4 & 5 & 4 & No & Yes & No & No & No & 29\% \\
 \hline
 NII\_UIT\_1\_Stacey & 3 & 3 & 7 & No & No & No & No & No & 7\% \\
 \hline
 NII\_UIT\_2\_Stacey & 3 & 3 & 7 & No & No & No & No & No & 7\% \\
 \hline
 NII\_UIT\_3\_Stacey & 3 & 2 & 5 & No & No & No & No & No & 8\% \\
 \hline
 NII\_UIT\_4\_Stacey & 3 & 2 & 6 & No & No & No & No & No & 7\% \\
 \hline
\end{tabular}
\end{table*}

\subsubsection{Observations}
This is the first year of the video summarization task. Due to this, the decision was taken to require that teams submit results for 4 different runs, specified by a maximum number of shots and maximum summary length in seconds. While it was found that the conditions for run 3 scores higher, this is based on too few submissions to make a reliable judgment. Hence for the next year of the task these same conditions will apply. Following that, the plan is to specify that runs differentiate between techniques applied and configurations used. All submitted summaries used all of the maximum number of shots allowed, however no submitted summaries were anywhere near as long as the specified maximum length, thus the maximum length will be greatly reduced for the next year of the task.

We should also note that this year a significant amount of time and effort was spent trying to get the data agreement set with the donor (BBC) which may have adversely affected the number of teams who did not get enough time to work on and finish the task. 

We now summarize the approaches taken by teams. Team MeMAD proposed a fan-driven and character-centered approach to video summarization. In addition to the BBC Eastenders data made available to teams participating in the task, they also used fan-made content from the BBC Eastenders Fandom Wiki and character images crawled from a search engine. In the first step of their approach, they scraped synopsis found on the Eastenders Fandom Wiki with the hypothesis that every sentence represents an important event to be included in summaries. In parallel, they extracted shots from the series in which the three target queries appear using Face Celebrity Recognition. This detects faces using Multi-task Cascaded Convolutional Neural Networks (MTCNN) algorithm \footnote{https://github.com/ipazc/mtcnn}, and the FaceNet \footnote{https://github.com/davidsandberg/facenet} model is applied to get face embeddings.

Team NII\_UIT proposed a framework to generate final summaries by combining the score of a person's facial detection scores and a self-attention based network. They divided their network into three separate modules - segmentation, score, selection. Videos were divided into short segments using the shot time information on the BBC Eastenders data. They then calculated an importance score for each shot by combining the person score of facial detection similarity, and representation score. For person score, they use their INS 2019 system which includes a face detector, a face descriptor, and a face matching component. For calculating the representation score, they used VASNet \cite{fajtl2018summarizing}, an approach to sequence transformation for video summarization based on soft, self-attention mechanism. The final importance score of each shot was calculated by summing the person face and representation score.

\subsubsection{Conclusions}
This was the first year of the Video Summarization task. Teams were asked to produce summaries of the major life events of three target characters from within a specified time frame of the BBC Eastenders series. The major challenges of this task were to locate only shots for the target queries and to identify those shots most likely to have been considered major life events.

There were a total of 2 finishing teams out of 12 participating teams in this year's task. All 2 finishing teams submitted notebook papers and presented their approaches at the TRECVID workshop. It should also be noted that the long delay in teams gaining access to the data set may have adversely affected the number of teams who were able to complete the task.

\section{Summing up and moving on}
In this overview paper to TRECVID 2020, we provided basic information for all
tasks we run this year and particularly on the goals, data, evaluation mechanisms, and metrics used. 
Further details about each particular group's approach and performance for each task 
can be found in that group's site report. The raw results for each submitted run 
can be found at the online proceeding of the workshop \cite{tv20pubs}.
Finally, we are looking forward to continuing a new evaluation cycle in 2021 after
refining the current tasks and introducing any potential new tasks.
\section{Authors' note}
TRECVID would not have happened in 2020 without support from the
National Institute of Standards and Technology (NIST). The research
community is very grateful for this. Beyond that, various individuals
and groups deserve special thanks:
\begin{itemize}

\item{Koichi Shinoda of the TokyoTech team agreed to host a copy of 
IACC.2 data.}

\item{Georges Qu\'{e}not provided the master shot reference for the
IACC.3 videos.}

\item{The LIMSI Spoken Language Processing Group and Vocapia Research
provided ASR for the IACC.3 videos.}

\item{Luca Rossetto of University of Basel for providing the V3C dataset collection.}

\item{Noel O'Connor and Kevin McGuinness at Dublin City University
  along with Robin Aly at the University of Twente worked with NIST
  and Andy O'Dwyer plus William Hayes at the BBC to make the BBC
  EastEnders video available for use in TRECVID. Finally, Rob Cooper at BBC
  facilitated the copyright licence agreement for the Eastenders data.}

\item{Jeffrey Liu and Andrew Weinert of MIT Lincoln Laboratory for supporting the DSDI task
      by making the LADI dataset available and helping with the testing dataset preparations.} 
\end{itemize}

Finally we want to thank all the participants and other contributors
on the mailing list for their energy and perseverance.

\section{Acknowledgments} 
The ActEV NIST work was supported by the Intelligence Advanced Research Projects Activity (IARPA), agreement~IARPA-16002, order R18-774-0017. The authors would like to thank Kitware, Inc. for annotating the dataset. 
The Video-to-Text work has been partially supported by Science Foundation Ireland (SFI) as a part of the Insight Centre at Dublin City University (12/RC/2289) and grant  number  13/RC/2106 (ADAPT Centre for Digital Content Technology, \url{www.adaptcentre.ie})  at  Trinity College Dublin.
We would like to thank Tim Finin and Lushan Han of University of Maryland, Baltimore County for providing access to the semantic similarity metric.
Finally, the TRECVID team at NIST would like to thank all external coordinators for their efforts across the different tasks they helped to coordinate.

\bibliography{video}

\clearpage
\onecolumn

\appendix
\section{Ad-hoc query topics - 20 unique} 
\label{appendixA}
\begin{description}\itemsep0pt \parskip0pt

\item[641]  Find shots showing an aerial view of buildings near water in the daytime
\item[642]  Find shots of a person paddling kayak in the water
\item[643]  Find shots of people dancing or singing while wearing costumes outdoors
\item[644]  Find shots of sailboats in the water
\item[645]  Find shots of a person wearing a necklace
\item[646]  Find shots of a woman sitting on the floor
\item[647]  Find shots of people or cars moving on a dirt road
\item[648]  Find shots of a man in blue jeans outdoors
\item[649]  Find shots of someone jumping while snowboarding
\item[650]  Find shots of one or more people drinking wine
\item[651]  Find shots of one or more people skydiving
\item[652]  Find shots of a little boy smiling
\item[653]  Find shots of group of people clapping
\item[654]  Find shots of one or more persons exercising in a gym
\item[655]  Find shots of one or more persons standing in a body of water
\item[656]  Find shots of a long haired man
\item[657]  Find shots of a woman with short hair indoors
\item[658]  Find shots of two or more people under a tree
\item[659]  Find shots of a church from the inside
\item[660]  Find shots of train tracks during the daytime

\end{description}

\section{Ad-hoc query topics - 20 progress topics} 
\label{appendixB}
\begin{description}\itemsep0pt \parskip0pt
\item[591]  Find shots of a person holding an opened umbrella outdoors
\item[592]  Find shots of a person reading a paper including newspaper
\item[593]  Find shots of one or more women models on a catwalk demonstrating clothes
\item[594]  Find shots of people doing yoga
\item[595]  Find shots of a person sleeping
\item[596]  Find shots of fishermen fishing on a boat
\item[597]  Find shots of a shark swimming under the water
\item[598]  Find shots of a man in a clothing store
\item[599]  Find shots of a person in a bedroom
\item[600]  Find shots of a person's shadow
\item[601]  Find shots of a person jumping with a motorcycle
\item[602]  Find shots of a person jumping with a bicycle
\item[603]  Find shots of people hiking
\item[604]  Find shots of bride and groom kissing
\item[605]  Find shots of a person skateboarding
\item[606]  Find shots of people queuing
\item[607]  Find shots of two people kissing who are not bride and groom
\item[608]  Find shots of two people talking to each other inside a moving car
\item[609]  Find shots of people walking across (not down) a street in a city
\item[610]  Find shots showing electrical power lines

\end{description}

\section{Instance search topics - 20 unique} 
\label{appendixC}
\begin{description} \itemsep0pt

\item[9299] Find Ian sitting on couch
\item[9300] Find Billy sitting on couch
\item[9301] Find Ian Holding paper - including photos/envelope,notebooks, magazines, etc
\item[9302] Find Bradley Holding paper - including photos/envelope,notebooks, magazines, etc
\item[9303] Find Billy Holding paper - including photos/envelope,notebooks, magazines, etc
\item[9304] Find Max Drinking
\item[9305] Find Dot Drinking
\item[9306] Find Pat Holding cloth - including jackets, coats, kitchen towels, cleaning towels, etc
\item[9307] Find Heather Holding cloth - including jackets, coats, kitchen towels, cleaning towels, etc
\item[9308] Find Ian Crying
\item[9309] Find Heather Crying
\item[9310] Find Max smoking a cigarette - including holding a cigarette between fingers
\item[9311] Find Dot smoking a cigarette - including holding a cigarette between fingers
\item[9312] Find Pat smoking a cigarette - including holding a cigarette between fingers
\item[9313] Find Stacey Laughing
\item[9314] Find Pat Laughing
\item[9315] Find Max Going up or down the stairs
\item[9316] Find Bradley Going up or down the stairs
\item[9317] Find Max holding a phone / handset - including talking on phone
\item[9318] Find Stacey holding a phone / handset - including talking on phone

\end{description}

\section{Instance search topics - 20 progress topics} 
\label{appendixD}
\begin{description} \itemsep0pt

\item[9279] Find Phil Sitting on a couch
\item[9280] Find Heather Sitting on a couch
\item[9281] Find Jack Holding phone
\item[9282] Find Heather Holding phone
\item[9283] Find Phil Drinking
\item[9284] Find Shirley Drinking
\item[9285] Find Jack Kissing
\item[9286] Find Denise Kissing
\item[9287] Find Phil Opening door and entering room / building
\item[9288] Find Sean Opening door and entering room / building
\item[9289] Find Shirley Shouting
\item[9290] Find Sean Shouting
\item[9291] Find Stacey Hugging
\item[9292] Find Denise Hugging
\item[9293] Find Max Opening door and leaving room / building
\item[9294] Find Stacey Opening door and leaving room / building
\item[9295] Find Max Standing and talking at door
\item[9296] Find Dot Standing and talking at door
\item[9297] Find Jack Closing door without leaving
\item[9298] Find Dot Closing door without leaving
    
\end{description}

\end{document}